\documentclass[preprint,review]{elsarticle}

\usepackage[T1]{fontenc}

\usepackage{lineno,hyperref}
\usepackage{caption}
\usepackage{subcaption}

\usepackage{algorithm}
\usepackage{algpseudocode}

\usepackage{multirow}

\usepackage{bbm}
\usepackage{dsfont}

\usepackage{setspace}
\usepackage[percent]{overpic}

\usepackage{xcolor} % http://www.ctan.org/tex-archive/macros/latex/contrib/xcolor
\usepackage{tikz}
\usetikzlibrary{fit,calc}
\colorlet{neg}{red!40}
\colorlet{pos}{cyan!60}

\journal{Pattern Recognition}

\usepackage{booktabs}

\usepackage{dsfont}
\usepackage{xcolor}
\usepackage{multirow}
\usepackage[normalem]{ulem}
\useunder{\uline}{\ul}{}

\newcommand{\R}{\mathds{R}}

\newcommand{\Ls}{\mathcal{L}}
\newcommand{\T}{\mathcal{T}}
\newcommand{\D}{\mathcal{D}}

\newcommand{\F}{\mathcal{F}}

\DeclareMathAlphabet\mathbfcal{OMS}{cmsy}{b}{n}

\usepackage[symbol]{footmisc}
\title{Meta-Learners for Few-Shot Weakly-Supervised Medical Image Segmentation}

\author{Hugo Oliveira\corref{cor1}}
\author{Pedro H. T. Gama}
\author{Isabelle Bloch}
\author{Roberto Marcondes Cesar Jr}

\begin{document}

\begin{frontmatter}

\renewcommand*{\thefootnote}{\fnsymbol{footnote}}

% \doublespacing

\begin{abstract}
Most uses of Meta-Learning in visual recognition are very often applied to image classification, with a relative lack of works in other tasks {such} as segmentation and detection. We propose a generic Meta-Learning framework for few-shot weakly-supervised segmentation in medical imaging domains. We conduct a comparative analysis of meta-learners from distinct paradigms adapted to few-shot image segmentation in different sparsely annotated radiological tasks. The imaging modalities include 2D chest, mammographic and dental X-rays, as well as 2D slices of volumetric tomography and resonance images. Our experiments consider a total of 9 meta-learners, 4 backbones and multiple target organ segmentation tasks. We explore small-data scenarios in radiology with varying weak annotation styles and densities. Our analysis shows that metric-based meta-learning approaches achieve better segmentation results in tasks with smaller domain shifts in comparison to the meta-training datasets, while some gradient- and fusion-based meta-learners are more generalizable to larger domain shifts.
\end{abstract}


\end{frontmatter}

% \newpage
\renewcommand*{\thefootnote}{\arabic{footnote}}

\newcommand{\currprop}{\textwidth}

%\linenumbers

% Setting table column separation.
\setlength\tabcolsep{4.5pt} % default: 6pt

\section{Introduction}
\label{sec:introduction}

% During the last decade, automated visual recognition has been largely reshaped by deep learning. Powerful end-to-end trainable neural networks based on convolutional operations \cite{krizhevsky2012imagenet} shifted the main limitation of machine learning models from the lack of representation capacity of shallow feature extractors to the lack of sufficient annotated samples. More recently, machine learning algorithms have become ubiquitous in medical image analysis, being an integral part of almost any image-based pathology detection/diagnosis pipeline. Deep learning has been aiding physicians on multiple fronts, including disease diagnosis, cancer detection, surgery planning, and organ segmentation \cite{mcbee2018deep}.

Despite the widespread use of deep learning in medical imaging, neural networks are still subject to a series of limitations that hamper their use in most real-world medical settings. The main hurdle in implementing Deep Neural Networks (DNNs) \cite{krizhevsky2012imagenet} into clinical practice is the data-driven nature of these models, which usually require hundreds, or even thousands of samples per class to fit properly, running the risk of suffering from underfitting or overfitting in small-data scenarios. A compounding factor to this problem is the presence of domain shift in real-world cases. In medical imaging, domain shift can be introduced to a task due to changes in imaging equipment or settings, differences in the cohort of training/test samples, imaging and label modalities, etc. Regarding segmentation tasks, an additional complication in the use of deep learning is the cost of producing the labels to be fed to the algorithms, as dense pixelwise labels are known to be expensive and to require highly specialized anatomical knowledge from the physician. In this case, sparse labels fed to weakly-supervised segmentation algorithms can be a good compromise between annotation cost and performance \cite{peng2021medical}. Still, the majority of works on one-/few-shot image segmentation do not consider the case of sparsely labeled images, only focusing on the case of fully labeled support sets \cite{oliveira2022domain}.%hu2019attention,zhang2019canet,zhang2020sg}. % \cite{shaban2017one,hu2019attention,zhang2019pyramid,zhang2019canet,zhang2020sg}. 

In this paper, we address the task of domain generalization from Few-shot Weakly-supervised Segmentation (FWS). In other words, we are interested in mitigating the limitations of semantic image segmentation with DNNs in problems with small-data, weak supervision for segmentation, and completely distinct image and label spaces during training and testing.

% In RGB images, ImageNet \cite{deng2009imagenet} pretraining is often a prior knowledge introduced into few-shot segmentation and detection models \cite{shaban2017one,rakelly2018few,wang2019panet,hu2019attention,zhang2019pyramid,zhang2019canet,zhang2020sg}. The large variety of ImageNet samples is used as a strong feature extractor that can be leveraged to achieve good segmentation results with very few samples by replacing the prediction head of a traditional Convolutional Neural Network (CNN) by a segmentation head. More recently, pretraining on ImageNet samples without using their label information on Self-Supervised Learning (SSL) has been proven to be a more robust initialization to such tasks \cite{he2020momentum,chen2021exploring} than supervised ImageNet pretraining. However, such strategies are often limited to natural images, not presenting considerable gains in comparison to random initializations for domains such as medical imaging. Additionally, works that claim to solve FWS in non-RGB domains often rely on task-specific priors \cite{bearman2016s} that are not portable to other visual domains or Class Activation Maps (CAMs) \cite{du2021weakly} that assume supervised pretraining.

One common key point in few-shot learning is to introduce some form of prior knowledge into the models. A simple solution to this is to pretrain the model on a larger dataset. In RGB image domains there are several datasets with thousands or even millions of annotated samples (i.e. ImageNet~\cite{deng2009imagenet}, Instagram 1B~\cite{yalniz2019billion}, Pascal VOC~\cite{everingham2015pascal}, Flickr30k~\cite{young2014image}, etc.), which can be used to produce feature extractors to be leveraged in order to achieve good segmentation results with very few samples. Recently, Self-Supervised Learning (SSL) has been proven to be a more robust initialization to such tasks %\cite{he2020momentum,chen2021exploring}
\cite{jing2020self} than supervised ImageNet pretraining. However, such strategies are often limited to natural images, not presenting considerable gains in comparison to random initializations for domains such as medical imaging. Another approach to introduce prior knowledge is to use meta-learning training. Already being successfully used in other pattern recognition tasks, it has been employed in few-shot image classification \cite{hospedales2021meta}, with recent approaches for semantic segmentation gaining popularity \cite{luo2022meta}. Still, the majority of works rely on ImageNet pretraining as feature extractors, while also not conducting tests on weak labels. In this scenario, there is currently a gap in reliable methods for FWS on non-RGB images.

We propose a framework capable of strong pretraining using Meta-Learning on radiological images that does not assume task-specific priors, does not require previous pretraining of the backbone prior to our meta-training phase%(e.g. on ImageNet \cite{deng2009imagenet}, MS COCO \cite{lin2014microsoft} or Pascal VOC \cite{everingham2015pascal})
, and, therefore, can be generalized to any FWS task in radiology.
% Differently from Domain Adaptation (DA) tasks, the solutions described in this work concern tasks wherein the target domain is related, but not available during training, which is a requirement for most knowledge transfer approaches based on transduction \cite{arnold2007comparative}. Our tasks are quite close to Domain Generalization, following the taxonomy proposed by Zhang \textit{et al.} \cite{zhang2017transfer}, as our framework uses pixelwise labeled data from multiple radiology domains (i.e. chest x-rays, mammographies, dental x-rays, etc) aiming to be generalizable to other medical imaging tasks (i.e. computed tomography, magnetic resonance, positron emission tomography, etc). From the dense masks of the meta-dataset we simulate multiple weak (sparse) annotation styles and densities during the meta-training in order to enforce agnosticism towards the target weak labeling.
An overview of the proposed framework can be seen in Figure~\ref{fig:overview}. During its meta-training phase, it uses a meta-dataset $\mathbf{\D} = \{ \D_{1}, \D_{2}, \dots, \D_{n} \}$, then it is deployed on an out-of-distribution (OOD) few-shot target task $\F$, where the single highly generalizable model $\phi$, trained via a selective supervised loss function $\Ls$, is used as a predictor. As discussed in the following sections, $\phi$ can be trained in several distinct ways, such as second-order optimization, metric learning, and late fusion.
Differently from Domain Adaptation (DA) tasks, the solutions described in this work concern tasks wherein the target domain is related to the meta-training data, but not available during training. %, which is a requirement for most knowledge transfer approaches based on transduction \cite{arnold2007comparative}.
Our tasks are closer to Domain Generalization \cite{wang2022generalizing} scenarios,
%, following the taxonomy proposed by Zhang \textit{et al.}~\cite{zhang2017transfer}, 
since our framework uses pixelwise labeled data from multiple radiology domains (chest X-rays, mammographies, dental X-rays, etc.) aiming to be generalizable to other medical imaging tasks (i.e. computed tomography, magnetic resonance, positron emission tomography, etc.). From the dense ground truth masks of the meta-dataset, we simulate multiple weak (sparse) annotation styles and densities during the meta-training in order to enforce agnosticism toward the target weak labeling.

\renewcommand{\currprop}{0.9\textwidth}
\begin{figure*}[!ht]
    \centering
    \includegraphics[clip,trim=1.0in 0in 0in 0in,width=\currprop]{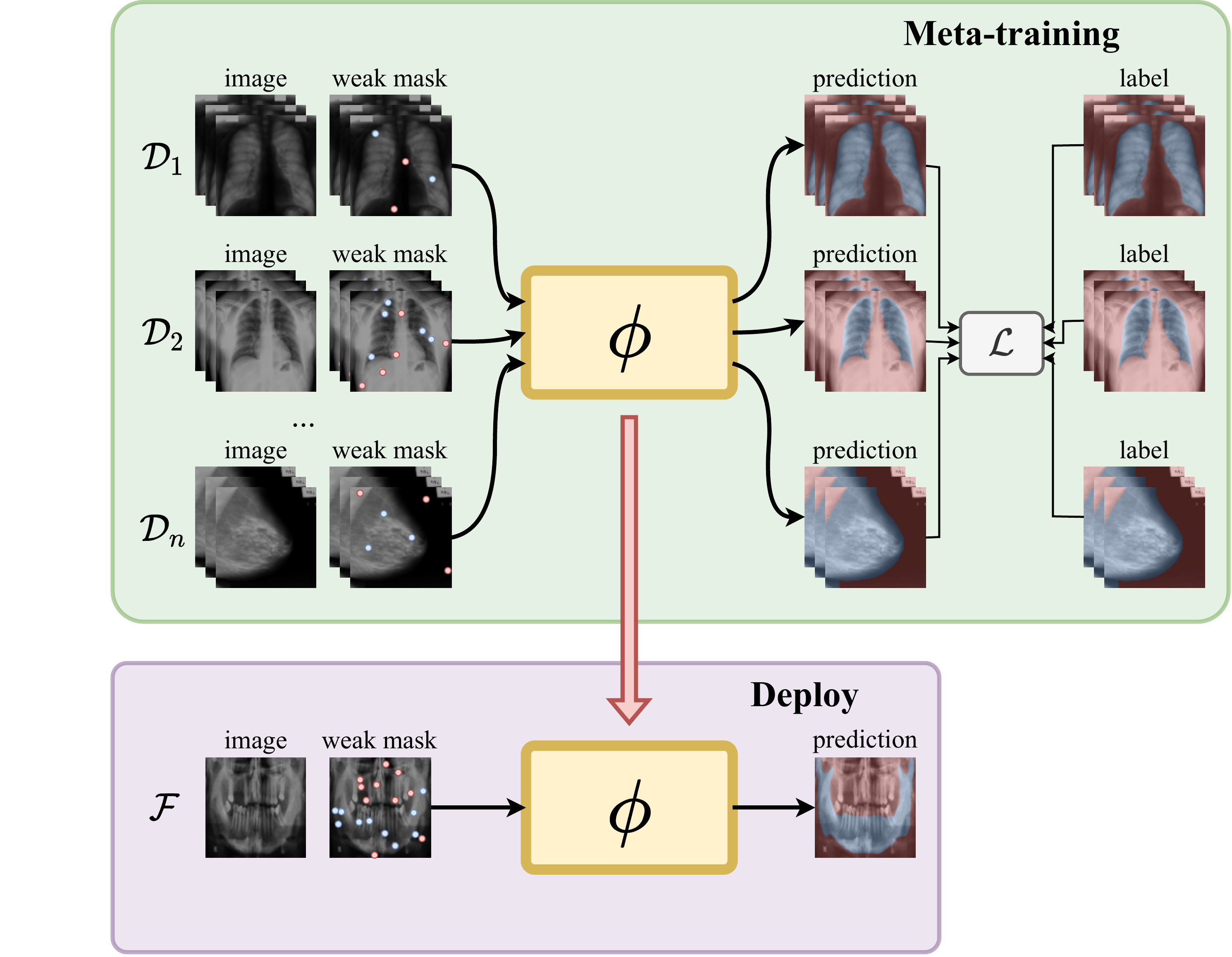}
    \caption{Overview of a meta-learner $\phi$ for FWS being pretrained in an episodic fashion on multiple related tasks $\{\D_{1}, \D_{2}, \dots, \D_{n}\}$ pertaining to a meta-dataset (top) and being deployed to a target few-shot task $\F$ with a sparsely annotated support set.}
    \label{fig:overview}
\end{figure*}

% \todo{Domains distinct from RGB images that do not have ImageNet for pretraining. Also, many algorithms are only designed for sparse labeling scenarios (i.e. classification or regression) and not for dense labeling (i.e. segmentation, detection).}

% \todo{Cost of annotations and label availability in medical imaging makes weakly-supervised segmentation especially useful in such applications.}

% \todo{Most of the literature on FWS with sparse labels rely on either supervised pretraining \cite{rakelly2018few}, strong task-specific priors \cite{bearman2016s} or Class Activation Maps (CAMs) \cite{du2021weakly}.}

% \todo{Mention Meta-Learning, SSL and Semi-Supervised Learning as alternatives to fully supervised learning.}

We highlight the main contributions of this work as: 1) a generalizable approach to porting meta-learners from multiple paradigms (see related work in Section~\ref{sec:related}) for image classification to be applied to FWS (Section~\ref{sec:meta_learning}); 2) a detailed performance analysis of the multiple Meta-Learning algorithms being adapted to multiple FWS in 2D radiology (Section~\ref{sec:results_intra}); and 3) analysis of multiple annotation styles and sparsity parameters (Section~\ref{sec:results_weak_annotations});  %4) proposal (Section~\ref{sec:variations}) and analysis (Section~\ref{sec:results_variations}) of strategies for improving the performance of existing meta-learners.

This work is organized as follows. Section~\ref{sec:related} reviews the literature and discusses the taxonomy adopted in this paper to compare the different approaches. Section~\ref{sec:meta_learning} describes the assessed meta-learning methods while the experimental setup is explained in Section~\ref{sec:setup}. The experimental results are presented and discussed in Section~\ref{sec:results}, with conclusions presented in Section~\ref{sec:conclusion}.

\section{Related Work}
\label{sec:related}

\subsection{Meta-Learning for Visual Recognition}
\label{sec:related_metalearning}

Meta-Learning methods leverage multi-task learning in order to improve generalization for OOD tasks in image classification \cite{hospedales2021meta}. The idea is that, by generalizing for multiple datasets and tasks in the meta-training phase, the model is more well-equipped to deal with fully novel unseen tasks in the deployment phase.
We highlight three distinct approaches for achieving Meta-Learning that are important to this paper, despite other paradigms (such as black-box or Bayesian approaches) also being common: %1) weight prediction methods, often based on recurrent networks \cite{santoro2016meta,mishra2017simple} capable of producing weights of a fully novel network capable of few-shot learning;
1) gradient-based -- or optimization-based -- methods, which acquire task-specific parameters via optimization of first- \cite{finn2017model,nichol2018reptile} or second-order derivatives \cite{finn2017model,li2017metasgd,raghu2019rapid}; 2) metric learning \cite{snell2017prototypical,wang2019panet}, wherein instead of directly predicting class probabilities, methods focus on learning distances across samples from similar and dissimilar classes; 3) fusion-based approaches, which leverage the intermediary representations of the support set to guide the predictions in the query set \cite{lee2019meta,bertinetto2018meta,rakelly2018few}
%\cite{hu2019attention,zhang2019canet,zhang2020sg,bertinetto2018meta,rakelly2018few}
via identity mapping (i.e. concatenation, multiplication, addition, etc.).

\textbf{Gradient-based methods} yield -- often through second-order optimization -- specific models for each task $\T$ in a meta-batch.
% , as depicted for the digit classification scheme in Figure~\ref{fig:metalearners}.
Each subtask $\T$  computes its own temporary parameters, optimized using the task loss $\Ls_{\T}$. Finn \textit{et al.}~\cite{finn2017model} introduce the MAML algorithm, a second-order framework, which updates its model parameters by taking averages of cost's gradients of the specific task models evaluated on new samples of the task $\T$. Nichol \textit{et al.} \cite{nichol2018reptile} propose the Reptile algorithm, which only uses first-order gradient information by updating its weights in the direction of the difference between task specific parameters and global parameters. Also optimized via second-order derivatives, MetaSGD~\cite{li2017metasgd} aims to automatically learn step sizes for the SGD optimizer in addition to the model parameters. Another second-order algorithm similar to the MAML was introduced by Raghu \textit{et al.} \cite{raghu2019rapid}. Named ANIL, the method follows the MAML algorithm, with the novelty that in the \textit{inner} loop, instead of updating all the parameters, this strategy only updates the ones related to the network output head, i.e. the last classification layers.

% While first-order approaches uses approximations of gradients to update the $\phi$ model, such as \cite{nichol2018reptile}, where it is updated by moving the weights in the direction of the difference between $\phi_{\T}$ and $\phi$, in second-order approaches, such as \cite{finn2017model}, the model is updated by taking averages of gradients of the cost the specific $\phi_{\T}$ models evaluated in new samples of the task $\T$. 

\textbf{Metric-based approaches} train a single model $\phi$ in multiple tasks $\T$. The objective of these approaches is to obtain an agnostic mapper to an embedding space where similar samples are closer than dissimilar ones.
Snell \textit{et al.} \cite{snell2017prototypical} propose the Prototypical Networks (ProtoNets), a model that tries to learn an embedding function that computes \textit{prototypes} to a class (e.g. a $d$-dimensional vector that represents a class) and uses the distance to these prototypes for inference. For each class, features extracted from samples of their support -- the labeled set of images of a task -- are averaged to create its prototype vector. During inference, the feature extracted from a query image is compared with the prototypes, and the class of the closest prototype, according to some distance metric, is assigned to the query.

\textbf{Fusion-based approaches}, on the other hand, learn a single model $\phi$ where information of support sets is used to enhance the prediction of the query images. Similarly to metric-based methods, fusion-based approaches for meta-learning rely on an internal embedding from a neural network. However, instead of computing cross-sample similarities through a distance function (e.g. Euclidean, cosine, etc.) on the embeddings, such methods perform some form of late fusion (i.e. concatenation, addition, multiplication, cross-attention, etc.) on the support embeddings/labels and query samples. 
%This fusions can be used to ``guide'' inference from a small amount of annotated data \cite{bertinetto2018meta,lee2019meta}.
The Ridge Regression Differentiable Discriminator (R2D2) \cite{bertinetto2018meta} uses the support embeddings and labels to train a fully tensorial logistic or ridge regressor using least-squares, which admits closed-form solutions. Similarly, MetaOptNet \cite{lee2019meta} leverages a highly discriminative embedding generated from a neural network to train a differentiable SVM. In both methods, the regressor obtained from the support data is then applied to the query samples through a simple matrix-matrix product, resulting in a few-shot classifier guided by the few labeled support examples.

\subsection{Meta-Learning for Image Segmentation}
\label{sec:related_meta_segmentation}

% Given the success of meta-learning approaches in OOD classification problems, it was a natural progression to start seeing such approaches to semantic segmentation problems. However, perharps due to the intrinsic difficult of said problem and the expected structured output, meta-learning approaches are still underexplored for this scenario. Furthermore, meta-learning solutions for OOD semantic segmentation problems with weak supervision are still uncommon.

The most successful methods for FWS are Guided Nets \cite{rakelly2018few} and PANets~\cite{wang2019panet}. Guided Nets rely on pretrained backbones to extract features from both support and query data, and apply late fusion in these embeddings to guide the prediction over the query set from the support codes. By contrast, PANets rely on a framework similar to ProtoNets \cite{snell2017prototypical} to compute prototypes for each class in the embedding space of the support set instead of leveraging late fusion to guide the prediction over the query. PANets also introduce Prototype Alignment Regularization (PAR) in the training phase for better label efficiency, wherein the query labels are also used to compute prototypes to predict the segmentations of support samples. Similarly to PANets, ProtoSeg \cite{gama2022weakly} also use prototypes for conducting FWS, but instead of repurposing pretrained backbones, this approach is trained directly on related tasks from scratch in order to allow for inference over non-RGB images.% (i.e. medical imaging and remote sensing).
%Guided Nets are fusion-based methods that use a single backbone network to extract features from support and query images. Information in support images is fused with a simple operation, where first the features are averaged, then used as scaling factors to the query features. In contrast, PANets are metric-based approaches for the same problem, which also use a single backbone to extract features of both sets of images. PANets use a prototype approach akin to the Prototypical Networks \cite{snell2017prototypical}, one prototype being computed for each class using the features extracted from support images. Their novelty is the inclusion of the Prototype Alignment Regularization (PAR) during training, wherein the query labels are also used to compute prototypes to predict the segmentations of support samples. One key common aspect of both methods -- Guided Nets and PANets -- is the strong reliance on a large VGG backbone \cite{simonyan2014very} pretrained on the ImageNet dataset \cite{deng2009imagenet}. This hampers their effectiveness in domains that differ from usual RGB images while limiting their direct use to target domains wherein there is enough labeled data for strong pretraining. %without heavy training with a large amount of data in said domains, which can make their use unfeasible. 

Hendryx \textit{et al.} \cite{hendryx2019meta} adapt the Reptile \cite{nichol2018reptile} and First-Order-MAML (FOMAML) \cite{finn2017model} to the problem of semantic segmentation. Their main contribution, however, is the introduction of the EfficientLab architecture, which is a convolutional network for semantic segmentation. Weakly-supervised Segmentation Learning (WeaSeL) \cite{gama2021learning} applies the well-known MAML second-order optimization-based framework \cite{finn2017model} to FWS in medical imaging by training it directly on tasks related to radiology, with the downside of being less efficient than first-order approaches.
%It uses the EfficientNet \cite{tan2019efficientnet} network for encoding the images and a new atrous spatial pyramid pooling (ASPP) to upscale the feature maps while having skip connections at multiple resolutions.
% 
% Gama \textit{et al.} \cite{gama2021learning} introduce WeaSeL for few-shot semantic segmentation with sparse annotations. This gradient-based approach uses two nested loops. In the inner loop when optimizing for sampled tasks simulated sparse annotations are used. the dense annotations are used in order to compute the outer loss function. The learner can thus be trained in a scenario similar to the few-shot target task to extract information from the weak supervision in the inner loop. The authors evaluated the WeaSeL method on lung segmentation datasets.
% In a subsequent work, Gama \textit{et al.} \cite{gama2022weakly} also introduced the ProtoSeg method. It is a metric-based technique that uses class prototypes. The method learns an embedding function that maps each pixel of the image to a feature vector. The class prototypes are the averages of the feature vectors of the pixels belonging to that class. The authors evaluated the methods using a set of medical segmentation datasets as well ass some remote sensing tasks.

Even though multiple works on meta-learners specifically designed for segmentation have appeared during the last years, there is still a gap in such methods that can work with weakly-annotated images. Aiming at encouraging a larger use of Meta-Learning for FWS tasks, in the remainder sections of this text we propose generalizable pipelines for porting meta-learners designed for image classification to weakly-supervised segmentation tasks, and test these approaches in real-world medical tasks.

\section{Meta-Learners for Weakly Supervised Segmentation}
\label{sec:meta_learning}

We use most of the problem definitions from Gama \textit{et al.} \cite{gama2021learning}. We consider a training dataset $\D$ as a set of pairs $(\mathbf{x}, \mathbf{y})$ of images $\mathbf{x} \in \R^{H \times W \times B}$ with dimensions $H \times W$ and $B$ bands/channels, and semantic labels $\mathbf{y} \in \R^{H\times W}$. For each batch fed to an algorithm there are two partitions, named the \textit{support} set ($\D^{sup}$) and the \textit{query} set ($\D^{qry}$), such that $\D^{sup} \cap \D^{qry} = \emptyset$. We define a segmentation task $\T$ as a tuple $\T = \{\D^{sup}, \D^{qry}, t\}$ (or, $\T = \{\D, t\}$), where $t$ is a target class or set of classes. In our setting, all segmentation tasks are binary during both meta-training and deployment, so $t$ is a single class that can be referred to as \textit{positive/foreground} in opposition to the \textit{negative/background} class. $\D^{qry} = \{ \mathbf{x}^{qry}, \mathbf{y}^{qry} \}$ is composed of a set of images ($\mathbf{x}^{qry}$) and associated densely labeled segmentation ground truth ($\mathbf{y}^{qry}$), while the support set $\D^{sup} = \{ \mathbf{x}^{sup}, \mathbf{y}^{sup} \}$ contains another subset of images from the same dataset $\D$ as $\D^{qry}$, but paired with a weakly-supervised mask $\mathbf{y}^{sup}$. We employ several distinct strategies detailed in the supplementary materials of this manuscript to procedurally acquire the weak labels $\mathbf{y}^{sup}$ from the dense segmentation masks of the meta-training datasets.

A few-shot semantic segmentation task $\F$ is a specific type of segmentation task. The difference is that the samples of $\D^{sup}$ have their labels sparsely annotated, and the labels in $\D^{qry}$ are absent or unknown during training/tuning. Moreover, the number of samples $k = |\D^{sup}|$ is small, e.g., $20$, $10$ or even less. %We can also refer to a few-shot task as a $k$-shot task, where $k$ is the number of support samples.

The problem is then defined as follows. Given a few-shot task $\F$ and a set of segmentation tasks $\mathbfcal{T} = \{ \T_1, \T_2, \dots, \T_n \}$, we want to segment the images from $\D^{qry}_{\F}$ using information from tasks in $\mathbfcal{T}$ and information from $\D^{sup}_{\F}$.
We also require that no pair of image/labels of $\F$ is present in $\mathbfcal{T}$, in order to assure that the only semantic information about $\F$ is in its support annotations.

In order to teach the model through supervised inputs during the meta-training phase, we employ supervised loss functions $\Ls_{sup}$. As not all pixels in an FWS task are labeled, we leverage the pixelwise Selective Cross-Entropy (SCE) loss function~\cite{gama2021learning,gama2022weakly} in its binary form to conduct our supervised training on the labeled pixels in a ground truth $\mathbf{y}$ and prediction logits $\hat{\mathbf{y}}$:
\begin{equation}
    \Ls_{sce}(\mathbf{y}, \hat{\mathbf{y}}) = - \frac{1}{N} \sum_{j=1}^{N} \mathds{1}_{j} \left[ \mathbf{y}_{j} \log \hat{\mathbf{y}}_{j} + (1 - \mathbf{y}_{j}) \log (1 - \hat{\mathbf{y}}_{j}) \right].
    \label{eq:sce}
\end{equation}
In Equation~\ref{eq:sce}, $N$ is the number of labeled pixels in $\mathbf{y}$, $j$ is an index iterating over all pixels, $\hat{\mathbf{y}}_j$ is the probability predicted for each pixel $j$ of being classified as pertaining from the positive class, and $\mathds{1}_j \in \{0, 1\}$ is a flag indicating whether a pixel $j$ has a valid annotation or not. $\Ls_{sce}$ is applicable to gradient-, metric- and fusion-based methods, with only the form of computing the logits $\hat{\mathbf{y}}_{j}$ varying across the different paradigms. We employ the SCE loss function in our experiments for all methods, as we observed early on that other supervised segmentation loss functions (i.e. Dice \cite{milletari2016v}, Focal \cite{lin2017focal}, etc.) did not achieve the same performance as SCE.

% Apart from the $\Ls_{sce}$, we also conducted exploratory experiments with selective versions of the Dice Similarity Coefficient (DSC) \cite{milletari2016v} and Focal \cite{lin2017focal} losses, also very common choices for segmentation tasks. %These losses proved to be more unstable in our framework and, thus, were eliminated from the pipeline in its current form.

Our proposed pipeline for FWS via Meta-Learning was conceived as a general strategy to convert algorithms originally designed for image classification to semantic segmentation, in theory being applicable to any gradient-, metric- or fusion-based method. We leverage knowledge gathered from previous works~\cite{gama2021learning,gama2022weakly} on FWS using MAML \cite{finn2017model} and ProtoNets \cite{snell2017prototypical} to generalize the frameworks to other state-of-the-art Meta-Learning algorithms \cite{li2017metasgd,bertinetto2018meta,raghu2019rapid,lee2019meta,nichol2018reptile}. Sections~\ref{sec:gradient_fws},~\ref{sec:metric_fws} and~\ref{sec:fusion_fws} discuss how each Meta-Learning paradigm can be ported to FWS tasks.

\subsection{Gradient-based Meta-Learning for FWS}
\label{sec:gradient_fws}

In order to adapt gradient-based meta-learners to FWS we employ FCNs and Encoder-Decoder architectures, and we divide each network architecture into two distinct parts: 1) a feature extraction component $\phi$; and 2) a segmentation head $h$. $\phi$ receives support images $\mathbf{x}^{sup}$ and outputs embedded representations of the pixels in these images $\mathbf{f}^{sup} = \phi(\mathbf{x}^{sup})$, while $h$ inputs $\mathbf{f}^{sup}$ and outputs segmentation predictions $\hat{\mathbf{y}}^{sup}$. For both FCNs and Encoder-Decoders, $h$ is simply the last convolutional block responsible for the final pixelwise classification, while the feature extractor $\phi$ comprises all previous layers in the architecture -- be it a sequence of Encoder/Decoder blocks or a CNN backbone.

The gradient-based pipeline can be observed in a graphical manner in Figure~\ref{fig:metaseg_gradient} for one single inner loop. As one can see, a support image $\mathbf{x}^{sup}$ is initially fed through $\phi$, generating features $\mathbf{f}^{sup}$, which are then fed to $h$, yielding a segmentation prediction $\hat{\mathbf{y}}^{sup}$. The prediction is then compared to the sparsely supervised ground truth $\mathbf{y}^{sup}$, from the support set, in the few labeled points available through $\Ls_{sup}$. The inner loop update function $U_{in}$ operates on the gradients obtained through first- \cite{nichol2018reptile} or second-order \cite{finn2017model,li2017metasgd} optimization, depending on the meta-learning algorithm of choice. $U_{in}$ returns the task-specific feature extractor $\phi_{\star}$ and segmentation head $h_{\star}$, which are then fed with the query image $\mathbf{x}^{qry}$, resulting in a prediction $\hat{\mathbf{y}}^{qry}$ for the query set. $\hat{\mathbf{y}}^{qry}$ and $\mathbf{y}^{qry}$ are then compared through $\Ls_{sup}$, yielding gradients that can be backpropagated to meta-models $\phi$ and $h$.

\renewcommand{\currprop}{0.4\textheight}
\renewcommand{\currprop}{0.415\textheight}

\begin{figure}[!ht]
    \centering
    \includegraphics[height=\currprop]{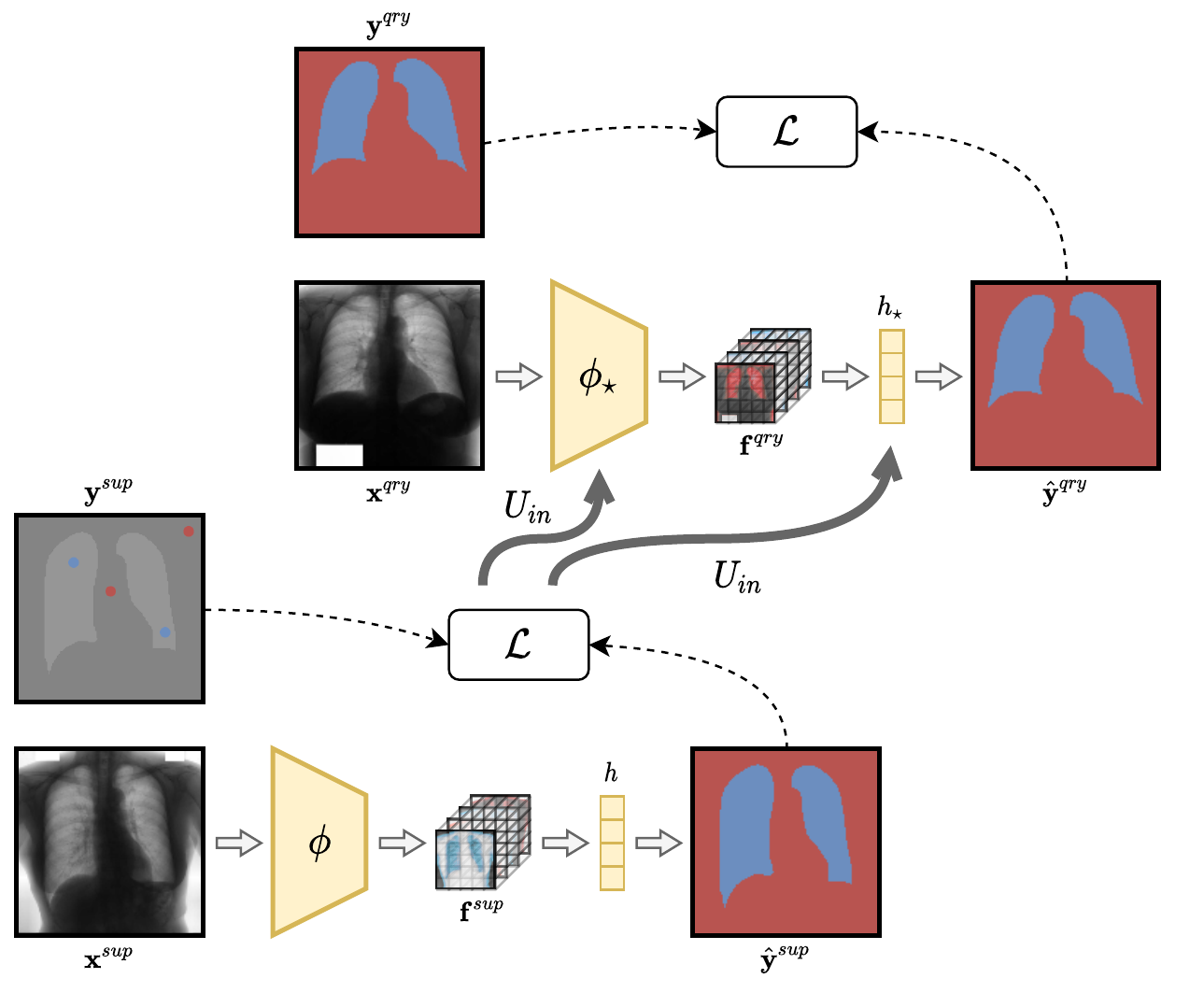}
    \caption{Graphical illustration of one inner loop iteration for gradient-based FWS methods.}
    \label{fig:metaseg_gradient}
\end{figure}

In a real-world implementation of this idea, this procedure is repeated for all tasks randomly sampled in a meta-batch, each one with distinct $\phi_{\star}$ and $h_{\star}$ parameter sets, and yielding different gradients to be backpropagated to $\phi$ and $h$. The gradients can then be merged -- usually via averaging -- to update $\phi$ and $h$ to more generalist parameter sets, highly adaptable to multiple tasks at once. This update to $\phi$ and $h$ from the gradients obtained on the query set from the task-specific models is the outer loop of the optimization-based meta-learning algorithm. Additionally, it is highly desirable that the support image -- or, realistically, support batch of images -- is fed multiple times through $\phi$ and $h$, with the gradients being accumulated to generate $\phi_{\star}$ and $h_{\star}$ before feeding the query set to the task-specific models. The number of iterations through the support set used to compute $\phi_{\star}$ and $h_{\star}$ is a hyperparameter of gradient-based meta-learners, often being limited by the amount of memory in the GPU and/or training time constraints.

% The optimization-based algorithms adapted in our work were MAML \cite{finn2017model}, MetaSGD \cite{li2017metasgd}, ANIL \cite{raghu2019rapid}, a version of MetaSGD with second-order gradients being applied only to the optimization of task specific heads named MetaANIL,
% and Reptile \cite{nichol2018reptile}.

\subsection{Metric-based Meta-Learning for FWS}
\label{sec:metric_fws}

\begin{figure}[!ht]
    \centering
    \includegraphics[height=\currprop]{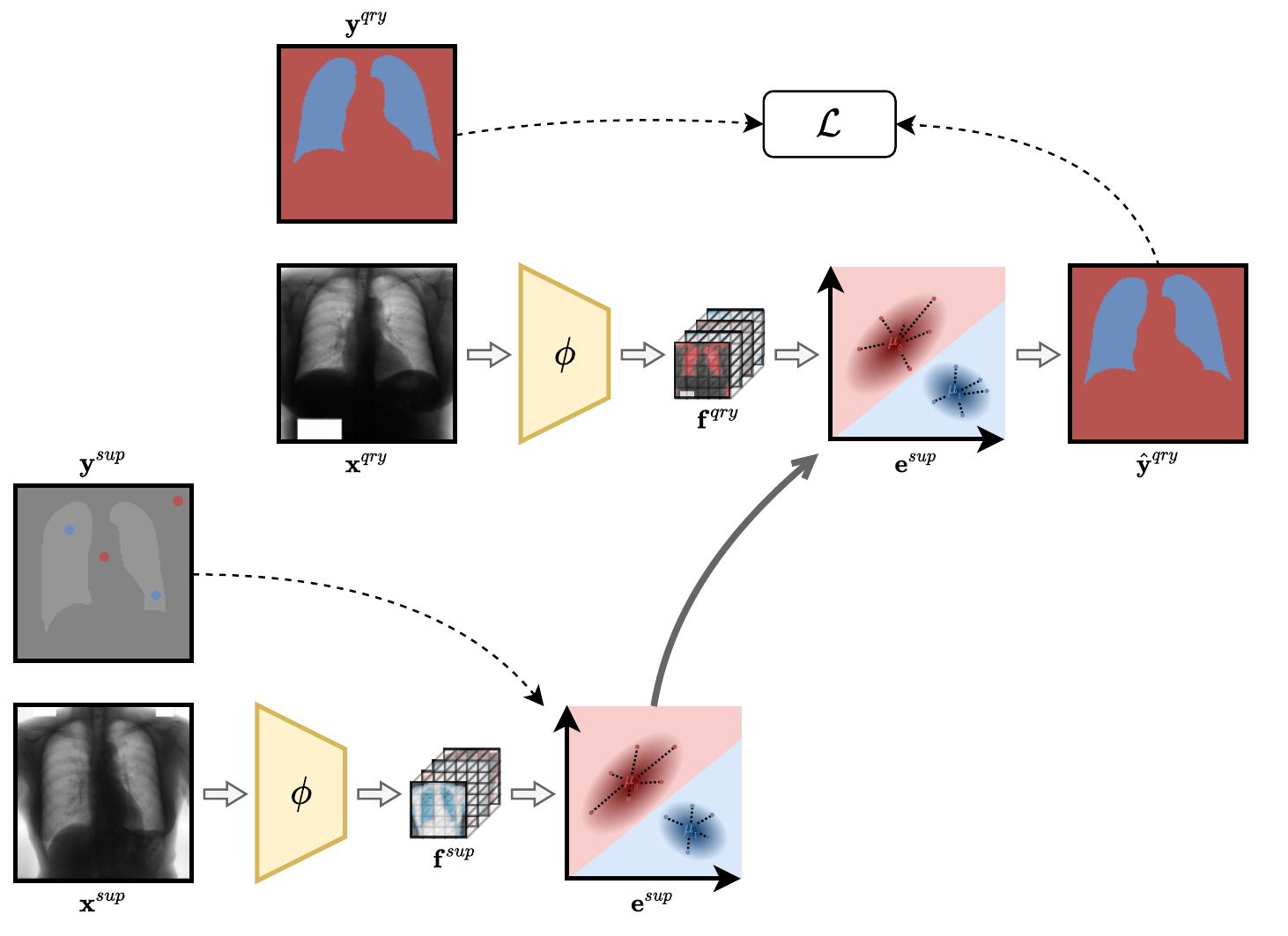}
    \caption{Graphical illustration of one inner loop iteration for metric-based FWS methods.}
    \label{fig:metaseg_metric}
\end{figure}

As discussed in Section~\ref{sec:related}, metric-learning can also be used to achieve meta-learning through multi-task meta-training and some clever use of support set annotations \cite{snell2017prototypical,wang2019panet,gama2022weakly} that are leveraged distinctly from an optimization-based approach. Instead of using the support set to tune task-specific models that should perform well on the query, metric-based methods instead use the labels from the support to compute prototypes in the embedding space, which can then be used as pivots to compute distances to query samples. This process is similar to a k-Nearest Neighbors or a Nearest Class Mean \cite{mensink2013distance} algorithm operating on the embedding space. This allows for extremely low-shot learning regimes (i.e. one-shot) and even zero-shot learning, the latter being unfeasible with most gradient-based methods.

We adapted metric-based meta-learners for FWS following the pipeline presented in Figure~\ref{fig:metaseg_metric} for one single inner loop iteration. Differently from most gradient-based approaches, the feature extractor $\phi$ remains frozen during the inner loops. The embedded spaces for the support $\mathbf{f}^{sup}$ and query $\mathbf{f}^{qry}$ sets, both in $\R^{C \times H \times W}$, are fundamental to pixelwise classification in this approach, assuming $C$ output channels to $\phi$. The few support labels available are used to compute pixel prototypes for each class, similarly to Snell \textit{et al.} \cite{snell2017prototypical}, which do this at an image level. As all tasks are binary in our approach, two centroids $\mu_{0}, \mu_{1} \in \R^{C}$ are computed for the negative and positive classes, respectively; resulting in an embedded representation $\mathbf{e}^{sup}$. Prototypes only take into account the labeled pixels from $\mathbf{y}^{sup}$, ignoring the unannotated ones from $\mathbf{f}^{sup}$.

The query set features $\mathbf{f}^{qry} = \phi(\mathbf{x}^{qry})$ are then projected onto the space $\mathbf{e}^{sup}$, where the distances of each query pixelwise feature vector can be computed in relation to $\mu_{0}$ and $\mu_{1}$ according to some distance metric $d$ (e.g. Euclidean \cite{snell2017prototypical,gama2022weakly}, cosine \cite{wang2019panet}, Mahalanobis, Manhattan, etc.). Logits can then be computed for each query pixel according to their distances to the centroids of the negative and positive classes, allowing them to be fed to a supervised loss function $\Ls_{sup}$. The gradients obtained from $\Ls_{sup}(\mathbf{y}^{qry}, \hat{\mathbf{y}}^{qry})$ can then be backpropagated through the pipeline, reaching the trainable parameters $\phi$. Similarly to the gradient-based methods, multiple inner loops such as the one shown in Figure~\ref{fig:metaseg_metric} are conducted in each meta-training iteration, with the gradients $\nabla_{\phi}\Ls_{sup}$ being added or averaged before updating $\phi$ on the outer loop.

% The main distinction across similarity-based meta-learners regards the distance $d$ chosen for computing query logits in comparison to the support centroids and loss regularizations that improve label efficiency \cite{wang2019panet}. For this work, we chose two metric-based meta-learners for FWS: 1) ProtoNets \cite{snell2017prototypical} and PANets \cite{wang2019panet}.

\subsection{Fusion-based Meta-Learning for FWS}
\label{sec:fusion_fws}

\begin{figure}[!ht]
    \centering
    \includegraphics[height=\currprop]{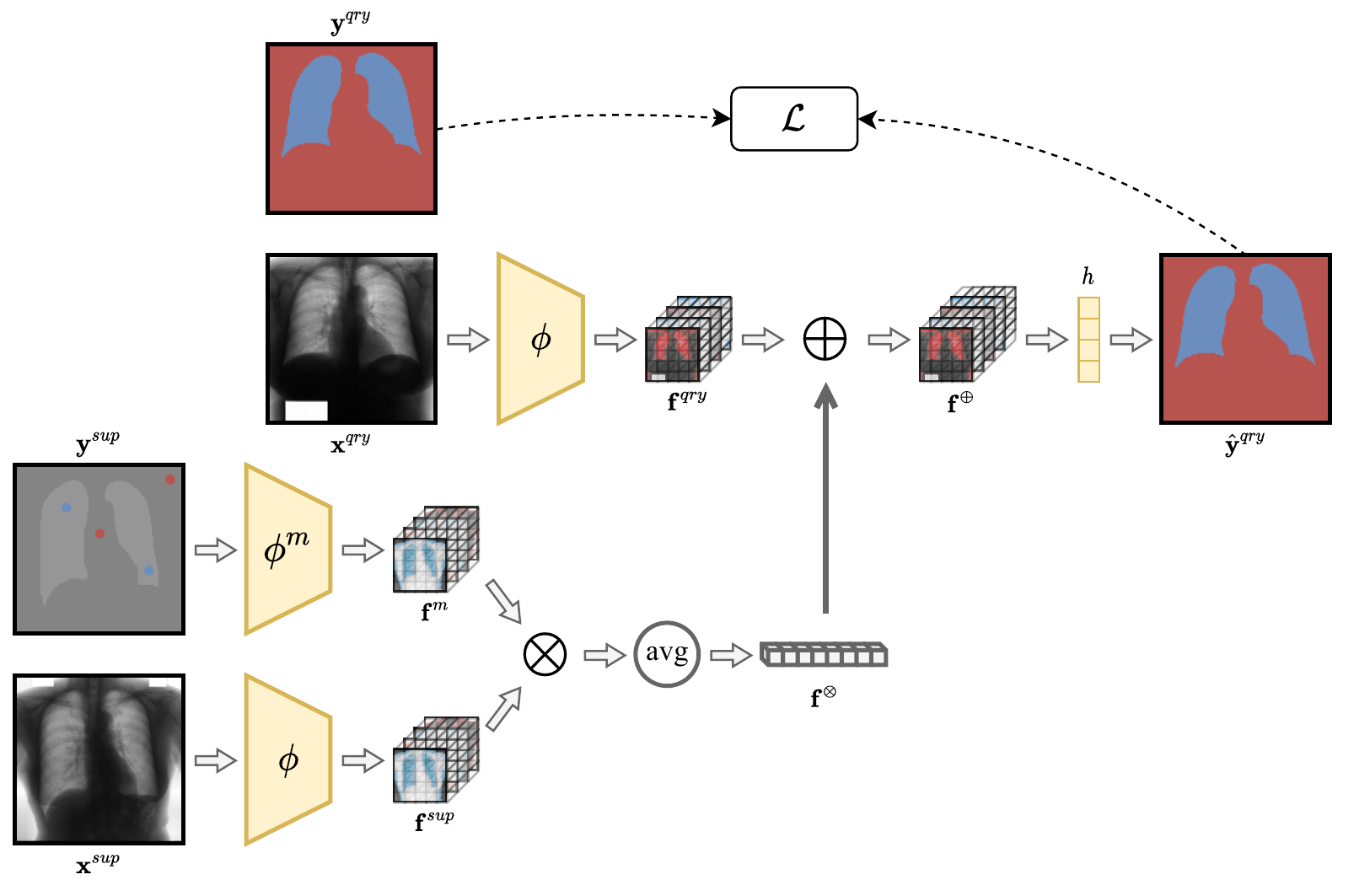}
    \caption{Graphical illustration of one inner loop iteration for fusion-based FWS methods.}
    \label{fig:metaseg_fusion}
\end{figure}

Fusion-based methods work by using the support set images and labels to ``guide'' \cite{zhang2020sg,rakelly2018few} the classification or segmentation of the desired classes on the query set samples via late feature fusion between feature representations $\mathbf{f}^{sup}$ and $\mathbf{f}^{qry}$. Figure~\ref{fig:metaseg_fusion} exemplifies the pipeline of a fusion-based algorithm along one inner loop, starting with the feature extraction from the support and query images via $\phi$ -- yielding $\mathbf{f}^{sup}$ and $\mathbf{f}^{qry}$ -- and the extraction of features from the sparse support mask $\mathbf{y}^{sup}$, resulting in $\mathbf{f}^{m}$. Feature representations $\mathbf{f}^{m}$ and $\mathbf{f}^{sup}$ are then fused using a function $\otimes$ and averaged into 1D vector $\mathbf{f}^{\otimes}$, responsible for guiding $\phi$ in segmenting query images according to the background and foreground classes in $\mathbf{y}^{sup}$. The guiding vector is then fused to the query features $\mathbf{f}^{qry}$ using another mapping $\oplus$ and resulting in $\mathbf{f}^{\oplus}$. The fused feature maps $\mathbf{f}^{\oplus}$ can then be passed through a segmentation head $h$, resulting in predictions $\hat{\mathbf{y}}^{qry}$, which is subsequently fed to $\Ls_{sup}$.

As shown in Figure~\ref{fig:metaseg_fusion}, some fusion-based approaches also extract features from the sparsely labeled support segmentation mask $\mathbf{y}^{sup}$ through a generic model referred to as $\phi^{m}$. Guided Nets \cite{rakelly2018few}, for instance, repurpose the backbone~$\phi$ to share parameters with $\phi^{m}$, even though we found that keeping distinct parameters sets for $\phi$ and $\phi^{m}$ resulted in much more stable pretraining using this strategy. %However, learning features from $\mathbf{f}^{sup}$ is not a constant to all fusion-based methods, and other strategies only use the masks to change the distribution of support image features \cite{zhang2020sg} or leverage the support labels to train simple differentiable classifiers (e.g. ridge/logistic regression \cite{bertinetto2018meta} or support vector machine \cite{lee2019meta}).
Another important aspect of fusion-based methods is the choice of the two functions: $\otimes$ -- that merges support image and label features; and $\oplus$ -- responsible for fusing the support and query feature representations. Common choices for these functions are concatenation, matrix multiplication, addition, pixelwise multiplication or even trainable attention modules \cite{oliveira2022domain}.% \cite{zhang2019pyramid}. Three distinct fusion-based meta-learners were tested in our FWS experiments: Guided Nets \cite{rakelly2018few}, MetaOptNet \cite{lee2019meta}, R2D2 \cite{bertinetto2018meta}.

% \section{Variations to Meta-Learning Paradigms}
% \label{sec:variations}

% \todo{Mention variations of existing algorithms (i.e. attention \cite{schlemper2018attention,jetley2018learn}, CRF \cite{krahenbuhl2011efficient,zheng2015conditional}, positional encoding \cite{xu2021positional,chu2021conditional} implemented with simple concatenation and using FiLM \cite{perez2018film}, deep supervision \cite{huang2020unet}, non-local operations \cite{wang2018non}, etc).}

\section{Experimental Setup}
\label{sec:setup}

Our experiments compare both well-known FWS algorithms in the literature (PANets \cite{wang2019panet}, Guided Nets \cite{rakelly2018few}, WeaSeL \cite{gama2021learning} -- referred to as MAML and ProtoSeg \cite{gama2022weakly} -- referred to as ProtoNets) and novel algorithms based on meta-learners designed for few-shot classification and ported to FWS (MetaSGD \cite{li2017metasgd}, ANIL \cite{raghu2019rapid}, Reptile \cite{nichol2018reptile}, R2D2 \cite{bertinetto2018meta} and MetaOptNet \cite{lee2019meta}).

We explored different neural network architectures as backbones for the meta-learners, including: U-Net (\textbf{U}) \cite{gama2022weakly,ronneberger2015u}, EfficientLab-6-3 (\textbf{E}) \cite{hendryx2019meta}, DeepLabv3 (\textbf{D}) \cite{chen2017rethinking} and an FCN with a ResNet-12 (\textbf{R}) \cite{lee2019meta} backbone. %The selection of architectures was aimed at achieving diversity by including Encoder-Decoders \cite{ronneberger2015u}, FCNs with state-of-the-art CNN backbones, and other architectures specifically designed to work on meta-learning algorithms \cite{vinyals2016matching,lee2019meta,hendryx2019meta,ye2020few}. However, most of these architectures proved to yield redundant results, while others had convergence problems or simply underfit the meta-dataset, which led us to select only the U-Net (\textbf{U}), ResNet-12 (\textbf{R}), EfficientLab (\textbf{E}) and DeepLabv3 (\textbf{D}) architectures for our full experimental evaluation.
% We explored different neural network architectures as backbones for the meta-learners, including: U-Net \cite{gama2022weakly,ronneberger2015u}, EfficientLab-6-3 \cite{hendryx2019meta,tan2021efficientnetv2}, DeepLabv3 \cite{chen2017rethinking} and FCNs with four distinct backbones based on the 4-layered CNN from Vinyals \textit{et al.} \cite{vinyals2016matching}, ResNet-12 \cite{lee2019meta,he2016deep} and Wide ResNet-28 (WRN28) \cite{ye2020few,zagoruyko2016wide}. The selection of architectures was aimed at achieving diversity by including Encoder-Decoders \cite{ronneberger2015u}, FCNs with state-of-the-art CNN backbones, and other architectures specifically designed to work on meta-learning algorithms \cite{vinyals2016matching,lee2019meta,hendryx2019meta,ye2020few}. However, most of these architectures proved to yield redundant results, while others had convergence problems or simply underfit the meta-dataset, which led us to select only the U-Net (\textbf{U}), ResNet-12 (\textbf{R}), EfficientLab (\textbf{E}) and DeepLabv3 (\textbf{D}) architectures for our full experimental evaluation.

As the performance of few-shot algorithms are notoriously variable according to the chosen support set samples, all results shown in Section~\ref{sec:results} were computed according to a 5-fold cross-validation procedure, with paired samples for each fold in all algorithms. As all of our tasks are binary by experimental design, the simple Intersection over Union (IoU, also known as Jaccard index) between the positive and negative classes was our main measure for assessing the performance of FWS meta-learners.

Our implementation of meta-learners for FWS was coded using the Pytorch\footnotemark\footnotetext{\url{https://pytorch.org/}} and learn2learn\footnotemark\footnotetext{\url{http://learn2learn.net/}} libraries. Experiments were conducted in machines running RTX 2070 GPUs with ~8GB of memory, so many hyperparameters for the meta-learning algorithms (i.e. meta batch size, number of inner loops, number of filters in the architectures, adaptation steps in 2$^{nd}$ order gradient-based methods, etc.) were chosen aiming to fit this capacity. Whenever possible, we set the default optimizer for our algorithms as Adam \cite{kingma2014adam}, unless in methods that use quadratic program solvers, such as MetaOptNet \cite{lee2019meta} or R2D2 \cite{bertinetto2018meta}, which rely on standard Stochastic Gradient Descent (SGD). As time limitation was our main consideration in designing the experiments shown in this work, we leveraged knowledge from previous works \cite{gama2021learning,gama2022weakly} to set most of the other hyperparameters that do not directly affect the memory usage of meta-learners. Readers can refer to our official implementation\footnotemark\footnotetext{\url{https://github.com/hugo-oliveira/fsws_metalearning}} for details.

\subsection{Datasets and Preprocessing}
\label{sec:datasets}

\noindent\textbf{2D Data:} As our methods were adapted solely to 2D data, we use multiple 2D radiological datasets with pixelwise annotations for the meta-training phase of the meta-learners. We follow the experimental setup of Gama \textit{et al.} \cite{gama2021learning,gama2022weakly}, borrowing their radiology datasets for the meta-training and test phases. Sets of Chest X-Rays (CXRs)%\cite{JSRTshiraishi2000development,jaeger2014two,NIHwang2017chestx,NIHtang2019xlsor}
, Mammographic X-Rays (MXRs)%\cite{moreira2012inbreast,MIASsuckling1994mammographic}
, Dental X-Rays (DXRs) %\cite{IVISIONsilva2018automatic,PANORAMICabdi2015automatic}
and Digitally Reconstructed Radiographs (DRRs) %\cite{LIDCarmato2011lung,LIDColiveira20203d}
are used as source 2D datasets. The CXR datasets contain lung, heart and clavicle annotations, while MXRs are labeled for breast region and pectoral muscle. DXR images contain reference segmentations for teeth and lower mandible, and DRRs are labeled for ribs. Not all datasets in each imaging modality are labeled for all organs, as we used solely public organ pixelwise labels for each dataset. %Some of the 2D source datasets were also used as target datasets by being excluded from the meta-training phase when they were selected as the target domains. %The 2D dataset/task pairs used as target were: 1) JSRT \cite{JSRTshiraishi2000development} on the \textit{Both Lungs} task; 2) OpenIST\footnote{\url{https://github.com/pi-null-mezon/OpenIST}} on the \textit{Both Lungs} task; 3) MIAS \cite{MIASsuckling1994mammographic} on the \textit{Pectoral Muscle} task; and 4) Panoramic Dental X-Rays \cite{PANORAMICabdi2015automatic} on the \textit{Lower Mandible} task.

\noindent\textbf{2D alices from 3D Data:} Aiming to properly test the label efficiency of the few-shot meta-learners on fully distinct domains in relation to the meta-training ones, we also acquired 2D slices from volumetric datasets (e.g. Computed Tomography -- CT; and Magnetic Resonance Imaging -- MRI) \cite{simpson2019large,antonelli2022medical,oliveira2021automatic}. In order to gather consistent 2D slices from the 3D datasets, we used the center slices of each target organ of interest in the craniocaudal axis of the patients. These data were fully absent from any meta-training conducted during this work, being only used as target domains/tasks.

% \Isa{MAYBE ADD FOOTNOTES WITH THE URL WHERE ALL THESE DATA CAN BE FOUND?}

\noindent\textbf{Preprocessing:} During Meta-Training we preprocess the support and query images with Contrast Limited Adaptive Histogram Equalization (CLAHE) \cite{pizer1987adaptive}, followed by random transformations (rotations, horizontal and vertical flips, etc.). At last, we resize the images to $140 \times 140$ pixels, finally performing a random crop to $128 \times 128$ pixels. We noticed that the random data augmentation operations used in our experiments only worked when they were consistent across whole batches. Aiming to standardize the experiments, during the deployment of our model on the target tasks we do not perform any random operation. The preprocessing in the deployment stage consists simply in applying CLAHE and resizing to $128 \times 128$ pixels.

% \todo{Mention that experiments were conducted with an official implementation in PyTorch, with the aid of the learn2learn library. All experiments were run on machines equiped with RTX 2070 GPUs -- which allows for runtime comparisons -- and hyperparameters (i.e. meta batch size, number of inner loops, adaptation steps, etc) were tuned in order to occupy at most the ~8GB of RAM in each GPU. Provide link to the official implementation and demo.}

% \todo{Mention that, as we could not run grid searches for each algorithm, other hyperparameters were chosen according to previous knowledge from \cite{gama2021learning,gama2022weakly,bertinetto2018meta,lee2019meta}. Mention the choice of optimizers for each algorithm: by default Adam, with exceptions to algorithms that did not behave well with Adam (i.e. MetaOptNet, R2D2, etc). In such cases, SGD was used. Mention the \textit{metalr} and \textit{fastlr} parameters.}

\subsection{Experiment Organization}
\label{sec:organization}

In order to test the efficacy of the meta-learners shown in Section~\ref{sec:meta_learning} for FWS, we designed an experimental setup aiming to assess the performance of each algorithm in multiple radiological image segmentation tasks. We use the \textit{points} weak annotation style in order to assess the performance of algorithms in very small data scenarios for two distinct few-shot tasks: $\F_{id}$ -- the OpenIST dataset\footnote{\url{https://github.com/pi-null-mezon/OpenIST}} in the task of lung segmentation (\textit{OpenIST-lungs}); and $\F_{ood}$ -- the Panoramic Dental X-Rays \cite{PANORAMICabdi2015automatic} in the task of inferior mandible segmentation (\textit{Panoramic-mandible}).

Task $\F_{id}$ represents a relatively ``in-distribution'' (ID) task very close to many domains used during meta-training, as both the Shenzhen/Mongomery sets %\cite{jaeger2014two}
and the Chest X-Ray 14 dataset %\cite{NIHtang2019xlsor}
are very similar to the OpenIST data and also contain lung annotations. Task $\F_{ood}$, however, is a very out-of-distribution task in comparison to the meta-training domains, as no other dataset contains segmentation for the inferior mandible and only one other DXR dataset (IVisionLab) %\cite{IVISIONsilva2018automatic}
is used during the meta-training for the quite distinct task of teeth segmentation. In order to avoid the meta-learners simply overfitting on the target dataset% \cite{rajendran2020meta}
, we hold out each target domain in an experiment from the meta-training phase, assuring that the first time the methods/networks have seen each target dataset is during the test phase.

Additionally to the experiments on ID and OOD 2D images, we tested our meta-learners pretrained on the 2D CXRs, MXRs and DXRs in four tasks from three volumetric datasets: 1) StructSeg \cite{simpson2019large} on the \textit{Both Lungs} and \textit{Heart} tasks (\textit{StructSeg-lungs} and \textit{StructSeg-heart}); 2) Medical Imaging Segmentation Decathlon (MSD) \cite{antonelli2022medical} slices on the \textit{Spleen} task (\textit{MSD-spleen}); and 3) private data from Oliveira \textit{et al.} \cite{oliveira2021automatic} in \textit{Cerebellum} segmentation on pediatric MRI data (\textit{STAP-cerebellum}). We selected these tasks as targets due to the relatively large size of these organs in the images in comparison to the whole volumes. This selection was conducted in order to avoid the inherent difficulties of compensating for domain shift in datasets with imbalanced target classes %\cite{hsu2015unsupervised,tsai2016domain,yan2017mind}.
\cite{hsu2015unsupervised}.

While the meta-training is conducted with the largest variability possible for the choice of support/query set samples and weak support masks, we fixed the seeds of all randomly chosen variables in the sparsification algorithms for the support set of the target task $\F$, forcing them to be exactly the same on a pixel-level for all samples.

% Another aspect that we chose to study was the choice of supervised loss function for the algorithms that allow variation in this aspect. The standard SCE \cite{gama2021learning,gama2022weakly} (Equation~\ref{eq:sce}) is a default choice in most supervised learning methods for classification/segmentation, while the Dice loss \cite{milletari2016v} is often paired with SCE in medical tasks. The Focal loss \cite{lin2017focal} is a segmentation- and detection-focused loss aimed at solving class imbalance, while the structural similarity index (SSIM) loss \cite{wang2003multiscale} has been adopted in many state-of-the-art medical image segmentation tasks during recent years \cite{huang2020unet}. All of these loss functions were tested in early experiments, however only the SCE loss yielded satisfactory results in our experiments. Thus, all experiments shown in Section~\ref{sec:results} are related to SCE experiments.

At last, in order to evaluate the performance of the best meta-learners in other weakly supervised segmentation styles, we conduct a series of experiments with \textit{grid}, \textit{scribbles}, and \textit{skeleton} annotations, as depicted in Figure~\ref{fig:weak_annotation_styles}. These experiments are presented in Section~\ref{sec:results_weak_annotations} for \textit{Panoramic-mandible}, \textit{OpenIST-lungs}, \textit{StructSeg-lungs}, \textit{StructSeg-heart}, \textit{MSD-spleen} and \textit{STAP-cerebellum}.

\renewcommand{\currprop}{\textwidth}

\begin{figure}[!ht]
    \centering
    \includegraphics[width=\currprop]{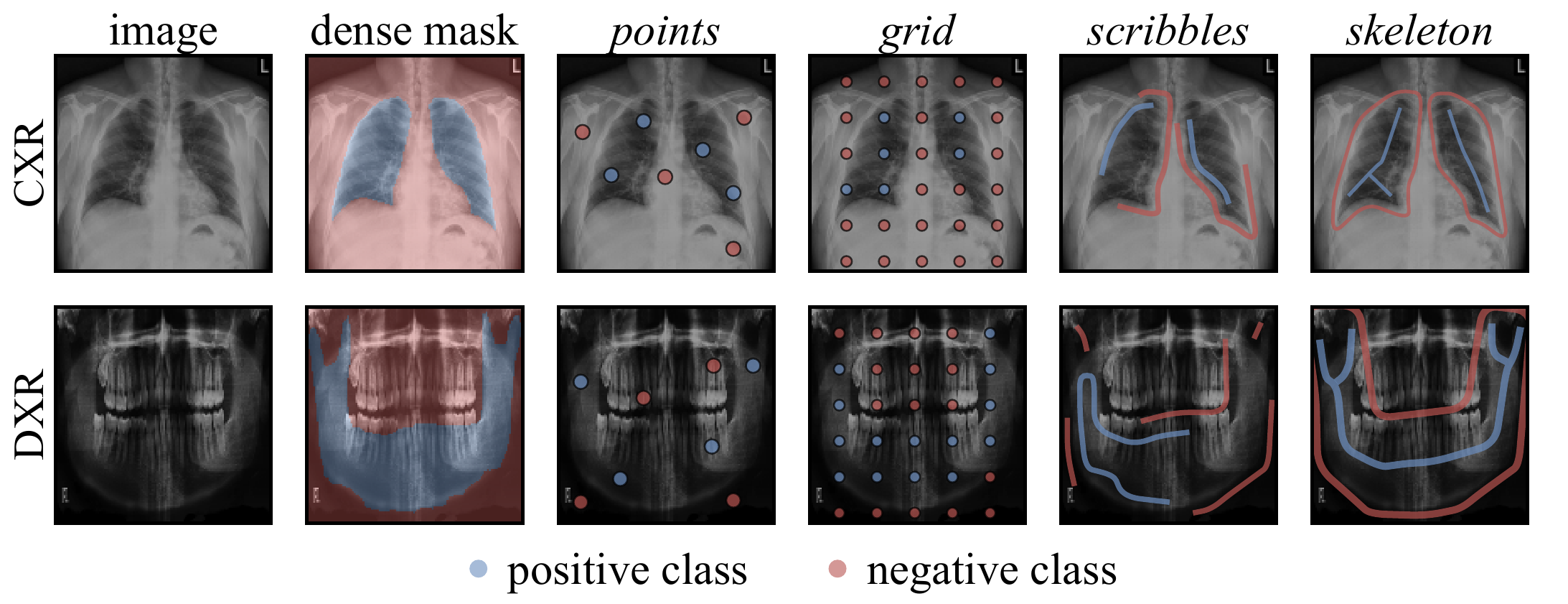}
    \caption{Weak annotation styles (4 last columns) studied in this work for two distinct radiological tasks: 1) \textit{OpenIST-lungs} (top row); and 2) \textit{Panoramic-mandible} (bottom row).}
    \label{fig:weak_annotation_styles}
\end{figure}

% \subsection{Annotation Style Experiments}
% \label{sec:weak_annotation_experiments}

% \todo{The best overall algorithms of each paradigm are compared in experiments concerning distinct annotation styles.}

\section{Results and Discussion}
\label{sec:results}

% \todo{Introduce this section and present the organization of following subsections...}

\subsection{Results Intra Meta-Learning Paradigms}
\label{sec:results_intra}

We present initially the results in the \textit{points} annotation style for 1- and 5-shot scenarios, each one with 4 distinct label configurations: 1) 1 annotated point (p=1) with a dilation radius of 1 (r=1); 2) p=1/r=3; 3) p=5/r=1; and 4) p=5/r=3. These experiments correspond to the \textit{Panoramic-mandible} task, where the domain shift is larger than \textit{OpenIST-lungs}, as the focus of this work is in Domain Generalization instead of simple cross-dataset transfer learning. This results in 8 distinct annotation settings, which is the number of result columns in Tables~\ref{tab:results_baseline},~\ref{tab:results_gradient},~\ref{tab:results_metric} and~\ref{tab:results_fusion}. %Throughout the presentation of our results, the 4 backbones analysed in this work are coded by the letters \textbf{U} (U-Net), \textbf{R} (FCN with ResNet-12), \textbf{E} (EfficientLab) and \textbf{D} (DeepLabv3).
Bold values in result tables represent the best overall results for each label configuration (column), while underlined values highlight the backbones with the best performance for a meta-learner. All metrics within 0.01 of IoU difference from the best result are highlighted in those tables in order to point not only to the best overall algorithm/backbone pairs but also other strategies with comparable performance.

First, in order to assess the best from scratch baseline backbone, we show the results for the 4 distinct segmentation architectures in Table~\ref{tab:results_baseline}. From these initial experiments, we observe that U-Nets and ResNet-12 achieve the best overall results in the few-shot experiments, motivating us to present only ResNet-12 as the main baseline with no pretraining for the remaining of our analysis.

\begin{table}[!ht]
    \centering
    \caption{IoU results for the query set predicted by the \textit{from scratch} baseline with four segmentation backbones (\textbf{U}, \textbf{R}, \textbf{E} and \textbf{D}) directly trained on the few-shot weakly-supervised support samples with the \textit{points} annotation style.}
    \label{tab:results_baseline}
    \begin{tabular}{@{}cccccccccccc@{}}
        \toprule
        \multicolumn{2}{c}{\multirow{2}{*}{\textbf{Method}}} &  & \multicolumn{4}{c}{\textbf{1-shot}}                                                                                                                                                                                                               &  & \multicolumn{4}{c}{\textbf{5-shot}}                                                                                                                                                                                                               \\ \cmidrule(lr){4-7} \cmidrule(l){9-12} 
        \multicolumn{2}{c}{}                                 &  & \textbf{\begin{tabular}[c]{@{}c@{}}p=1\\ r=1\end{tabular}} & \textbf{\begin{tabular}[c]{@{}c@{}}p=1\\ r=3\end{tabular}} & \textbf{\begin{tabular}[c]{@{}c@{}}p=5\\ r=1\end{tabular}} & \textbf{\begin{tabular}[c]{@{}c@{}}p=5\\ r=3\end{tabular}} &  & \textbf{\begin{tabular}[c]{@{}c@{}}p=1\\ r=1\end{tabular}} & \textbf{\begin{tabular}[c]{@{}c@{}}p=1\\ r=3\end{tabular}} & \textbf{\begin{tabular}[c]{@{}c@{}}p=5\\ r=1\end{tabular}} & \textbf{\begin{tabular}[c]{@{}c@{}}p=5\\ r=3\end{tabular}} \\ \cmidrule(r){1-2} \cmidrule(lr){4-7} \cmidrule(l){9-12} 
        \multirow{4}{*}{\textbf{Baseline}}    & \textbf{U}   &  & {\ul \textbf{.371}}                                        & {\ul \textbf{.386}}                                        & .444                                                       & .451                                                       &  & {\ul \textbf{.459}}                                        & {\ul \textbf{.445}}                                        & .530                                                       & .545                                                       \\
                                              & \textbf{R}   &  & .312                                                       & .311                                                       & {\ul \textbf{.478}}                                        & {\ul \textbf{.482}}                                        &  & .443                                                       & {\ul \textbf{.450}}                                        & {\ul \textbf{.556}}                                        & {\ul \textbf{.558}}                                        \\
                                              & \textbf{E}   &  & .093                                                       & .098                                                       & .114                                                       & .128                                                       &  & .202                                                       & .234                                                       & .160                                                       & .167                                                       \\
                                              & \textbf{D}   &  & .273                                                       & .276                                                       & .231                                                       & .219                                                       &  & .224                                                       & .186                                                       & .243                                                       & .225                                                       \\ \bottomrule
    \end{tabular}
\end{table}

We present results for algorithms in each Meta-Learning paradigm separately in order to sort the more label efficient algorithms for comparisons cross-paradigms. Tables~\ref{tab:results_gradient},~\ref{tab:results_metric} and~\ref{tab:results_fusion} show, respectively, the IoU metrics for the gradient-, metric- and fusion-based methods for the 4 backbones analysed in this work, as well as the baseline with ResNet-12.

Table~\ref{tab:results_gradient} shows IoU results for pure $2^{nd}$ order \cite{finn2017model,li2017metasgd}, pure $1^{st}$ order \cite{nichol2018reptile}, as well as optimization-based strategies that use both $2^{nd}$ and $1^{st}$ order gradients \cite{raghu2019rapid}. It is clear that ANIL-\textbf{D} achieved the best overall results in gradient-based methods, yielding the best performances in all FWS configurations. %In fact, the only other strategy that even reached close results to ANIL was MetaANIL; a very similar approach with MetaSGD being applied on the classifier instead of the traditional MAML shown in Raghu \textit{et al.} \cite{raghu2019rapid}.
Hence, the hybrid strategies of mixing $1^{st}$ order gradients to train the backbone and $2^{nd}$ order gradients to train the segmentation head were the most label efficient gradient-based methods by quite some margin. Most of this strategy's success can be attributed to the larger amount of adaptation steps that can be computed to the segmentation head (10 adaptation steps in ANIL) in comparison to applying $2^{nd}$ order gradients to the whole backbone (2 adaptation steps for MAML and MetaSGD). Thus, with additional GPU memory, one could theoretically improve the performance of full $2^{nd}$ order approaches in theory. However, we conducted early tests with up to 10 adaptation steps on MAML and MetaSGD and these approaches tended to apparently overfit on the few annotated pixels with more adaptation steps than 5, possibly due to the larger parameter capacity in relation to the amount of labeled data being fitted by the $2^{nd}$ order optimization in the whole backbone.

\begin{table}[!ht]
    \centering
    \caption{IoU results for five distinct gradient-based methods (MAML \cite{finn2017model,gama2021learning}, MetaSGD \cite{li2017metasgd}, ANIL \cite{raghu2019rapid} %, MetaANIL \cite{li2017metasgd,raghu2019rapid}
    and Reptile \cite{nichol2018reptile}) with the four segmentation backbones pretrained on our meta-dataset and tuned on the few-shot weakly-supervised support samples with the \textit{points} annotation style.}
    \label{tab:results_gradient}
    \begin{tabular}{@{}cccccccccccc@{}}
        \toprule
        \multicolumn{2}{c}{\multirow{2}{*}{\textbf{Method}}} &  & \multicolumn{4}{c}{\textbf{1-shot}}                                                                                                                                                                                                               &  & \multicolumn{4}{c}{\textbf{5-shot}}                                                                                                                                                                                                               \\ \cmidrule(lr){4-7} \cmidrule(l){9-12} 
        \multicolumn{2}{c}{}                                 &  & \textbf{\begin{tabular}[c]{@{}c@{}}p=1\\ r=1\end{tabular}} & \textbf{\begin{tabular}[c]{@{}c@{}}p=1\\ r=3\end{tabular}} & \textbf{\begin{tabular}[c]{@{}c@{}}p=5\\ r=1\end{tabular}} & \textbf{\begin{tabular}[c]{@{}c@{}}p=5\\ r=3\end{tabular}} &  & \textbf{\begin{tabular}[c]{@{}c@{}}p=1\\ r=1\end{tabular}} & \textbf{\begin{tabular}[c]{@{}c@{}}p=1\\ r=3\end{tabular}} & \textbf{\begin{tabular}[c]{@{}c@{}}p=5\\ r=1\end{tabular}} & \textbf{\begin{tabular}[c]{@{}c@{}}p=5\\ r=3\end{tabular}} \\ \cmidrule(r){1-2} \cmidrule(lr){4-7} \cmidrule(l){9-12} 
        \textbf{Baseline}                     & \textbf{R}   &  & {\ul .312}                                                 & {\ul .311}                                                 & {\ul .478}                                                 & {\ul .482}                                                 &  & {\ul .443}                                                 & {\ul .450}                                                 & {\ul .556}                                                 & {\ul .558}                                                 \\ \cmidrule(r){1-2} \cmidrule(lr){4-7} \cmidrule(l){9-12} 
        \multirow{4}{*}{\textbf{MAML}}        & \textbf{U}   &  & .205                                                       & .239                                                       & .321                                                       & .360                                                       &  & .319                                                       & .336                                                       & .476                                                       & .473                                                       \\
                                              & \textbf{R}   &  & .260                                                       & .269                                                       & .387                                                       & .394                                                       &  & {\ul .373}                                                 & {\ul .358}                                                 & .249                                                       & .243                                                       \\
                                              & \textbf{E}   &  & .179                                                       & .185                                                       & .253                                                       & .254                                                       &  & .206                                                       & .207                                                       & .301                                                       & .304                                                       \\
                                              & \textbf{D}   &  & {\ul .328}                                                 & {\ul .335}                                                 & {\ul .424}                                                 & {\ul .431}                                                 &  & .322                                                       & .337                                                       & {\ul .488}                                                 & {\ul .501}                                                 \\ \cmidrule(r){1-2} \cmidrule(lr){4-7} \cmidrule(l){9-12} 
        \multirow{4}{*}{\textbf{MetaSGD}}     & \textbf{U}   &  & {\ul .306}                                                 & {\ul .351}                                                 & .306                                                       & .304                                                       &  & .301                                                       & .300                                                       & .411                                                       & .435                                                       \\
                                              & \textbf{R}   &  & .268                                                       & .277                                                       & .269                                                       & .260                                                       &  & {\ul .340}                                                 & {\ul .390}                                                 & {\ul .441}                                                 & {\ul .473}                                                 \\
                                              & \textbf{E}   &  & .244                                                       & .246                                                       & {\ul .376}                                                 & {\ul .388}                                                 &  & .324                                                       & .320                                                       & .368                                                       & .366                                                       \\
                                              & \textbf{D}   &  & .222                                                       & .234                                                       & .173                                                       & .191                                                       &  & .256                                                       & .272                                                       & .421                                                       & .431                                                       \\ \cmidrule(r){1-2} \cmidrule(lr){4-7} \cmidrule(l){9-12} 
        \multirow{4}{*}{\textbf{ANIL}}        & \textbf{U}   &  & .327                                                       & .323                                                       & .369                                                       & .385                                                       &  & .361                                                       & .366                                                       & .475                                                       & .473                                                       \\
                                              & \textbf{R}   &  & .187                                                       & .203                                                       & .216                                                       & .260                                                       &  & .314                                                       & .341                                                       & .218                                                       & .190                                                       \\
                                              & \textbf{E}   &  & .361                                                       & .364                                                       & .504                                                       & .506                                                       &  & .484                                                       & .490                                                       & .507                                                       & .517                                                       \\
                                              & \textbf{D}   &  & {\ul \textbf{.454}}                                        & {\ul \textbf{.451}}                                        & {\ul \textbf{.532}}                                        & {\ul \textbf{.541}}                                        &  & {\ul \textbf{.546}}                                        & {\ul \textbf{.552}}                                        & {\ul \textbf{.619}}                                        & {\ul \textbf{.627}}                                        \\ \cmidrule(r){1-2} \cmidrule(lr){4-7} \cmidrule(l){9-12} 
        \multirow{4}{*}{\textbf{Reptile}}     & \textbf{U}   &  & {\ul .334}                                                 & {\ul .341}                                                 & {\ul .328}                                                 & .326                                                       &  & .349                                                       & .355                                                       & .365                                                       & .386                                                       \\
                                              & \textbf{R}   &  & .203                                                       & .205                                                       & {\ul .336}                                                 & {\ul .354}                                                 &  & {\ul .394}                                                 & {\ul .414}                                                 & {\ul .524}                                                 & {\ul .536}                                                 \\
                                              & \textbf{E}   &  & .206                                                       & .210                                                       & .281                                                       & .318                                                       &  & .367                                                       & .387                                                       & .314                                                       & .305                                                       \\
                                              & \textbf{D}   &  & .160                                                       & .161                                                       & .210                                                       & .205                                                       &  & .085                                                       & .080                                                       & .283                                                       & .317                                                       \\ \bottomrule
    \end{tabular}
\end{table}

Metric-based methods shown in Table~\ref{tab:results_metric}, in comparison to the baseline with ResNet-12 backbone, highlight the superiority of PANets \cite{wang2019panet} in comparison to ProtoNets \cite{snell2017prototypical} in all scenarios. The better performance of PANets can be attributed to two factors: the use of the cosine distance instead of Euclidean distance and the additional PAR regularization proposed by Wang \textit{et al.} \cite{wang2019panet} -- further explained in Section~\ref{sec:related_meta_segmentation}. EfficientLab outperformed other backbones in all but the two most sparsely labeled scenarios (1-shot/p=1), while the other architectures performed quite well in 1-shot/p=1/r=1 and 1-shot/p=1/r=3, but lacked the capacity of learning from more annotations as well as PANet-\textbf{E}.

% \Isa{IT IS NOT CLEAR WHY ResNet IS THE BEST BASELINE, WHILE THE TESTED APPROACHES PERFORM BETTER WITH ANOTHER BACK-BONE (D FOR ANIL, etc.)}

\begin{table}[!ht]
    \centering
    \caption{IoU results for two distinct metric-based methods (ProtoNets \cite{snell2017prototypical,gama2022weakly} and PANets \cite{wang2019panet}) with the four segmentation backbones pretrained on our meta-dataset and tuned on the few-shot weakly-supervised support samples with the \textit{points} annotation style.}
    \label{tab:results_metric}
    \begin{tabular}{@{}cclcccclcccc@{}}
        \toprule
        \multicolumn{2}{c}{\multirow{2}{*}{\textbf{Method}}} &  & \multicolumn{4}{c}{\textbf{1-shot}}                                                                                                                                                                                                               &  & \multicolumn{4}{c}{\textbf{5-shot}}                                                                                                                                                                                                               \\ \cmidrule(lr){4-7} \cmidrule(l){9-12} 
        \multicolumn{2}{c}{}                                 &  & \textbf{\begin{tabular}[c]{@{}c@{}}p=1\\ r=1\end{tabular}} & \textbf{\begin{tabular}[c]{@{}c@{}}p=1\\ r=3\end{tabular}} & \textbf{\begin{tabular}[c]{@{}c@{}}p=5\\ r=1\end{tabular}} & \textbf{\begin{tabular}[c]{@{}c@{}}p=5\\ r=3\end{tabular}} &  & \textbf{\begin{tabular}[c]{@{}c@{}}p=1\\ r=1\end{tabular}} & \textbf{\begin{tabular}[c]{@{}c@{}}p=1\\ r=3\end{tabular}} & \textbf{\begin{tabular}[c]{@{}c@{}}p=5\\ r=1\end{tabular}} & \textbf{\begin{tabular}[c]{@{}c@{}}p=5\\ r=3\end{tabular}} \\ \cmidrule(r){1-2} \cmidrule(lr){4-7} \cmidrule(l){9-12} 
        \textbf{Baseline}                     & \textbf{R}   &  & {\ul .312}                                                 & {\ul .311}                                                 & {\ul .478}                                                 & {\ul .482}                                                 &  & {\ul .443}                                                 & {\ul .450}                                                 & {\ul .556}                                                 & {\ul .558}                                                 \\ \cmidrule(r){1-2} \cmidrule(lr){4-7} \cmidrule(l){9-12} 
        \multirow{4}{*}{\textbf{ProtoNet}}    & \textbf{U}   &  & .371                                                       & .365                                                       & .364                                                       & .362                                                       &  & {\ul .454}                                                 & {\ul .462}                                                 & {\ul .558}                                                 & {\ul .563}                                                 \\
                                              & \textbf{R}   &  & .363                                                       & .364                                                       & .397                                                       & .395                                                       &  & {\ul .460}                                                 & {\ul .463}                                                 & .533                                                       & .540                                                       \\
                                              & \textbf{E}   &  & .319                                                       & .336                                                       & {\ul .445}                                                 & {\ul .453}                                                 &  & .448                                                       & .449                                                       & .521                                                       & .525                                                       \\
                                              & \textbf{D}   &  & {\ul .382}                                                 & {\ul .381}                                                 & .403                                                       & .399                                                       &  & {\ul .458}                                                 & {\ul .456}                                                 & {\ul .556}                                                 & {\ul .568}                                                 \\ \cmidrule(r){1-2} \cmidrule(lr){4-7} \cmidrule(l){9-12} 
        \multirow{4}{*}{\textbf{PANet}}       & \textbf{U}   &  & {\ul \textbf{.416}}                                        & {\ul \textbf{.411}}                                        & .419                                                       & .432                                                       &  & .511                                                       & .509                                                       & .554                                                       & .561                                                       \\
                                              & \textbf{R}   &  & {\ul \textbf{.423}}                                        & {\ul \textbf{.418}}                                        & .512                                                       & .510                                                       &  & .518                                                       & .522                                                       & .592                                                       & .600                                                       \\
                                              & \textbf{E}   &  & .295                                                       & .303                                                       & {\ul \textbf{.570}}                                        & {\ul \textbf{.581}}                                        &  & {\ul \textbf{.530}}                                        & {\ul \textbf{.539}}                                        & {\ul \textbf{.615}}                                        & {\ul \textbf{.622}}                                        \\
                                              & \textbf{D}   &  & {\ul \textbf{.419}}                                        & {\ul \textbf{.421}}                                        & .462                                                       & .462                                                       &  & .481                                                       & .484                                                       & .578                                                       & .584                                                       \\ \bottomrule
    \end{tabular}
\end{table}

\begin{table}[!ht]
    \centering
    \caption{IoU results for three distinct fusion-based methods (Guided Nets \cite{rakelly2018few}, R2D2 \cite{bertinetto2018meta} and MetaOptNet \cite{lee2019meta}) with the four segmentation backbones pretrained on our meta-dataset and tuned on the few-shot weakly-supervised support samples with the \textit{points} annotation style.}
    \label{tab:results_fusion}
    \begin{tabular}{@{}cclcccclcccc@{}}
        \toprule
        \multicolumn{2}{c}{\multirow{2}{*}{\textbf{\begin{tabular}[c]{@{}c@{}}Method\\ and\\ Backbone\end{tabular}}}} &  & \multicolumn{4}{c}{\textbf{1-shot}}                                                                                                                                                                                                               &  & \multicolumn{4}{c}{\textbf{5-shot}}                                                                                                                                                                                                               \\ \cmidrule(lr){4-7} \cmidrule(l){9-12} 
        \multicolumn{2}{c}{}                                 &  & \textbf{\begin{tabular}[c]{@{}c@{}}p=1\\ r=1\end{tabular}} & \textbf{\begin{tabular}[c]{@{}c@{}}p=1\\ r=3\end{tabular}} & \textbf{\begin{tabular}[c]{@{}c@{}}p=5\\ r=1\end{tabular}} & \textbf{\begin{tabular}[c]{@{}c@{}}p=5\\ r=3\end{tabular}} &  & \textbf{\begin{tabular}[c]{@{}c@{}}p=1\\ r=1\end{tabular}} & \textbf{\begin{tabular}[c]{@{}c@{}}p=1\\ r=3\end{tabular}} & \textbf{\begin{tabular}[c]{@{}c@{}}p=5\\ r=1\end{tabular}} & \textbf{\begin{tabular}[c]{@{}c@{}}p=5\\ r=3\end{tabular}} \\ \cmidrule(r){1-2} \cmidrule(lr){4-7} \cmidrule(l){9-12} 
        \textbf{Baseline}                      & \textbf{R}  &  & {\ul .312}                                                 & {\ul .311}                                                 & {\ul \textbf{.478}}                                        & {\ul \textbf{.482}}                                        &  & {\ul .443}                                                 & {\ul .450}                                                 & {\ul .556}                                                 & {\ul .558}                                                 \\ \cmidrule(r){1-2} \cmidrule(lr){4-7} \cmidrule(l){9-12} 
        \multirow{4}{*}{\textbf{Guided Net}}   & \textbf{U}  &  & .015                                                       & .013                                                       & .020                                                       & .012                                                       &  & .000                                                       & .000                                                       & .000                                                       & .000                                                       \\
                                               & \textbf{R}  &  & {\ul \textbf{.450}}                                        & {\ul \textbf{.449}}                                        & {\ul .447}                                                 & {\ul .443}                                                 &  & {\ul .450}                                                 & {\ul .450}                                                 & {\ul .449}                                                 & {\ul .447}                                                 \\
                                               & \textbf{E}  &  & .295                                                       & .298                                                       & .286                                                       & .306                                                       &  & .272                                                       & .272                                                       & .272                                                       & .291                                                       \\
                                               & \textbf{D}  &  & .150                                                       & .161                                                       & .140                                                       & .164                                                       &  & .254                                                       & .259                                                       & .260                                                       & .265                                                       \\ \cmidrule(r){1-2} \cmidrule(lr){4-7} \cmidrule(l){9-12} 
        \multirow{4}{*}{\textbf{R2D2}}         & \textbf{U}  &  & .320                                                       & .324                                                       & .292                                                       & .341                                                       &  & .390                                                       & .419                                                       & .531                                                       & .550                                                       \\
                                               & \textbf{R}  &  & {\ul .365}                                                 & {\ul .365}                                                 & {\ul .425}                                                 & .399                                                       &  & .487                                                       & .502                                                       & .596                                                       & .614                                                       \\
                                               & \textbf{E}  &  & .268                                                       & .272                                                       & .385                                                       & {\ul .422}                                                 &  & {\ul \textbf{.529}}                                        & {\ul \textbf{.526}}                                        & .677                                                       & .681                                                       \\
                                               & \textbf{D}  &  & {\ul .362}                                                 & {\ul .362}                                                 & {\ul .415}                                                 & {\ul .423}                                                 &  & .436                                                       & .460                                                       & {\ul \textbf{.706}}                                        & {\ul \textbf{.715}}                                        \\ \cmidrule(r){1-2} \cmidrule(lr){4-7} \cmidrule(l){9-12} 
        \multirow{4}{*}{\textbf{MetaOptNet}}   & \textbf{U}  &  & {\ul .383}                                                 & {\ul .379}                                                 & .415                                                       & .408                                                       &  & .424                                                       & .423                                                       & .458                                                       & .458                                                       \\
                                               & \textbf{R}  &  & {\ul .390}                                                 & {\ul .382}                                                 & .400                                                       & .382                                                       &  & .494                                                       & .489                                                       & .627                                                       & .640                                                       \\
                                               & \textbf{E}  &  & .257                                                       & .262                                                       & .399                                                       & {\ul .444}                                                 &  & {\ul .513}                                                 & {\ul .515}                                                 & {\ul .659}                                                 & {\ul .670}                                                 \\
                                               & \textbf{D}  &  & .358                                                       & {\ul .377}                                                 & {\ul .426}                                                 & {\ul .448}                                                 &  & .435                                                       & .486                                                       & {\ul .654}                                                 & {\ul .662}                                                 \\ \bottomrule
    \end{tabular}
\end{table}

Guided Nets, mainly with a ResNet-12 backbone, had a strong start in extremely low-data scenarios (i.e. 1-shot/p=1), but were unable to evolve to learn from more annotations, maintaining their performance close to 0.45 of IoU all throughout the other experiments. As for R2D2 and MetaOptNet, their most promising backbones reach relatively similar performances in very sparsely-annotated scenarios -- from 0.36 to 0.39 in 1-shot/p=1 -- with the considerable advantage that these architectures are able to continue learning from the larger amount of annotated data in 1-shot/p=5 and 5-shot. We highlight that R2D2-\textbf{E}, R2D2-\textbf{D} and MetaOptNet-\textbf{E}, MetaOptNet-\textbf{D} show good performances whenever larger amounts of annotated support pixels are provided.

From the results presented in Tables~\ref{tab:results_baseline},~\ref{tab:results_gradient},~\ref{tab:results_metric} and~\ref{tab:results_fusion}, we chose 7 distinct algorithm/backbone pairs for further analysing in the following sections: Baseline-\textbf{R}, ANIL-\textbf{E}/\textbf{D}, PANets-\textbf{R}/\textbf{E}, R2D2-\textbf{D} and MetaOptNet-\textbf{D}.

\subsection{Weak Annotation Styles}
\label{sec:results_weak_annotations}

While Section~\ref{sec:results_intra} focused on results for \textit{points} annotations, the present section shows results for three additional weakly-supervised mask styles: \textit{grid}, \textit{scribbles} and \textit{skeleton}.

\renewcommand{\currprop}{0.49\textwidth}
\begin{figure*}[!th]
    \centering
    \begin{subfigure}[b]{\currprop}
        % \centering
        \begin{overpic}[clip,trim=0.15in 0.1in 0in 0.5in,width=\columnwidth]{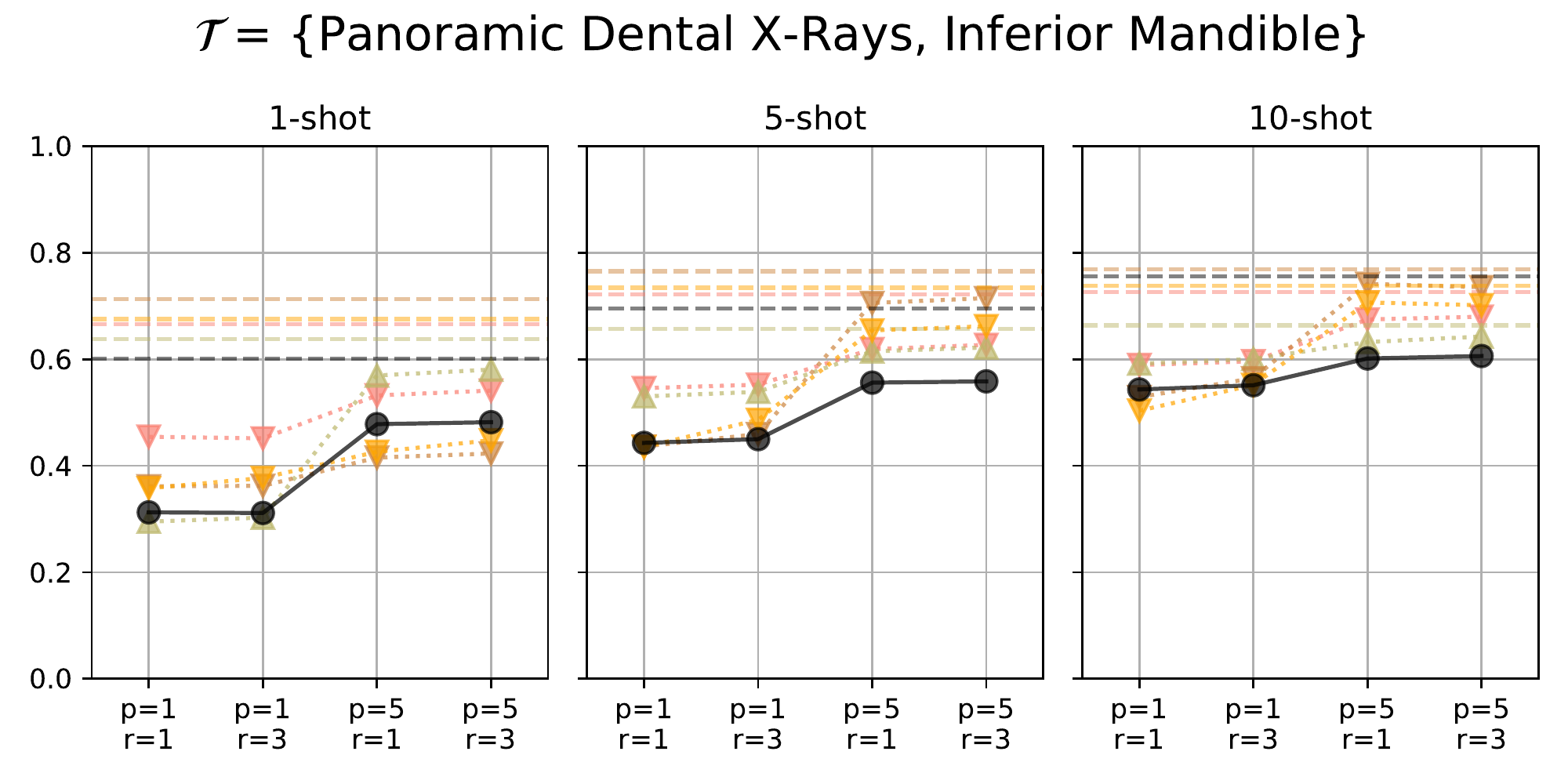}
            \put(63,7){\includegraphics[height=0.05\columnwidth]{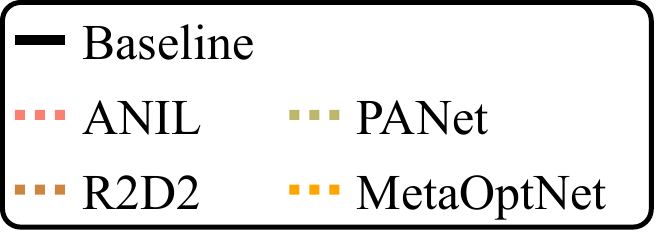}}
            \put(39,7){\includegraphics[height=0.05\columnwidth]{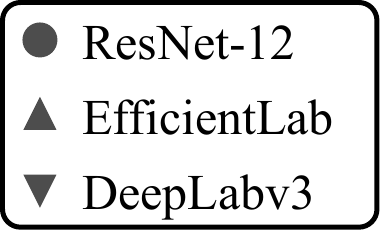}}
        \end{overpic}
        \caption{\textit{Panoramic-mandible}/\textit{points}.}
        \label{fig:results_weak_annotations_points_panoramic}
    \end{subfigure}
    \hfill
    \begin{subfigure}[b]{\currprop}
        % \centering
        \includegraphics[clip,trim=0.15in 0.1in 0in 0.5in,width=\columnwidth]{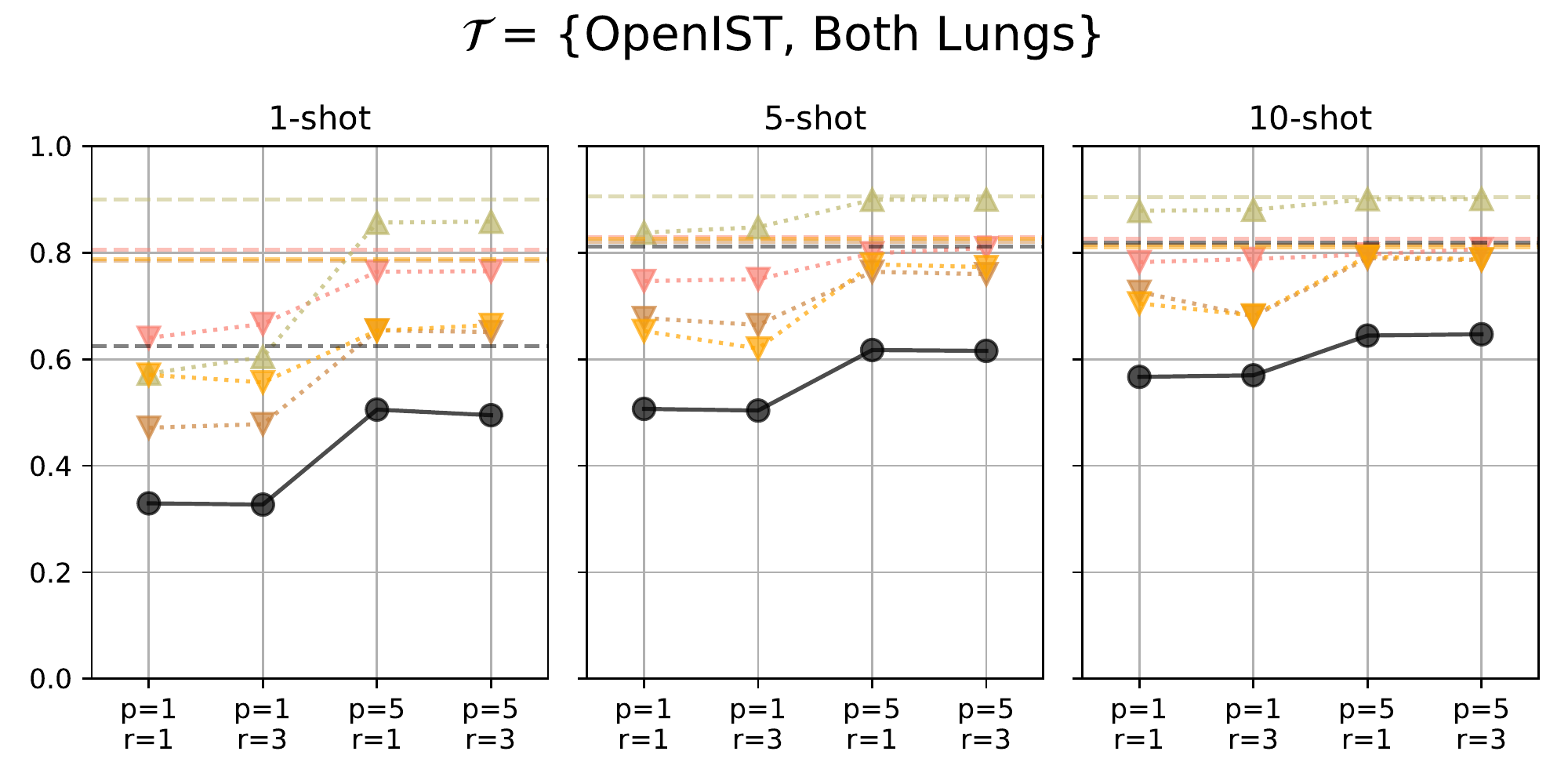}
        \caption{\textit{OpenIST-lungs}/\textit{points}.}
        \label{fig:results_weak_annotations_points_openist}
    \end{subfigure}
    \\
    \begin{subfigure}[b]{\currprop}
        % \centering
        \includegraphics[clip,trim=0.15in 0.1in 0in 0.5in,width=\columnwidth]{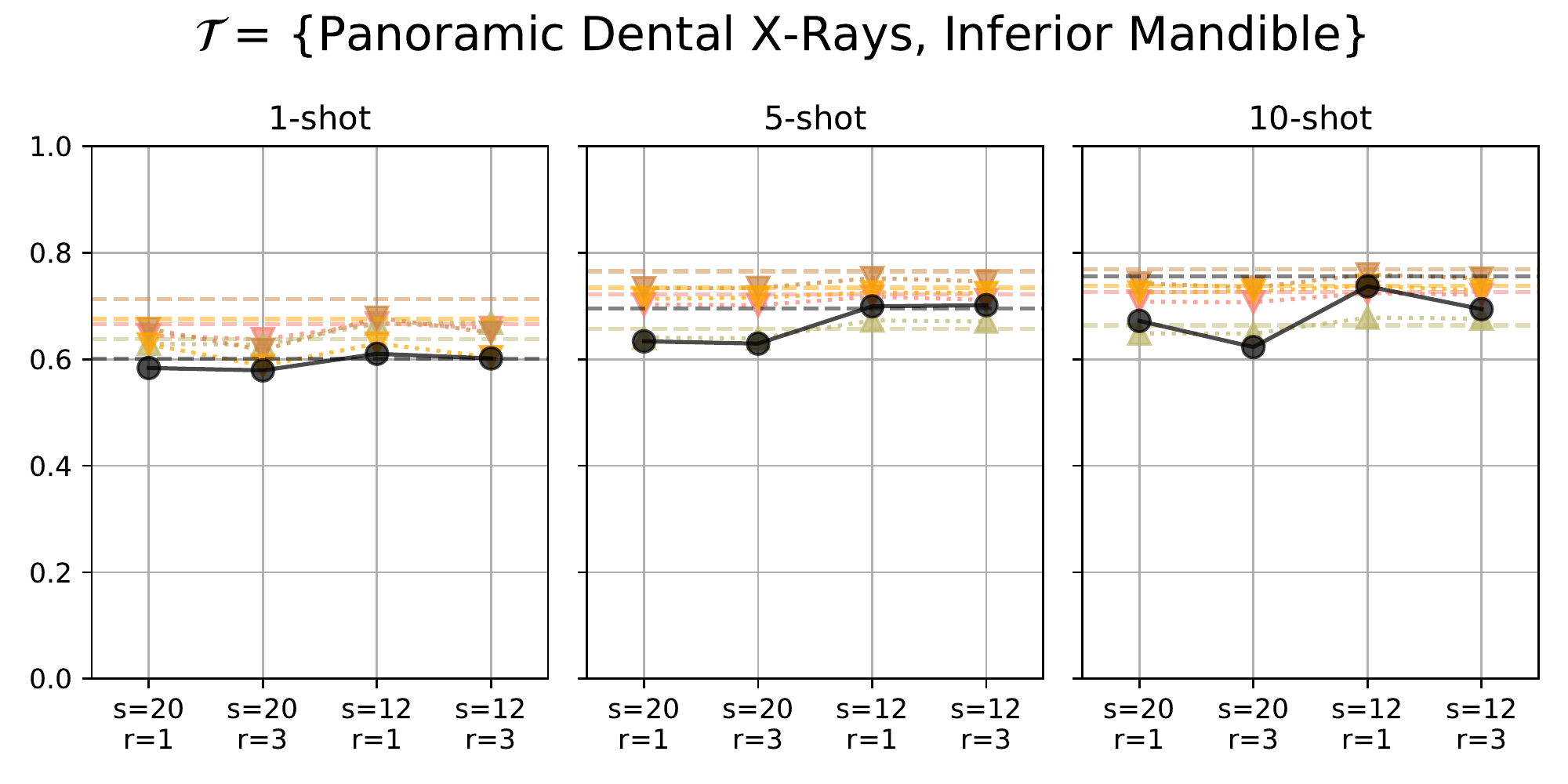}
        \caption{\textit{Panoramic-mandible}/\textit{grid}.}
        \label{fig:results_weak_annotations_grid_panoramic}
    \end{subfigure}
    \hfill
    \begin{subfigure}[b]{\currprop}
        % \centering
        \includegraphics[clip,trim=0.15in 0.1in 0in 0.5in,width=\columnwidth]{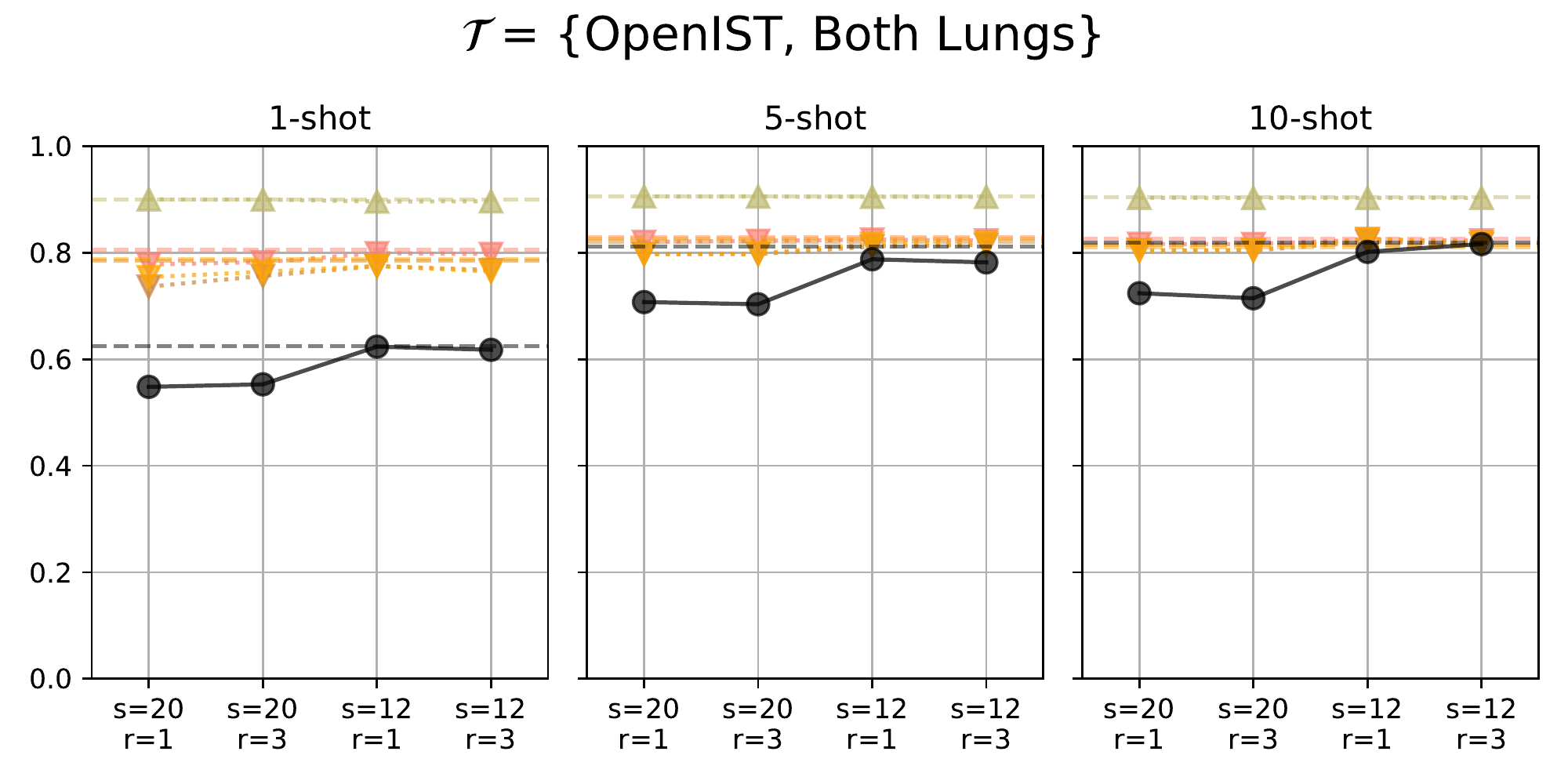}
        \caption{\textit{OpenIST-lungs}/\textit{grid}.}
        \label{fig:results_weak_annotations_grid_openist}
    \end{subfigure}
    \\
    \begin{subfigure}[b]{\currprop}
        % \centering
        \includegraphics[clip,trim=0.15in 0.1in 0in 0.5in,width=\columnwidth]{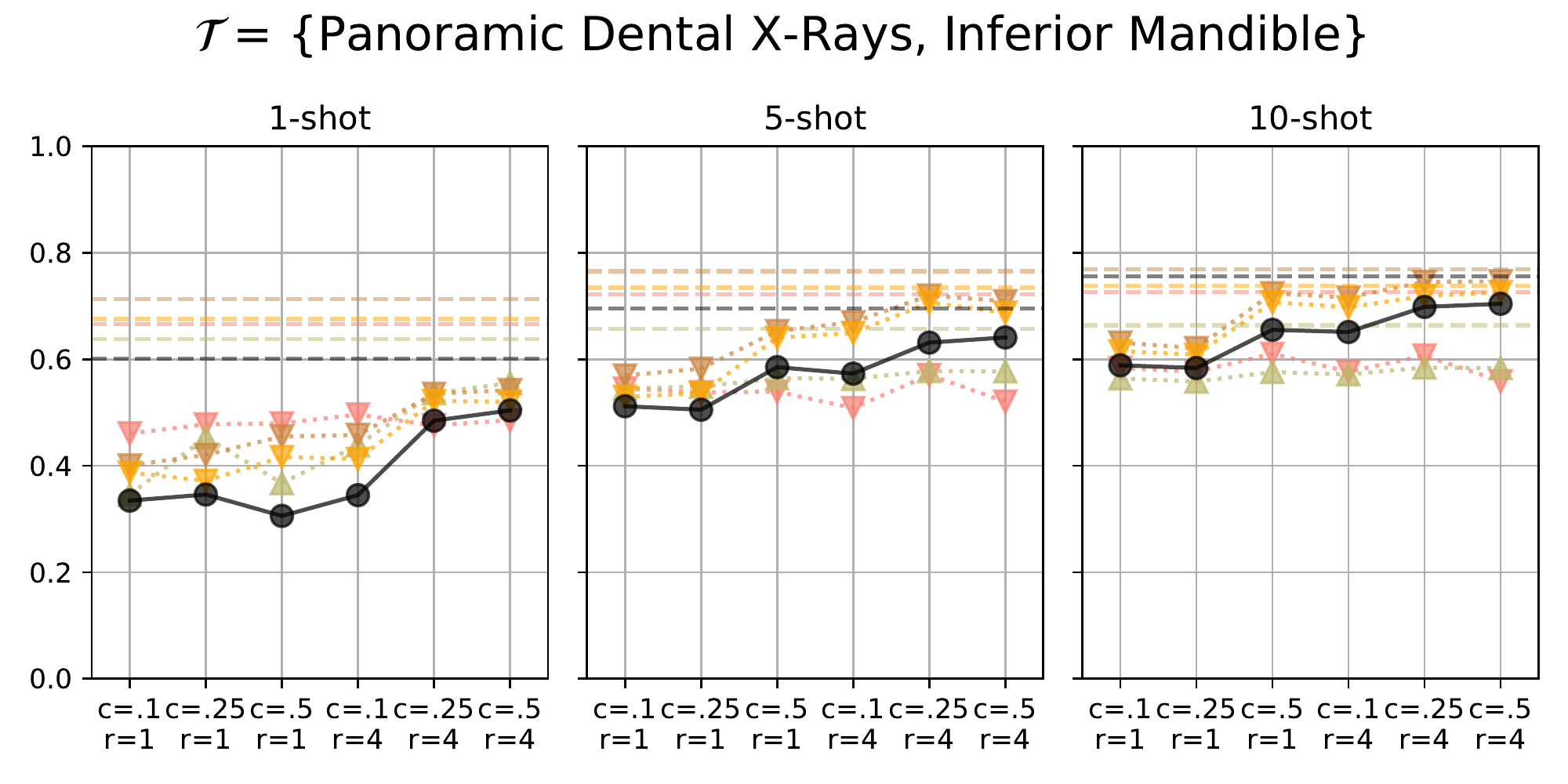}
        \caption{\textit{Panoramic-mandible}/\textit{scribbles}.}
        \label{fig:results_weak_annotations_scribbles_panoramic}
    \end{subfigure}
    \hfill
    \begin{subfigure}[b]{\currprop}
        % \centering
        \includegraphics[clip,trim=0.15in 0.1in 0in 0.5in,width=\columnwidth]{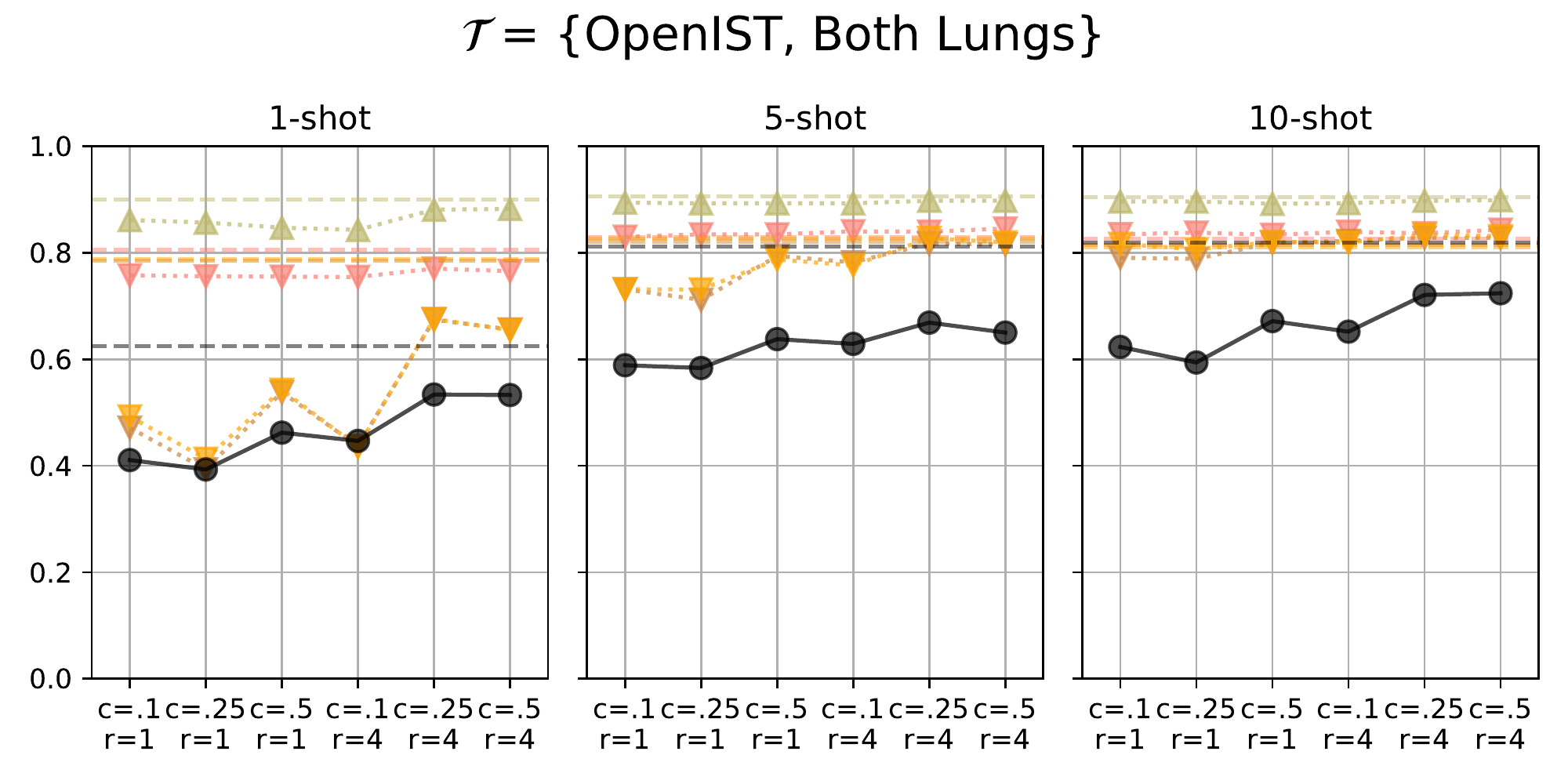}
        \caption{\textit{OpenIST-lungs}/\textit{scribbles}.}
        \label{fig:results_weak_annotations_scribbles_openist}
    \end{subfigure}
    \\
    \begin{subfigure}[b]{\currprop}
        % \centering
        \includegraphics[clip,trim=0.15in 0.1in 0in 0.5in,width=\columnwidth]{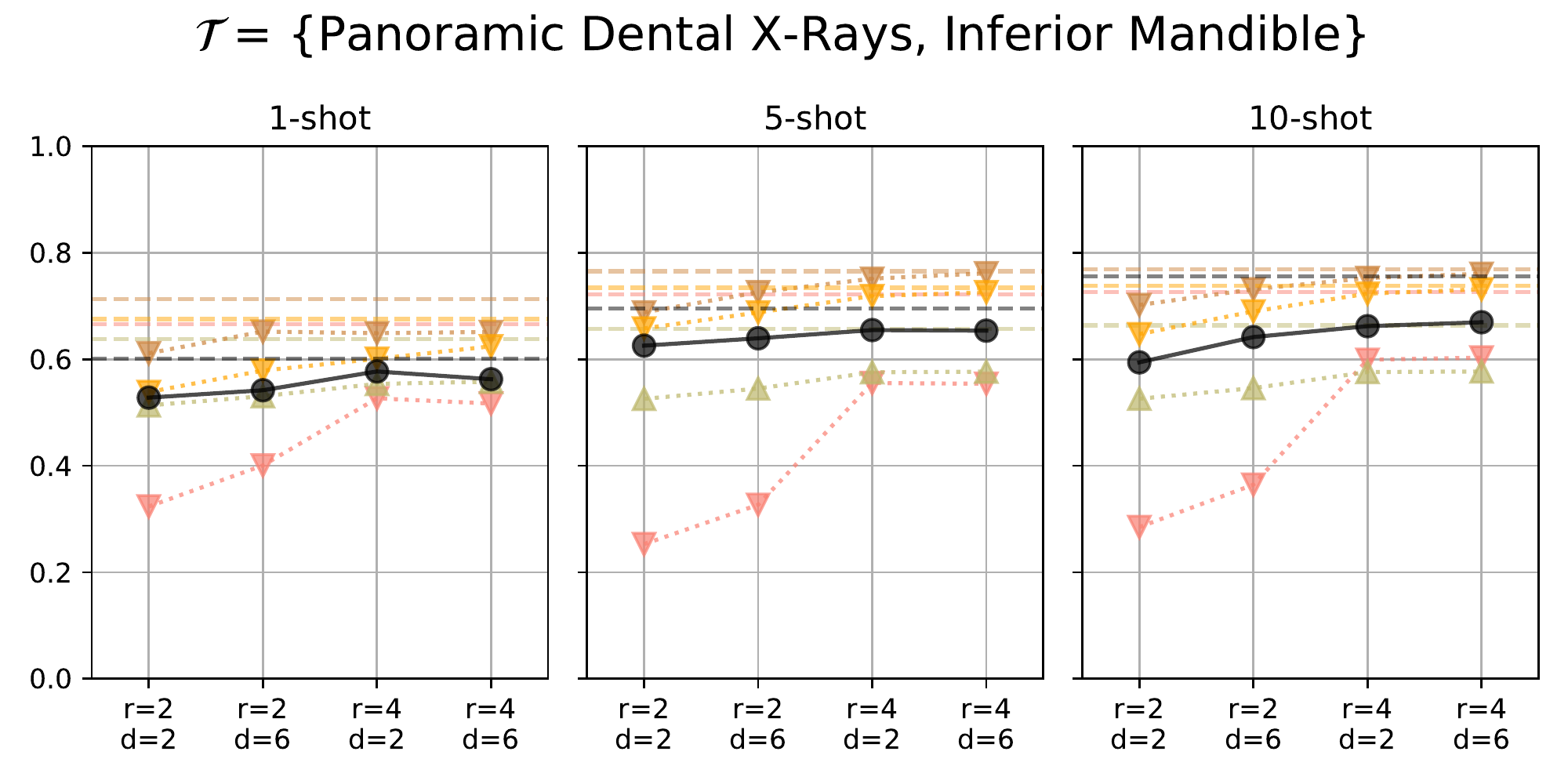}
        \caption{\textit{Panoramic-mandible}/\textit{skeleton}.}
        \label{fig:results_weak_annotations_skeleton_panoramic}
    \end{subfigure}
    \hfill
    \begin{subfigure}[b]{\currprop}
        % % \centering
        % \begin{overpic}[clip,trim=0.15in 0.1in 0in 0.5in,width=\columnwidth]{openist_both_lungs_skeleton_weak_annotations.pdf}
        %     \put(47.5,7){\includegraphics[width=0.50\columnwidth]{result_plot_legends_weak_methods.pdf}}
        %     \put(7,7){\includegraphics[width=0.220588235\columnwidth]{result_plot_legends_weak_networks.pdf}}
        % \end{overpic}
        \includegraphics[clip,trim=0.15in 0.1in 0in 0.5in,width=\columnwidth]{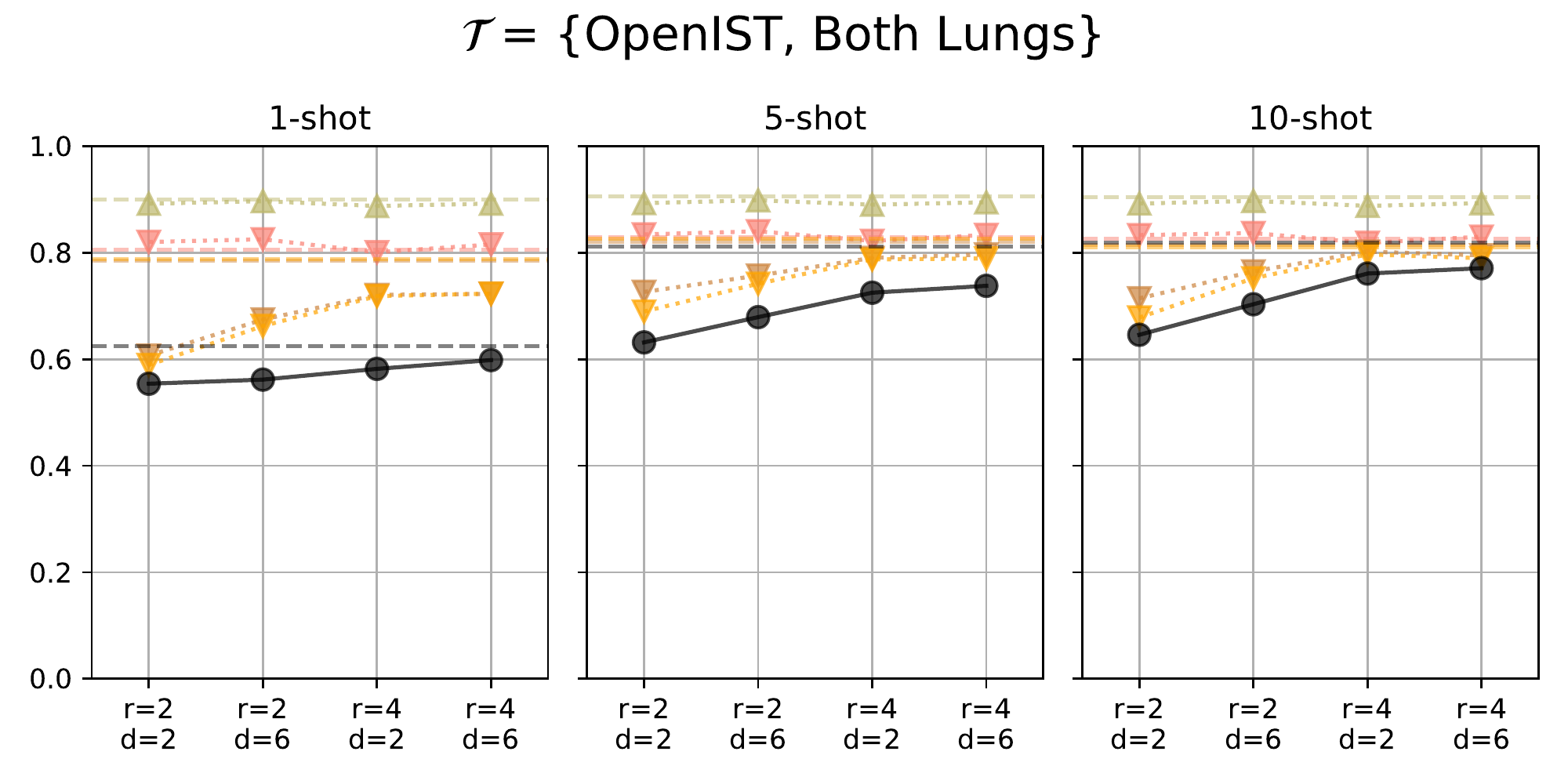}
        \caption{\textit{OpenIST-lungs}/\textit{skeleton}.}
        \label{fig:results_weak_annotations_skeleton_openist}
    \end{subfigure}
    \caption{IoU results for multiple weakly-supervised annotation styles in 1-, 5- and 10-shots for the best baseline backbone (\textbf{R}) and the two overall best algorithm/backbone pairs in each paradigm (ANIL-\textbf{D}, PANets-\textbf{E}, R2D2-\textbf{D} and MetaOptNet-\textbf{D}). Rows reflect distinct weakly-supervised annotation style, respectively: \textit{points}, \textit{grid}, \textit{scribbles}, \textit{skeleton}. The left column represents results for the \textit{Panoramic-mandible} task (large domain shift), while the rightmost column depicts results for the \textit{OpenIST-lungs}. Dotted lines reflect the performance of segmentation algorithms on the weakly-supervised support sets, while the dashed horizontal lines show the results tuned on the densely annotated support masks. Better viewed in color.}
    \label{fig:results_weak_annotations}
\end{figure*}

Figure~\ref{fig:results_weak_annotations} depicts some broader trends applied to all experiments. As expected, the \textit{OpenIST-lungs} segmentation task -- wherein the domain shift is smaller both in the pixel- and label-space in comparison to the meta-dataset -- achieves considerably better segmentation performances than the \textit{Panoramic-mandible} segmentation. Even in very sparsely labeled support sets (i.e. \textit{points} in 1-shot/p=1), some algorithms reach 0.6 or more of IoU, while the 5-shot/p=1 scenario presents performances for PANets with IoU larger than 0.8. All other annotation styles \textit{grid}, \textit{scribbles} and \textit{skeleton} in their most sparsely annotated scenarios even reach 0.9 of IoU for \textit{OpenIST-lungs}. Similarly to results reported by Gama \textit{et al.} \cite{gama2022weakly}, metric-based methods seem to be much more suitable to predict tasks in target domains where the domain shift from the meta-dataset is small. ANIL-\textbf{D} follows PANets quite closely in this task, while fusion-based approaches reached a lower ceiling in performance on \textit{OpenIST-lungs}.

One should notice that for OpenIST only annotations in randomly selected \textit{points} did not reach around 0.8 of IoU, while all of \textit{grid}, \textit{scribbles} and \textit{skeleton} did. This quickly reaching ceiling in performance in all but one annotation style implies that \textit{points} are a less label-efficient way of providing weak annotations.

Depending on the meta-learner, \textit{points} in its more sparse setting (1-shot/p=1) achieves between around 0.45/0.70 of IoU for the \textit{Panoramic-mandible} and \textit{OpenIST-lungs} tasks, respectively, while the highly efficient \textit{skeleton} annotation style in its more sparse setting (1-shot/r=2) yields 0.60/0.90 for MetaOptNet/PANets. In fact, annotations in \textit{skeleton} and \textit{scribbles} seem to me more efficient than \textit{points}, mainly due to the fact that the time spent in labeling single random points in an image is similar to the time spent in delineating a scribble or a skeleton for a commonly-shaped organ, while also providing more annotated points to tune the learner into the target task. Both draw-oriented annotations also provide a relatively clear guide for the algorithm indicating the organ borders, while randomly-sampled points do not provide this information.% for the meta-learners.

Another clearly seen trend in Figure~\ref{fig:results_weak_annotations} is the very small performance gap between sparse (dotted lines) and dense (dashed horizontal lines) annotations. While the gap is considerably large in \textit{Panoramic-mandible} for more sparsely annotated scenarios in all annotation styles, for \textit{OpenIST-lungs} in all scenarios and \textit{Panoramic-mandible} in more densely annotated scenarios it is negligible for the best meta-learners. %This is specially useful in FWS scenarios, as novel domains do not require large amounts of weakly-annotated data to train a segmentator.

% Additionally to the quantitative analysis conducted previously, we also show in Figure~\ref{fig:results_qualitative} sample segmentation predictions from 4 meta-learners and a baseline in 2D radiology data.
We show in Figure~\ref{fig:results_qualitative} sample segmentation predictions from 4 meta-learners and a baseline in 2D radiology data. Following the trends shown in Figure~\ref{fig:results_weak_annotations}, qualitative segmentation predictions in the \textit{Panoramic-mandible} task are better performed by the fusion-based methods (R2D2-\textbf{D} and MetaOptNet-\textbf{D}). As for the task with small domain shift, the metric-based PANets-\textbf{E} work better in \textit{OpenIST-lungs}, followed by the gradient-based ANIL-\textbf{D}. One can also derive from the 4 top lines in Figure~\ref{fig:results_qualitative} that the \textit{points} style is the least effective weak annotation modality for both \textit{Panoramic-mandible} and \textit{OpenIST-lungs}, while \textit{scribbles} and \textit{skeleton} labels show overall superior performance for the best meta-learners in each scenario.

\renewcommand{\currprop}{0.93\textwidth}
\begin{figure*}[!th]
    \centering
    \includegraphics[clip,trim=4.7in 0.4in 0.0in 0.35in,width=\currprop]{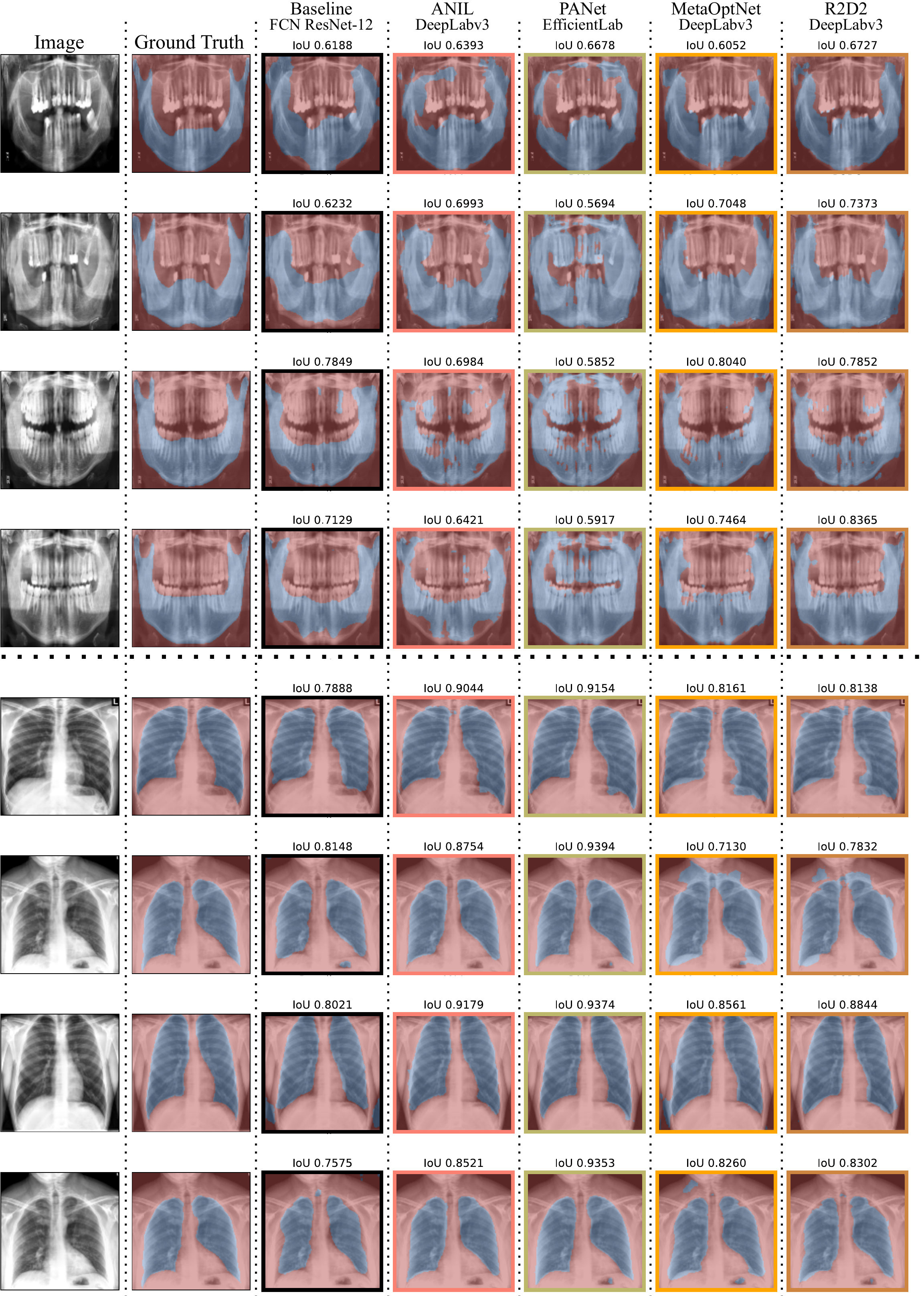}
    \caption{Qualitative results for FWS on two target 10-shot tasks: \textit{Panoramic-mandible} (top 4 rows) and \textit{OpenIST-lungs} (bottom 4 rows); for 4 distinct annotation styles: \textit{points} ($1^{st}$ and $5^{th}$ rows), \textit{grid} ($2^{nd}$ and $6^{th}$ rows), \textit{scribbles} ($3^{rd}$ and $7^{th}$ rows) and \textit{skeleton} ($4^{th}$ and $8^{th}$ rows). The positive class is represented in blue and the negative class is shown in red. For each task/annotation style, we show one sample of the query set ($\F^{qry}$). Segmentation predictions and sample IoU metrics for 4 of the best meta-learners (ANIL-\textbf{D}, PANet-\textbf{E}, MetaOptNet-\textbf{D} and R2D2-\textbf{D}) and Baseline-\textbf{R} are shown. Better viewed in color.}
    \label{fig:results_qualitative}
\end{figure*}

\subsection{Results for 2D Slices from Volumetric Data}
\label{sec:results_slices}

This section presents our experiments on 2D slices from 3D data, representing fully OOD tasks in relation to the meta-training datasets. Quantitative results for four tasks (\textit{StructSeg-lungs}, \textit{StructSeg-heart}, \textit{MSD-spleen} and \textit{STAP-cerebellum}) are shown in Figure~\ref{fig:results_weak_annotations_3d} for two annotation styles: \textit{grid} and \textit{skeleton}. As all 2D slice tasks are considerably more OOD in comparison to the meta-training datasets, we observe similar results to the ones presented for the \textit{Panoramic-mandible} task in Figure~\ref{fig:results_weak_annotations}, with fusion-based methods achieving higher performance than their counterparts. These best results obtained by R2D2 and MetaOptNet are followed by the optimization-based ANIL, with the similarity-based PANets showing performance close to the baseline in the majority of tasks and annotation styles. Weak \textit{grid} annotations seemed to perform better than the skeleton labeling style in 2D slice tasks, possibly due to the spatial uniformity of such annotations in tasks with large domain shifts, even though it is often more time-demanding to label a grid of points rather than simply draw a skeleton and outline of an organ.

\renewcommand{\currprop}{0.49\textwidth}
\begin{figure*}[!th]
    \centering
    \begin{subfigure}[b]{\currprop}
        % \centering
        \begin{overpic}[clip,trim=0.15in 0.1in 0in 0.5in,width=\columnwidth]{panoramic_mandible_points_weak_annotations.pdf}
            \put(63,7){\includegraphics[height=0.05\columnwidth]{result_plot_legends_weak_methods.pdf}}
            \put(39,7){\includegraphics[height=0.05\columnwidth]{result_plot_legends_weak_networks.pdf}}
        \end{overpic}
        \caption{\textit{StructSeg-lungs}/\textit{grid}.}
        \label{fig:results_weak_annotations_3d_grid_lungs}
    \end{subfigure}
    \hfill
    \begin{subfigure}[b]{\currprop}
        % \centering
        \includegraphics[clip,trim=0.15in 0.1in 0in 0.5in,width=\columnwidth]{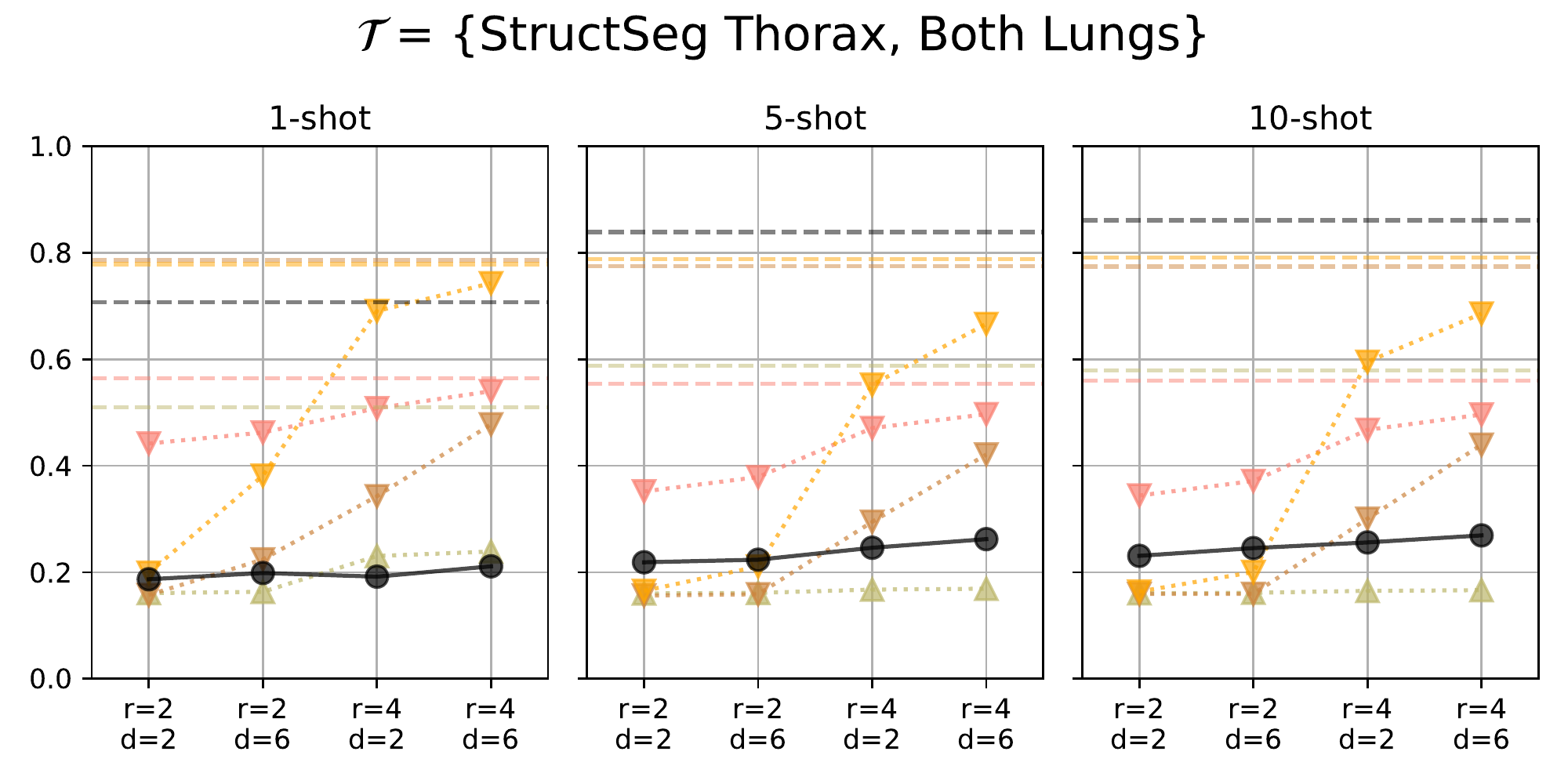}
        \caption{\textit{StructSeg-lungs}/\textit{skeleton}.}
        \label{fig:results_weak_annotations_3d_skeleton_lungs}
    \end{subfigure}
    \\
    \begin{subfigure}[b]{\currprop}
        % \centering
        \includegraphics[clip,trim=0.15in 0.1in 0in 0.5in,width=\columnwidth]{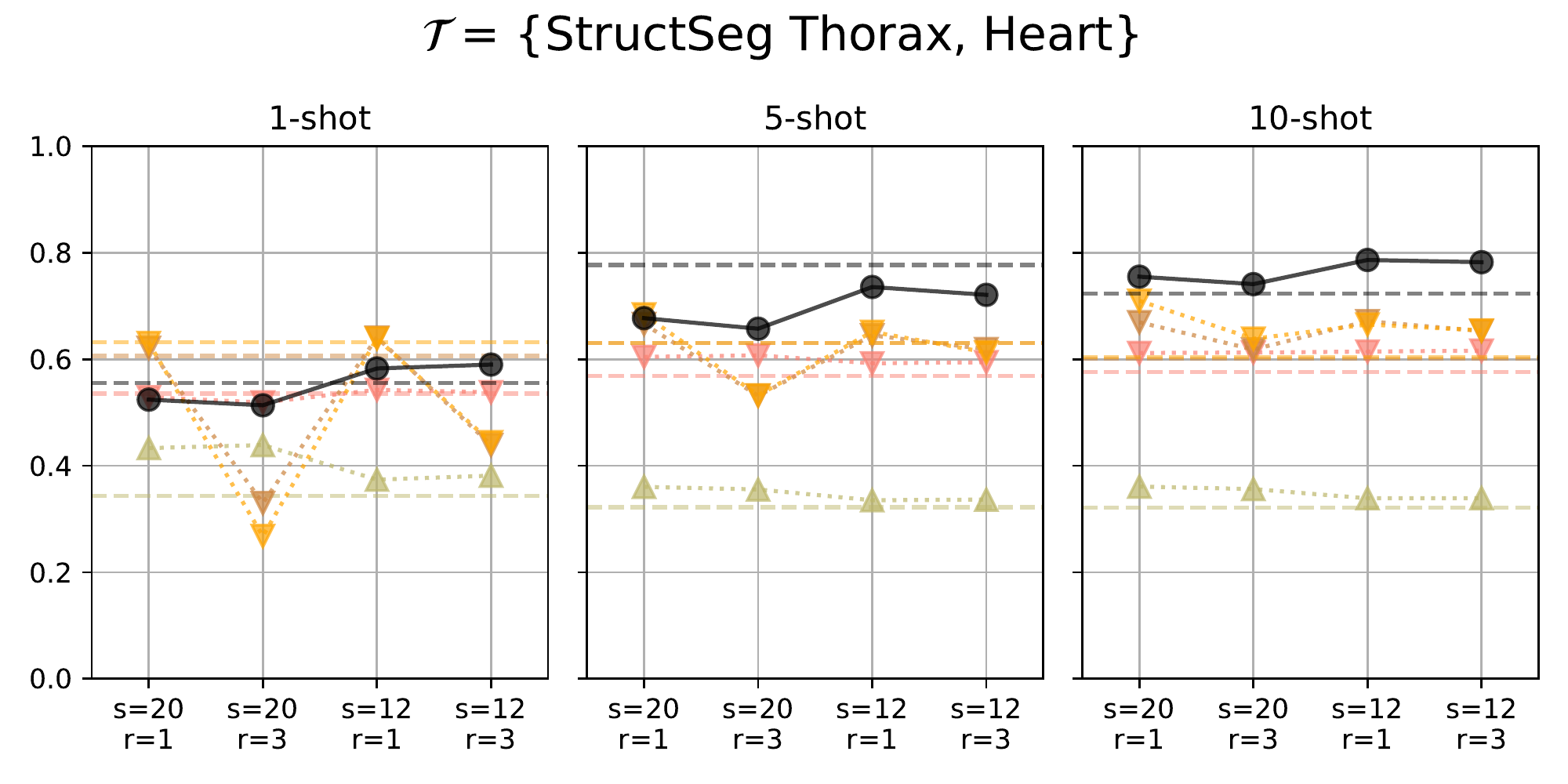}
        \caption{\textit{StructSeg-heart}/\textit{grid}.}
        \label{fig:results_weak_annotations_3d_grid_heart}
    \end{subfigure}
    \hfill
    \begin{subfigure}[b]{\currprop}
        % \centering
        \includegraphics[clip,trim=0.15in 0.1in 0in 0.5in,width=\columnwidth]{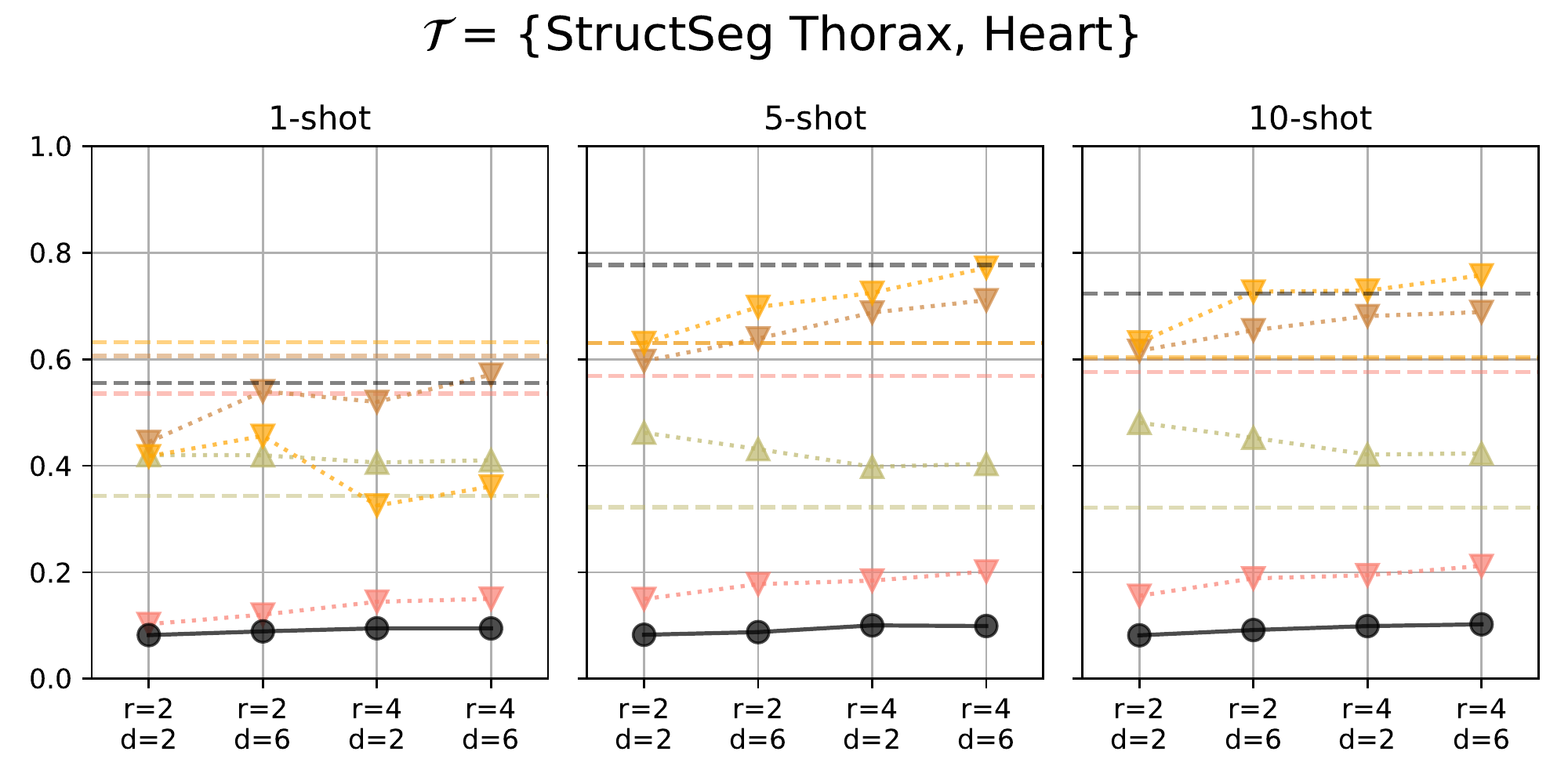}
        \caption{\textit{StructSeg-heart}/\textit{skeleton}.}
        \label{fig:results_weak_annotations_3d_skeleton_heart}
    \end{subfigure}
    \\
    \begin{subfigure}[b]{\currprop}
        % \centering
        \includegraphics[clip,trim=0.15in 0.1in 0in 0.5in,width=\columnwidth]{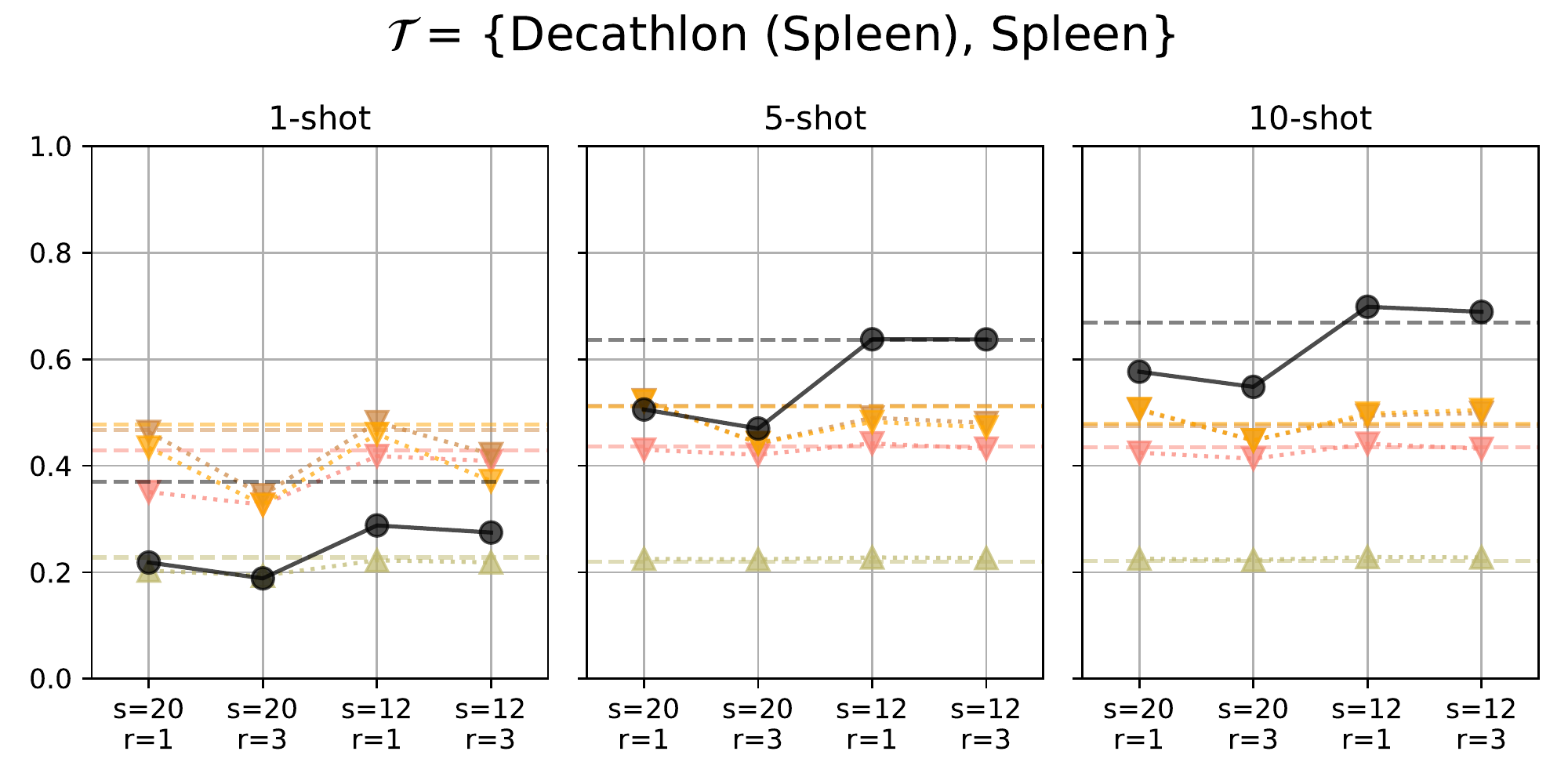}
        \caption{\textit{MSD-spleen}/\textit{grid}.}
        \label{fig:results_weak_annotations_3d_grid_spleen}
    \end{subfigure}
    \hfill
    \begin{subfigure}[b]{\currprop}
        % \centering
        \includegraphics[clip,trim=0.15in 0.1in 0in 0.5in,width=\columnwidth]{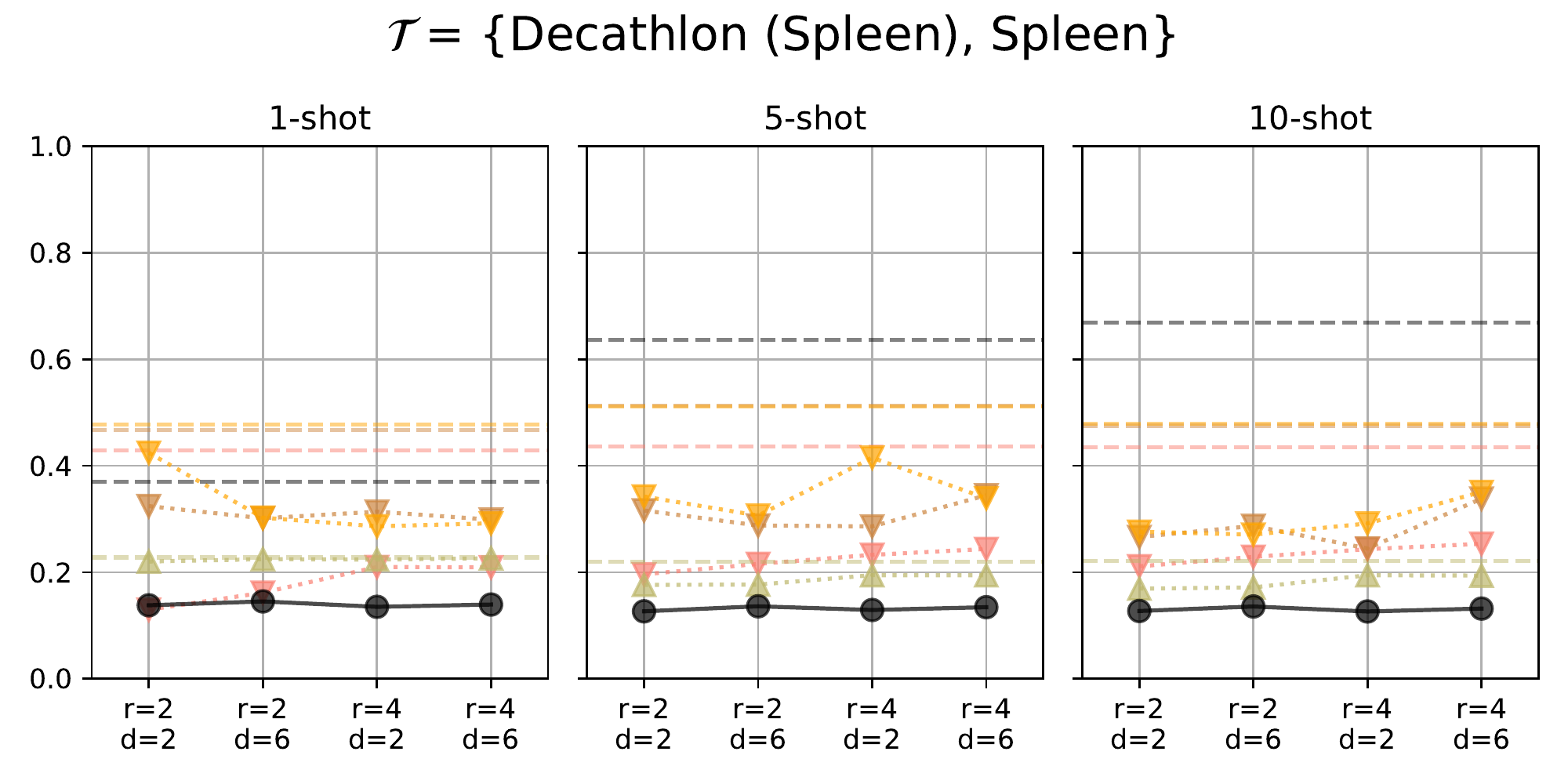}
        \caption{\textit{MSD-spleen}/\textit{skeleton}.}
        \label{fig:results_weak_annotations_3d_skeleton_spleen}
    \end{subfigure}
    \\
    \begin{subfigure}[b]{\currprop}
        % \centering
        \includegraphics[clip,trim=0.15in 0.1in 0in 0.5in,width=\columnwidth]{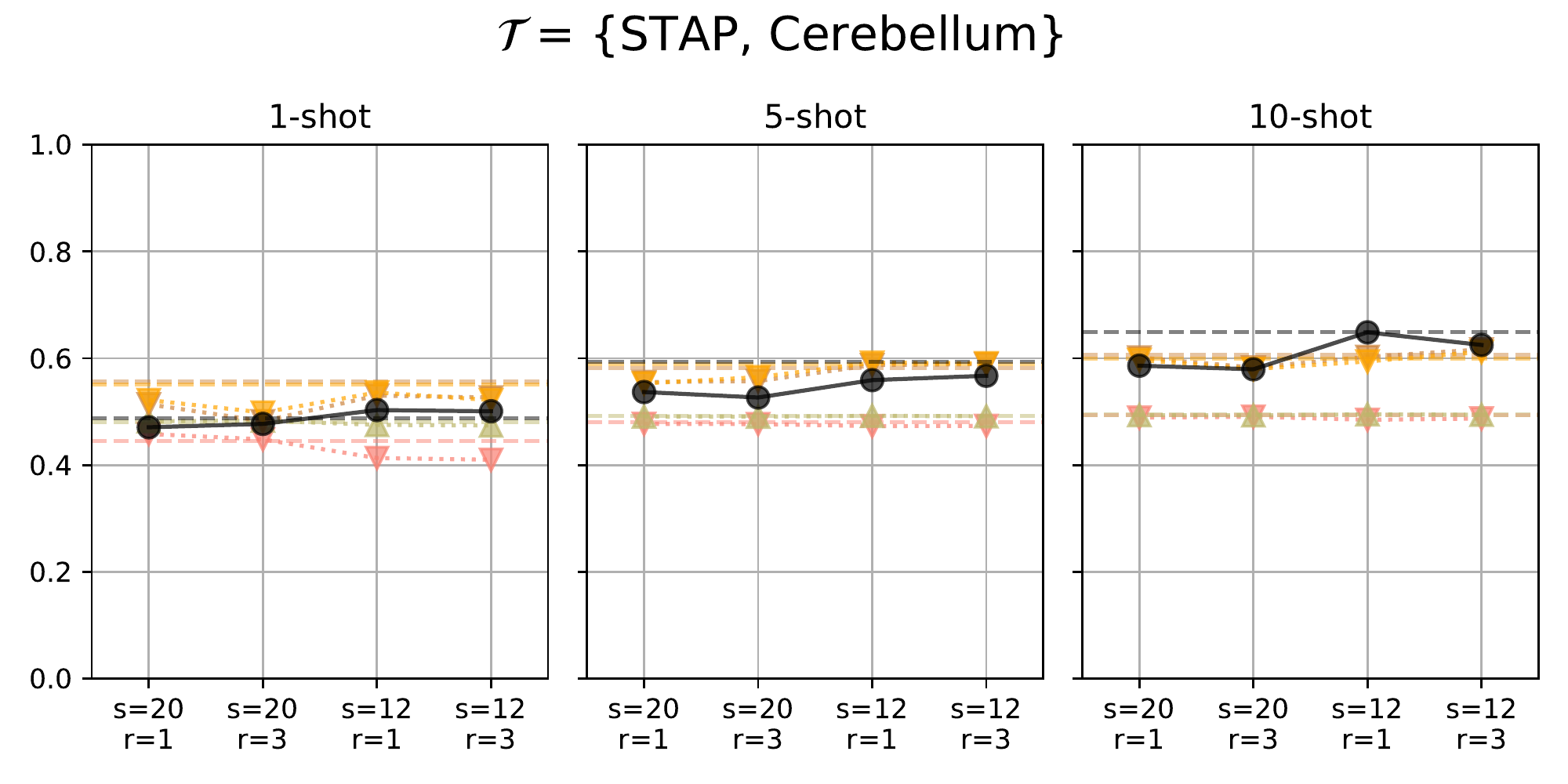}
        \caption{\textit{STAP-cerebellum}/\textit{grid}.}
        \label{fig:results_weak_annotations_3d_grid_cerebellum}
    \end{subfigure}
    \hfill
    \begin{subfigure}[b]{\currprop}
        % \centering
        \includegraphics[clip,trim=0.15in 0.1in 0in 0.5in,width=\columnwidth]{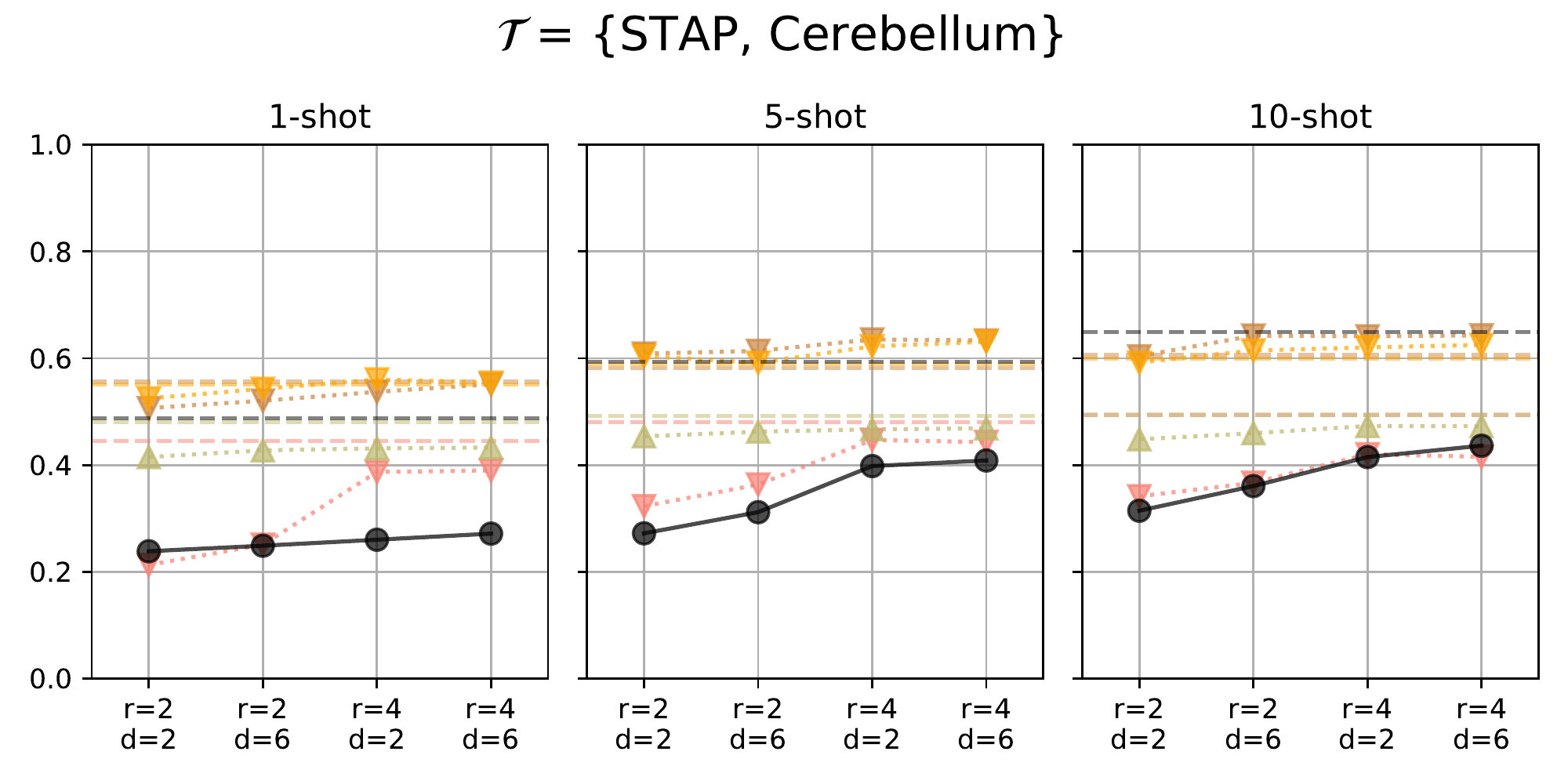}
        \caption{\textit{STAP-cerebellum}/\textit{skeleton}.}
        \label{fig:results_weak_annotations_3d_skeleton_cerebellum}
    \end{subfigure}
    \caption{IoU results for multiple weakly-supervised annotation styles in 1-, 5- and 10-shots for the best baseline backbone (\textbf{R}) and the two overall best algorithm/backbone pairs in each paradigm (ANIL-\textbf{D}, PANets-\textbf{E}, R2D2-\textbf{D} and MetaOptNet-\textbf{D}). Rows reflect distinct tasks, respectively: \textit{StructSeg-lungs}, \textit{StructSeg-heart}, \textit{MSD-spleen} and \textit{STAP-cerebellum}. The leftmost column represents results for \textit{grid} annotations, while the right column depicts \textit{skeleton} results. Dotted lines reflect the performance of segmentation algorithms on the weakly-supervised support sets, while the dashed horizontal lines depict the results tuned on the densely annotated support masks. Better viewed in color.}
    \label{fig:results_weak_annotations_3d}
\end{figure*}

\renewcommand{\currprop}{0.93\textwidth}
\begin{figure*}[!th]
    \centering
    \includegraphics[clip,trim=0.0in 0.45in 0.0in 0.0in,width=\currprop]{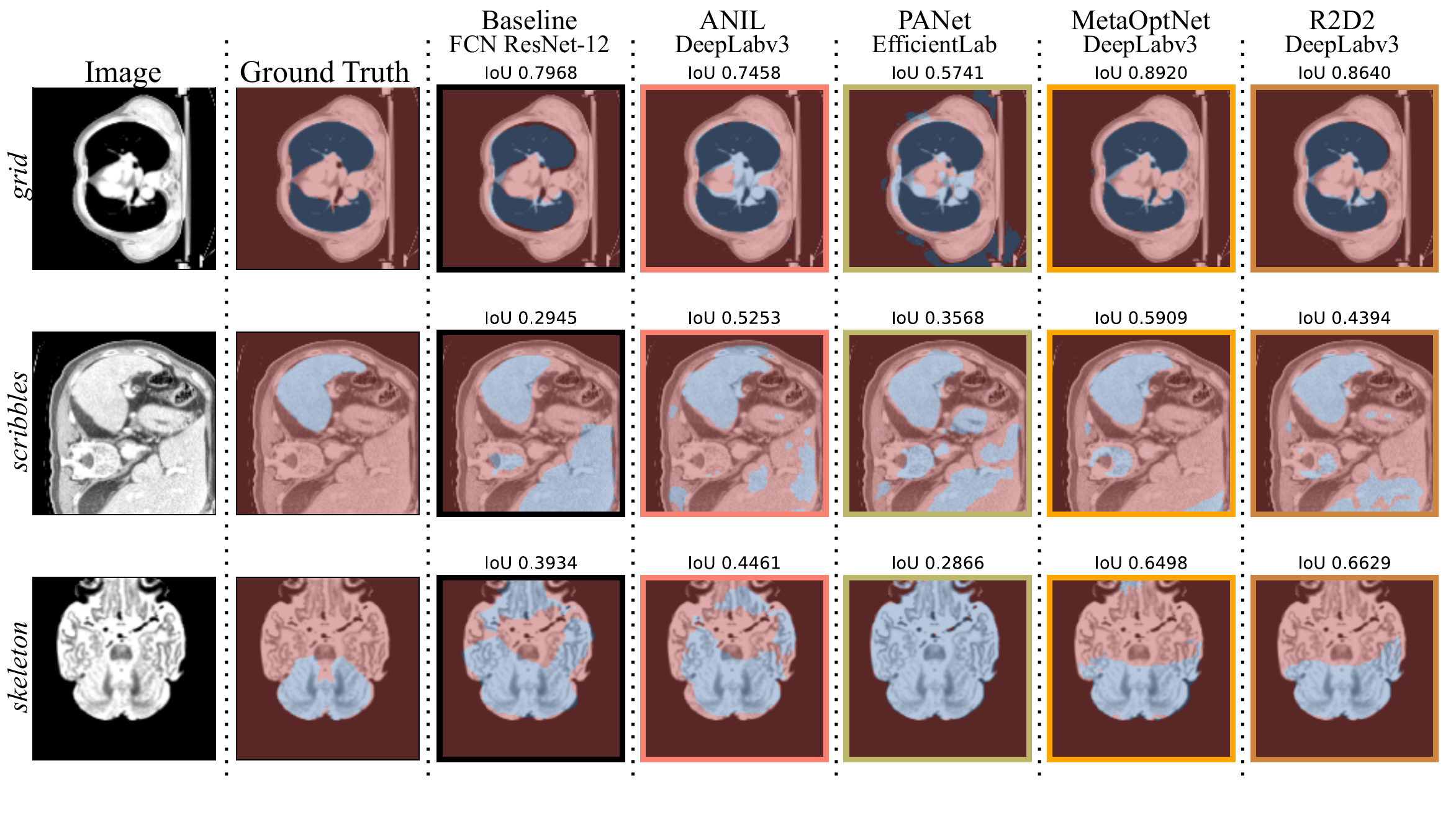}
    \caption{Qualitative results for FWS on three target tasks: \textit{StructSeg-lungs} on \textit{grid} annotations (top row), \textit{MSD-spleen} on \textit{scribble} annotations (middle row) and \textit{STAP-cerebellum} on \textit{skeleton} annotations (bottom row). Meta-learners and annotation styles are analogous to Figure~\ref{fig:results_qualitative}. Better viewed in color.}
    \label{fig:results_qualitative_3d}
\end{figure*}

Another interesting trend shown in Figures~\ref{fig:results_weak_annotations_3d_skeleton_lungs},~\ref{fig:results_weak_annotations_3d_skeleton_heart},~\ref{fig:results_weak_annotations_3d_skeleton_spleen} and~\ref{fig:results_weak_annotations_3d_skeleton_cerebellum} is that the baseline method was clearly outperformed by all meta-learners in the \textit{skeleton} annotation style for all tasks, but PANets in the \textit{StructSeg-lungs} task. In contrast to that, the baseline had a very competitive performance for \textit{grid} annotations, even outperforming all meta-learners in most sparsity configurations in Figures~\ref{fig:results_weak_annotations_3d_grid_heart} (\textit{StructSeg-heart}) and~\ref{fig:results_weak_annotations_3d_grid_spleen} (\textit{MSD-spleen}). Again, we attribute this better performance of the baseline to the higher density of annotations in \textit{grid}. This implies that the baseline is less adaptable than meta-learners in highly sparse and unorthodox annotation styles.

Finally, we show segmentation results for three of the 2D slice tasks in Figure~\ref{fig:results_qualitative_3d}. As the domain shift is larger for these tasks, similarly to the \textit{Panoramic-mandible} task, R2D2 and MetaOptNet achieved the overall best FWS results in comparison to the baseline and gradient-/metric-based strategies. More specifically, the metric-based PANet performed particularly poorly in these high domain-shifted tasks, segmenting other organs as the structure of interest in the majority of cases. This implies that metric-based methods tend to be unable to correctly deal with the inherent ambiguity of FWS tasks, possibly due to the difficulties in creating an embedding space that is discriminative to a wide range of real-world unseen OOD tasks.

% \todo{Comparison across 2D datasets (metrics, proximity between sources and targets).}

% \todo{Comparison between segmentation architectures (metrics).}

% \todo{Comparison between algorithms (metrics and train/adapt/test time).}

% \todo{Comparison between annotation types (metrics).}

% \todo{Comparison between image resolutions (metrics and train/adapt/test time).}

% \todo{Comparisons with and without attention and positional encoding (metrics).}

% \todo{HR Experiments 256x256 and 512x512 (metrics and train/adapt/test time).}

\section{Conclusion}
\label{sec:conclusion}

In this work we proposed a method for generalizing meta-learners from three distinct paradigms (gradient-based \cite{finn2017model,li2017metasgd,raghu2019rapid,nichol2018reptile}, metric-based \cite{snell2017prototypical,wang2019panet} and fusion-based \cite{bertinetto2018meta,lee2019meta,rakelly2018few}) for FWS tasks without requiring ImageNet pretraining or strong domain-dependent priors. We chose to apply our methodology to radiology images, even though the same methods should also apply to other non-RGB domains such as histopathology, remote sensing, seismic images, or even 1D temporal signals.

For gradient-based methods, we observed that performance peaks when applying low-cost second-order gradient-based algorithms (i.e. ANIL \cite{raghu2019rapid}% and MetaANIL \cite{li2017metasgd,raghu2019rapid}
) in the segmentation head. These methods are also more computationally effective than second-order optimization-based methods applied to the whole network (i.e. MAML \cite{finn2017model} or MetaSGD \cite{li2017metasgd}), which should allow them to be applied in scenarios that require inference on higher spatial resolution images or even in 3D radiology.
% than apply second-order gradients across the whole model, including feature extractor. 
Fully first-order methods (i.e. Reptile \cite{nichol2018reptile}), however, were not able to reach the same performance as ANIL, even if they are relatively quicker and more scalable to larger backbones. Future works in gradient-based methods might include alternative supervised loss functions (i.e. Dice \cite{milletari2016v} or Focal \cite{lin2017focal}% or Structural Similarity Index \cite{wang2003multiscale}
) and/or more label-efficient segmentation heads \cite{huang2020unet}.

In metric-based meta-learners, the cosine distance and prototype alignment regularization of PANets \cite{wang2019panet} proved to be more powerful than the simple Euclidean distance of ProtoNets \cite{snell2017prototypical,gama2022weakly}. Similarity-based methods, however, are only able to generalize to novel domains that are quite close to the meta-training domains/tasks, possibly due to the lack of global context in such models. Early experiments during this work tried to insert pixel location information in the distance computation with no visible effects for ProtoNets and PANets. Future research directions might be more successful in integrating local information with global context in such a way that benefits segmentation tasks (i.e. via CRFs \cite{zheng2015conditional} or visual attention \cite{hu2019attention}).% by preventing unrelated organs to be segmented as positive pixels, as seen in Figure~\ref{fig:results_qualitative_3d}.

Previous works on FWS, such as Guided Nets \cite{rakelly2018few} and PANets \cite{wang2019panet}, proved unable to adapt to novel domains with large domain shifts in relation to the meta-training datasets. In such scenarios, fusion-based approaches~\cite{bertinetto2018meta,lee2019meta} were considered the optimal choices, followed by gradient-based methods~\cite{raghu2019rapid}. These methods are also more computationally effective than second-order optimization-based methods applied to the whole network (i.e. MAML \cite{finn2017model} or MetaSGD~\cite{li2017metasgd}), which should allow them to be applied in scenarios that require inference on higher spatial resolution images or even in 3D radiology.

Very challenging segmentation tasks that require context and texture analysis, with organs that do not have clearly defined borders (such as \textit{STAP-cerebellum} \cite{oliveira2021automatic}) are still quite hard to learn from in few-shot settings. In future works, we intend to port the methods presented in this letter to fully 3D data (CTs, MRIs, and PET scans) instead of reducing these volumes to 2D slices. We hope that the additional context from the 3$^{rd}$ dimension will aid the algorithms in learning these challenging 3D tasks, despite the higher computation cost involved in learning 3D convolutional kernels.

At last, another major limitation of our pipeline is the need for annotated data from related tasks, restricting the application of the Meta-Learning pretraining to domains wherein labeled data is available for multiple datasets. Aiming to mitigate this limitation, another promising direction might be to merge Meta-Learning with SSL pseudolabels in order to eliminate the need of annotated datasets from related domains.

% \todo{Mention that the results for EfficientLab using R2D2 and MetaOptNet (in comparison to ProtoNet and PANets) reinforce the original findings from MetaOptNet's paper \cite{lee2019meta}: "Linear classifiers offer better generalization than nearest neighbor classifiers at a modest increase in computational costs. Our experiments suggest that regularized linear models allow significantly higher embedding dimensions with reduced overfitting."}

% \todo{ROBERTO: Incluir seção STAP. Motivação: problema desafiador pois recém-nascidos e crianças estão em desenvolvimento, mencionar diferenças nas RMI, etc. Além disso, 3D. mostrar que os resultados não ficam bons. Metalearning tem potencial substancial devido à escassez de dados públicos. Colocar imagens de exemplo. Future work: 3D}

% \todo{Future works: guidelines to port Meta-Learners for weakly-supervised segmentation in 3D images (without using 2D slices). GPU RAM limitations and running times must be considered and algorithms as R2D2 and MetaOptNets (Ridge) seem to be very promising for this application.}

\section*{Acknowledgements}

The authors would like to thank FAPESP (grants \#2015/22308-2, \#2017/50236-1 and \#2020/06744-5), Serrapilheira Institute (grant \#R-2011-37776), and ANR (ANR-FAPESP project \#ANR-17-CE23-0021) for their financial support for this research.

\bibliography{mybibfile}

\begin{thebibliography}{10}
\expandafter\ifx\csname url\endcsname\relax
  \def\url#1{\texttt{#1}}\fi
\expandafter\ifx\csname urlprefix\endcsname\relax\def\urlprefix{URL }\fi
\expandafter\ifx\csname href\endcsname\relax
  \def\href#1#2{#2} \def\path#1{#1}\fi

\bibitem{krizhevsky2012imagenet}
A.~Krizhevsky, I.~Sutskever, G.~E. Hinton, {ImageNet Classification with Deep
  Convolutional Neural Networks}, {NIPS} 25.

\bibitem{peng2021medical}
J.~Peng, Y.~Wang, {Medical Image Segmentation with Limited Supervision: A
  Review of Deep Network Models}, {IEEE Access}.

\bibitem{oliveira2022domain}
H.~Oliveira, R.~M. Cesar, P.~H. Gama, J.~A. Dos~Santos, {Domain Generalization
  in Medical Image Segmentation via Meta-Learners}, in: {Conference on
  Graphics, Patterns and Images}, Vol.~1, IEEE, 2022, pp. 288--293.

\bibitem{deng2009imagenet}
J.~Deng, W.~Dong, R.~Socher, L.-J. Li, K.~Li, L.~Fei-Fei, {ImageNet: A
  Large-scale Hierarchical Image Database}, in: {CVPR}, IEEE, 2009, pp.
  248--255.

\bibitem{yalniz2019billion}
I.~Z. Yalniz, H.~J{\'e}gou, K.~Chen, M.~Paluri, D.~Mahajan, Billion-scale
  semi-supervised learning for image classification, arXiv preprint
  arXiv:1905.00546.

\bibitem{everingham2015pascal}
M.~Everingham, S.~Eslami, L.~Van~Gool, C.~K. Williams, J.~Winn, A.~Zisserman,
  {The Pascal Visual Object Classes Challenge: A Retrospective}, {IJCV} 111~(1)
  (2015) 98--136.

\bibitem{young2014image}
P.~Young, A.~Lai, M.~Hodosh, J.~Hockenmaier, From image descriptions to visual
  denotations: New similarity metrics for semantic inference over event
  descriptions, Transactions of the Association for Computational Linguistics 2
  (2014) 67--78.

\bibitem{jing2020self}
L.~Jing, Y.~Tian, {Self-Supervised Visual Feature Learning with Deep Neural
  Networks: A Survey}, {IEEE TPAMI}.

\bibitem{hospedales2021meta}
T.~Hospedales, A.~Antoniou, P.~Micaelli, A.~Storkey, {Meta-Learning in Neural
  Networks: A Survey}, {IEEE TPAMI} 44~(9) (2021) 5149--5169.

\bibitem{luo2022meta}
S.~Luo, Y.~Li, P.~Gao, Y.~Wang, S.~Serikawa, {Meta-Seg: A Survey of
  Meta-Learning for Image Segmentation}, {Pattern Recognition} (2022) 108586.

\bibitem{wang2022generalizing}
J.~Wang, C.~Lan, C.~Liu, Y.~Ouyang, T.~Qin, W.~Lu, Y.~Chen, W.~Zeng, P.~Yu,
  {Generalizing to Unseen Domains: A Survey on Domain Generalization}, {IEEE
  Transactions on Knowledge and Data Engineering}.

\bibitem{finn2017model}
C.~Finn, P.~Abbeel, S.~Levine, {Model-Agnostic Meta-Learning for Fast
  Adaptation of Deep Networks}, in: {ICML}, PMLR, 2017, pp. 1126--1135.

\bibitem{nichol2018reptile}
A.~Nichol, J.~Schulman, {Reptile: A Scalable Metalearning Algorithm}, arXiv
  preprint arXiv:1803.02999 2~(3) (2018) 4.

\bibitem{li2017metasgd}
Z.~Li, F.~Zhou, F.~Chen, H.~Li, {Meta-SGD: Learning to Learn Quickly for
  Few-Shot Learning}, arXiv preprint arXiv:1707.09835.

\bibitem{raghu2019rapid}
A.~Raghu, M.~Raghu, S.~Bengio, O.~Vinyals, {Rapid Learning or Feature Reuse?
  Towards Understanding the Effectiveness of MAML}, arXiv preprint
  arXiv:1909.09157.

\bibitem{snell2017prototypical}
J.~Snell, K.~Swersky, R.~Zemel, {Prototypical Networks for Few-Shot Learning},
  {NeurIPS} 30.

\bibitem{wang2019panet}
K.~Wang, J.~H. Liew, Y.~Zou, D.~Zhou, J.~Feng, {PANet: Few-shot Image Semantic
  Segmentation with Prototype Alignment}, in: {CVPR}, 2019, pp. 9197--9206.

\bibitem{lee2019meta}
K.~Lee, S.~Maji, A.~Ravichandran, S.~Soatto, {Meta-Learning with Differentiable
  Convex Optimization}, in: {CVPR}, 2019, pp. 10657--10665.

\bibitem{bertinetto2018meta}
L.~Bertinetto, J.~F. Henriques, P.~H. Torr, A.~Vedaldi, {Meta-Learning with
  Differentiable Closed-form Solvers}, arXiv preprint arXiv:1805.08136.

\bibitem{rakelly2018few}
K.~Rakelly, E.~Shelhamer, T.~Darrell, A.~A. Efros, S.~Levine, {Few-Shot
  Segmentation Propagation with Guided Networks}, arXiv preprint
  arXiv:1806.07373.

\bibitem{gama2022weakly}
P.~H.~T. Gama, H.~N. Oliveira, J.~Marcato, J.~Dos~Santos, {Weakly Supervised
  Few-Shot Segmentation Via Meta-Learning}, {IEEE Transactions on Multimedia}.

\bibitem{hendryx2019meta}
S.~M. Hendryx, A.~B. Leach, P.~D. Hein, C.~T. Morrison, {Meta-Learning
  Initializations for Image Segmentation}, arXiv preprint arXiv:1912.06290.

\bibitem{gama2021learning}
P.~H. Gama, H.~Oliveira, J.~A. dos Santos, {Learning to Segment Medical Images
  from Few-Shot Sparse Labels}, in: {Conference on Graphics, Patterns and
  Images}, IEEE, 2021, pp. 89--96.

\bibitem{milletari2016v}
F.~Milletari, N.~Navab, S.-A. Ahmadi, {V-net: Fully Convolutional Neural
  Networks for Volumetric Medical Image Segmentation}, in: {International
  Conference on 3D Vision}, IEEE, 2016, pp. 565--571.

\bibitem{lin2017focal}
T.-Y. Lin, P.~Goyal, R.~Girshick, K.~He, P.~Doll{\'a}r, {Focal Loss for Dense
  Object Detection}, in: {ICCV}, 2017, pp. 2980--2988.

\bibitem{mensink2013distance}
T.~Mensink, J.~Verbeek, F.~Perronnin, G.~Csurka, {Distance-based Image
  Classification: Generalizing to New Classes at Near-Zero Cost}, {IEEE TPAMI}
  35~(11) (2013) 2624--2637.

\bibitem{zhang2020sg}
X.~Zhang, Y.~Wei, Y.~Yang, T.~S. Huang, {SG-One: Similarity Guidance Network
  for One-Shot Semantic Segmentation}, {IEEE Transactions on Cybernetics}
  50~(9) (2020) 3855--3865.

\bibitem{ronneberger2015u}
O.~Ronneberger, P.~Fischer, T.~Brox, {U-net: Convolutional Networks for
  Biomedical Image Segmentation}, in: {MICCAI}, Springer, 2015, pp. 234--241.

\bibitem{chen2017rethinking}
L.-C. Chen, G.~Papandreou, F.~Schroff, H.~Adam, {Rethinking Atrous Convolution
  for Semantic Image Segmentation}, arXiv preprint arXiv:1706.05587.

\bibitem{kingma2014adam}
D.~P. Kingma, J.~Ba, {Adam: A Method for Stochastic Optimization}, arXiv
  preprint arXiv:1412.6980.

\bibitem{simpson2019large}
A.~L. Simpson, M.~Antonelli, S.~Bakas, M.~Bilello, K.~Farahani,
  B.~Van~Ginneken, A.~Kopp-Schneider, B.~A. Landman, G.~Litjens, B.~Menze,
  et~al., {A Large Annotated Medical Image Dataset for the Development and
  Evaluation of Segmentation Algorithms}, arXiv preprint arXiv:1902.09063.

\bibitem{antonelli2022medical}
M.~Antonelli, A.~Reinke, S.~Bakas, K.~Farahani, A.~Kopp-Schneider, B.~A.
  Landman, G.~Litjens, B.~Menze, O.~Ronneberger, R.~M. Summers, et~al., {The
  Medical Segmentation Decathlon}, {Nature Communications} 13~(1) (2022) 1--13.

\bibitem{oliveira2021automatic}
H.~Oliveira, L.~Penteado, J.~L. Maciel, S.~F. Ferraciolli, M.~S. Takahashi,
  I.~Bloch, R.~C. Junior, {Automatic Segmentation of Posterior Fossa Structures
  in Pediatric Brain MRIs}, in: {Conference on Graphics, Patterns and Images},
  IEEE, 2021, pp. 121--128.

\bibitem{pizer1987adaptive}
S.~M. Pizer, E.~P. Amburn, J.~D. Austin, R.~Cromartie, A.~Geselowitz, T.~Greer,
  B.~ter Haar~Romeny, J.~B. Zimmerman, K.~Zuiderveld, {Adaptive Histogram
  Equalization and Its Variations}, {Computer Vision, Graphics, and Image
  Processing} 39~(3) (1987) 355--368.

\bibitem{PANORAMICabdi2015automatic}
A.~H. Abdi, S.~Kasaei, M.~Mehdizadeh, {Automatic Segmentation of Mandible in
  Panoramic X-Ray}, {Journal of Medical Imaging} 2~(4) (2015) 044003.

\bibitem{hsu2015unsupervised}
T.~M.~H. Hsu, W.~Y. Chen, C.-A. Hou, Y.-H.~H. Tsai, Y.-R. Yeh, Y.-C.~F. Wang,
  {Unsupervised Domain Adaptation with Imbalanced Cross-Domain Data}, in: ICCV,
  2015, pp. 4121--4129.

\bibitem{huang2020unet}
H.~Huang, L.~Lin, R.~Tong, H.~Hu, Q.~Zhang, Y.~Iwamoto, X.~Han, Y.-W. Chen,
  J.~Wu, {UNet 3+: A Full-Scale Connected UNet for Medical Image Segmentation},
  in: {ICASSP}, IEEE, 2020, pp. 1055--1059.

\bibitem{zheng2015conditional}
S.~Zheng, S.~Jayasumana, B.~Romera-Paredes, V.~Vineet, Z.~Su, D.~Du, C.~Huang,
  P.~H. Torr, {Conditional Random Fields as Recurrent Neural Networks}, in:
  {ICCV}, 2015, pp. 1529--1537.

\bibitem{hu2019attention}
T.~Hu, P.~Yang, C.~Zhang, G.~Yu, Y.~Mu, C.~G. Snoek, {Attention-based
  Multi-Context Guiding for Few-Shot Semantic Segmentation}, in: {AAAI
  Conference on Artificial Intelligence}, Vol.~33, 2019, pp. 8441--8448.

\end{thebibliography}

% \tableofcontents

\end{document}

\begin{frontmatter}

\title{Meta-Learners for Few-Shot Weakly Supervised Medical Image Segmentation -- Supplementary Material}
% \tnotetext[mytitlenote]{\todo{footnote...}}

\author[aff1]{Hugo Oliveira}
\ead{oliveirahugo@ime.usp.br}
\ead[url]{https://sites.google.com/view/oliveirahugo}
\affiliation[aff1]{
    organization={Institute of Mathematics and Statistics, Universidade de São Paulo},
    % addressline={R. do Matão, 1010},
    city={São Paulo},
    country={Brazil}
}
\author[aff2]{Pedro H. T. Gama}
\affiliation[aff2]{
    organization={Department of Computer Science, Universidade Federal de Minas Gerais},
    % addressline={R. do Matão, 1010},
    city={Belo Horizonte},
    country={Brazil}
}
\author[aff3]{Isabelle Bloch}
\affiliation[aff3]{
    organization={Sorbonne Universit\'e, CNRS, LIP6},
    % addressline={R. do Matão, 1010},
    city={Paris},
    country={France}
}
\author[aff1]{Roberto Marcondes Cesar Jr}

% \begin{abstract}
% \todo{...}
% \end{abstract}

% \begin{keyword}
% \todo{\texttt{elsarticle.cls}\sep \LaTeX\sep Elsevier \sep template
% \MSC[2010] 00-01\sep  99-00}
% \end{keyword}

\end{frontmatter}

\newcommand{\currprop}{\textwidth}

\linenumbers

\section{Meta-Learners for Weakly Supervised Segmentation}
\label{sec:metalearners}

\subsection{Simulating Weak Support Labels}
\label{sec:weak_labels}

\begin{figure*}[!th]
    \centering
    \hfil
    \begin{subfigure}[b]{0.475\columnwidth}
        % \centering
        \includegraphics[width=0.95\columnwidth]{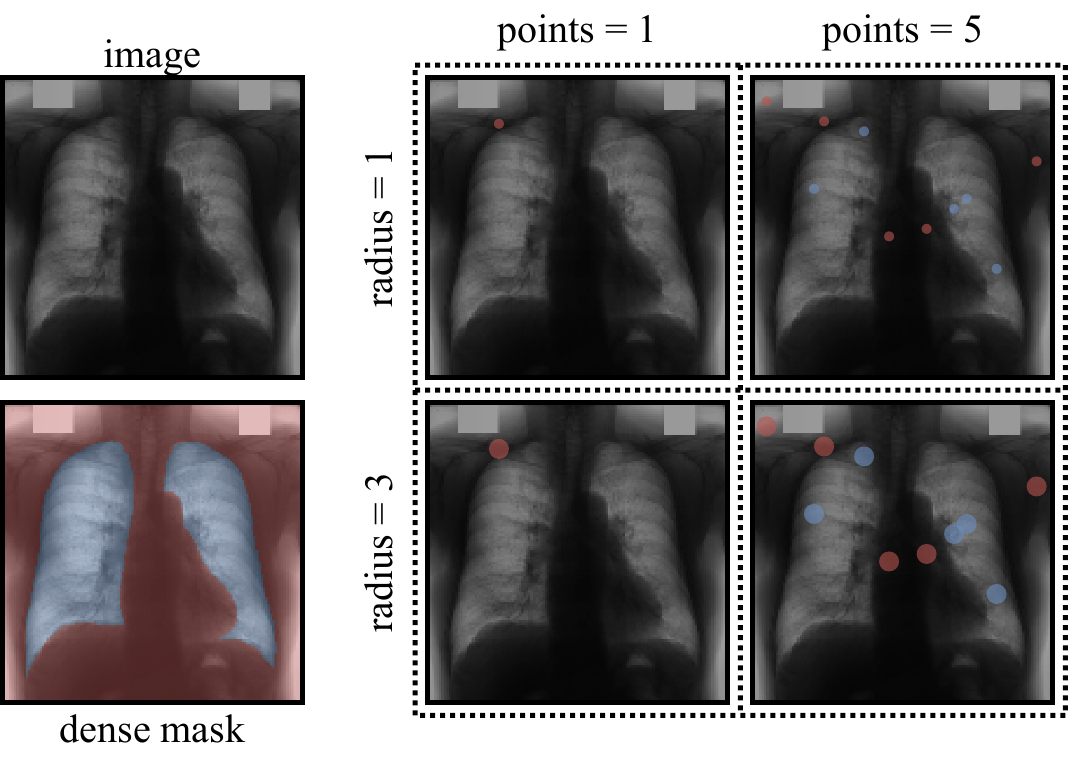}
        \caption{}
        \label{fig:weak_label_samples_points}
    \end{subfigure}
    \hfil
    \begin{subfigure}[b]{0.475\columnwidth}
        % \centering
        \includegraphics[width=0.95\columnwidth]{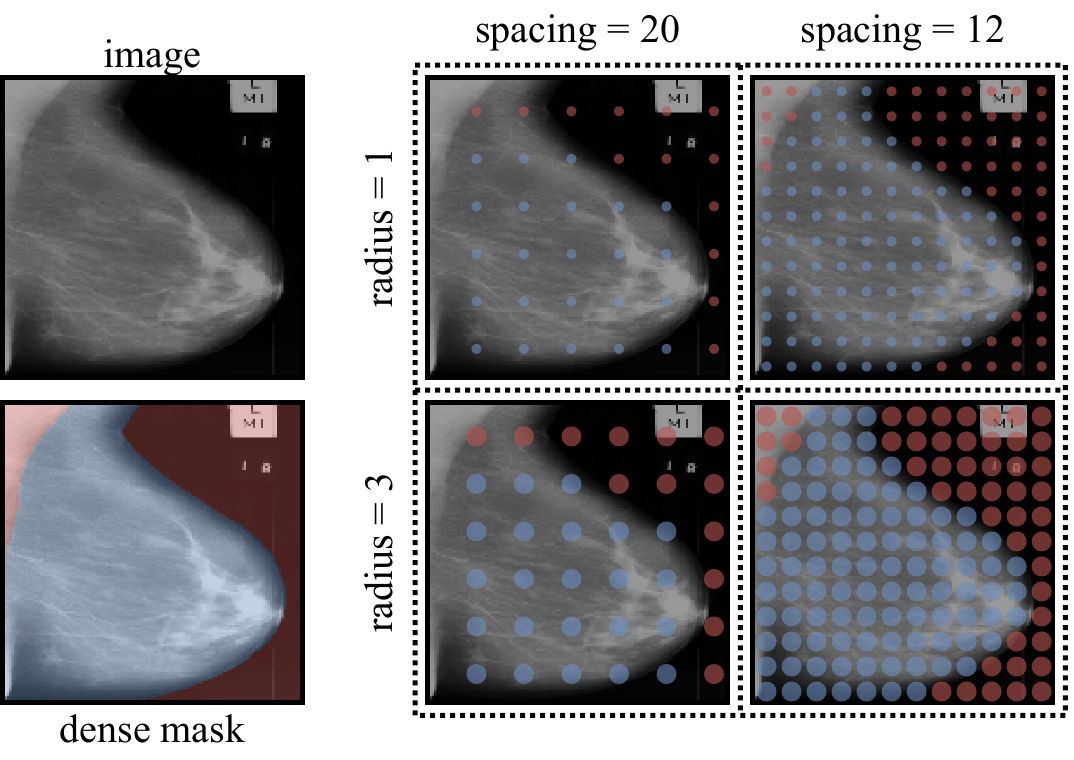}
        \caption{}
        \label{fig:weak_label_samples_grid}
    \end{subfigure}
    \hfil
    \\
    \hfil
    \begin{subfigure}[b]{0.475\columnwidth}
        % \centering
        \includegraphics[width=0.95\columnwidth]{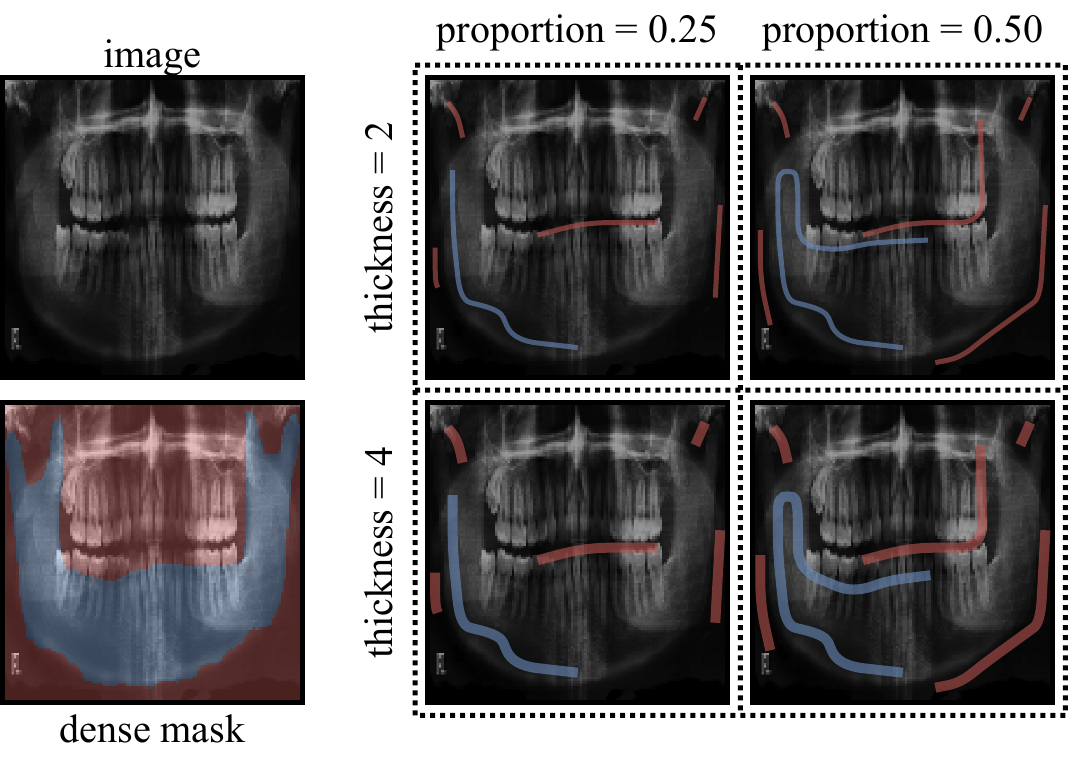}
        \caption{}
        \label{fig:weak_label_samples_scribbles}
    \end{subfigure}
    \hfil
    \begin{subfigure}[b]{0.475\columnwidth}
        % \centering
        \includegraphics[width=0.662121212\columnwidth]{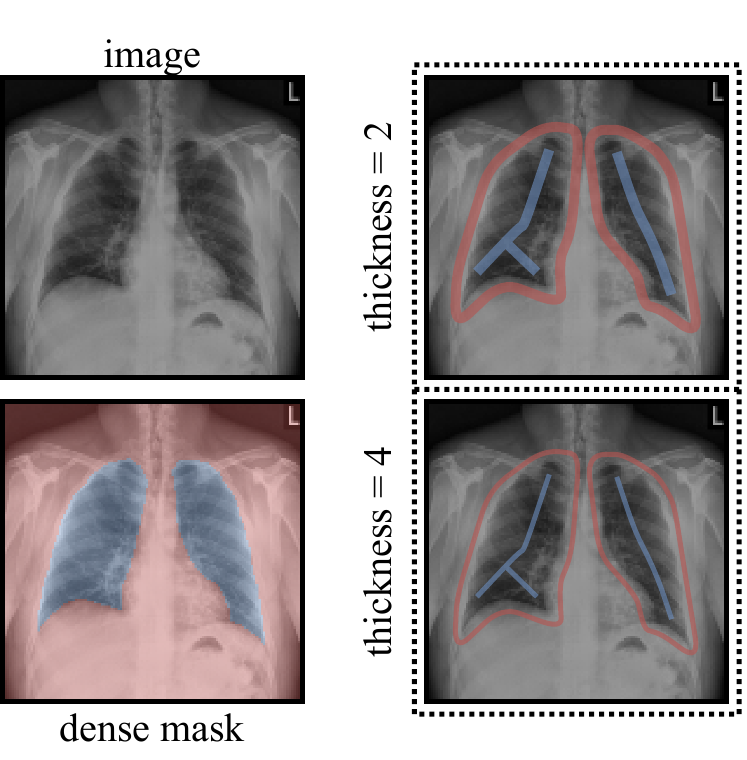}
        \caption{}
        \label{fig:weak_label_samples_skeleton}
    \end{subfigure}
    \hfil
    \caption{Weak labeling styles studied in this work: \textit{points}, \textit{grid}, \textit{scribbles} and \textit{skeleton}. The \textit{points} style can be parameterized by varying the number of points and radius of each labeled dot in the image, while \textit{grid} annotations vary in spacing and radius of each element. Line-based \textit{scribbles} annotations can be parameterized by changing the thickness and the proportion of the whole contour that is annotated, and, finally, \textit{skeleton} labels can be varied in density by changing the thickness of in- and outside lines.}
    \label{fig:weak_label_samples}
\end{figure*}

As discussed in the main text, the weakly annotated support set masks $\mathbf{y}^{sup}$ for each task $\T$ are generated from the dense labels by manipulating them using well-known image processing algorithms, such as morphological operations. We define and study four distinct types of weak annotations: \textit{points}, \textit{grid}, \textit{scribbles} and \textit{skeleton}. Examples for each annotation style can be seen in Figure~\ref{fig:weak_label_samples}.

Algorithms~\ref{alg:weak_annotations_check},~\ref{alg:weak_annotations_points_grid} and~\ref{alg:weak_annotations_scribbles_skeleton} describe the pseudo-algorithms used for computing weakly-supervised versions of the dense masks in the meta-dataset. In Algorithm~\ref{alg:weak_annotations_check} we define a generic function for checking for label consistency between the weak and dense masks, setting pixels with distinct known classes in \textsc{y} and \textsc{w} to unknown in order to remove inconsistencies in the support set labels. This function is called at the end of every sparsification method in Algorithms~\ref{alg:weak_annotations_points_grid} and~\ref{alg:weak_annotations_scribbles_skeleton}. Algorithm~\ref{alg:weak_annotations_points_grid} defined functions for computing two styles of point-based sparse masks: \textit{points} and \textit{grid}. In \textit{points}, \textsc{n\_pix} randomly sampled pixels are chosen to be set as known in the support set, with the possibility of dilating these dots to form larger disk regions. We highlight that the random choice of points in an image may be an unrepresentative sample of pixels and that a trained human oracle annotator would likely choose more representative points than the ones chosen stochastically. For the \textit{grid} annotation style the random nature of \textit{points} is mitigated by sampling a grid of points in the image and asking the oracle to select the positive ones. Again, the grid annotation style offers the choice of dilating the points to form circular disks in the image, allowing for more annotated support pixels to be fed to the model with no additional labor from the oracle.

\begin{algorithm}[!ht]
    % \setstretch{1.0}
    \caption{Algorithm for the self-checking function that aims to garantee that no wrongly annotated pixel is fed to the algorithms during the meta-training and tuning phases due to the sparsification algorithms. Indexing syntax for tensors in all algorithms are presented in a numpy-like fashion.}
    \label{alg:weak_annotations_check}
    \begin{algorithmic}[1]
    
        % \Require $x$: numpy-like array containing the image
        \Require \textsc{y}: numpy-like array containing the dense mask
        \Require \textsc{params}: dictionary containing sparsity parameters
        
        \Function{fix\_integrity}{\textsc{y}, \textsc{w}}
            
            \State \textsc{w}[\textsc{y} == 0 and \textsc{w} == 1] $\leftarrow$ unknown
            \State \textsc{w}[\textsc{y} == 1 and \textsc{w} == 0] $\leftarrow$ unknown
            
            \State \Return \textsc{w}
            
        \EndFunction
        
    \end{algorithmic}
    % \setstretch{2.0}
\end{algorithm}

\begin{algorithm}[!ht]
    % \setstretch{1.0}
    \caption{Algorithms for the sparsification process of the \textit{points} (\textsc{weak\_points}()) and \textit{grid} (\textsc{weak\_grid}()) annotation styles. Parameters \textsc{n\_pix} and \textsc{rad} control the density of \textit{points} annotations, while \textsc{spc} and \textsc{rad} control the density of \textit{grid} labels.}
    \label{alg:weak_annotations_points_grid}
    \begin{algorithmic}[1]
        
        % \Require $x$: numpy-like array containing the image
        \Require \textsc{y}: numpy-like array containing the dense mask
        % \Require \textsc{params}: dictionary containing sparsity parameters
        
        \Function{weak\_points}{\textsc{w}, \textsc{params}}
            
            \State Get \textsc{n\_pix} $\leftarrow$ \textsc{params}[`n\_pixels'] \textcolor{gray}{\Comment{Number of known points.}}
            \State Get \textsc{rad} $\leftarrow$ \textsc{params}[`radius'] \textcolor{gray}{\Comment{Radius for final dilation.}}
            
            \State Copy \textsc{y} to \textsc{w}
            
            \State Sample \textsc{n\_pix} negative and positive pixel indices to $\pi_{neg}$ and $\pi_{pos}$
            \State Set pixels in \textsc{w} such that \textsc{y} == 0 and index not in $\pi_{neg}$ to unknown
            \State Set pixels in \textsc{w} such that \textsc{y} == 1 and index not in $\pi_{pos}$ to unknown
            
            \State Dilate known areas: \textsc{w} $\leftarrow$ \Call{dilate}{\textsc{w}, \textsc{rad}}
            \State \textsc{w} $\leftarrow$ \Call{fix\_integrity}{\textsc{y}, \textsc{w}}
            
            \State \Return \textsc{w}
            
        \EndFunction
        
        \Function{weak\_grid}{\textsc{w}, \textsc{params}}
            
            \State Get \textsc{spc} $\leftarrow$ \textsc{params}[`spacing'] \textcolor{gray}{\Comment{Spacing between known pixels.}}
            \State Get \textsc{rad} $\leftarrow$ \textsc{params}[`radius'] \textcolor{gray}{\Comment{Radius for final dilation.}}
            
            \State Copy \textsc{y} to \textsc{w}
            \State Set all pixels in \textsc{w} to unknown
            
            \State Set pixels in grid intervals to known: \textsc{w}[::\textsc{spc}, ::\textsc{spc}] $\leftarrow$ \textsc{y}[::\textsc{spc}, ::\textsc{spc}]
            
            \State Dilate known areas: \textsc{w} $\leftarrow$ \Call{dilate}{\textsc{w}, \textsc{rad}}
            \State \textsc{w} $\leftarrow$ \Call{fix\_integrity}{\textsc{y}, \textsc{w}}
            
            \State \Return \textsc{w}
            
        \EndFunction
        
    \end{algorithmic}
    % \setstretch{2.0}
\end{algorithm}

\begin{algorithm}[!ht]
    % \setstretch{1.0}
    \caption{Algorithms for the sparsification process of the \textit{scribbles} (\textsc{weak\_scribbles}()) and \textit{skeleton} (\textsc{weak\_skeleton}()) annotation styles. Parameters \textsc{prop} and \textsc{rad} control the density of \textit{scribbles}, while \textsc{rad} control the density of \textit{skeleton} labels.}
    \label{alg:weak_annotations_scribbles_skeleton}
    \begin{algorithmic}[1]
        
        % \Require $x$: numpy-like array containing the image
        \Require \textsc{y}: numpy-like array containing the dense mask
        % \Require \textsc{params}: dictionary containing sparsity parameters
        
        \Function{weak\_scribbles}{\textsc{w}, \textsc{params}}
            
            \State Get \textsc{prop} $\leftarrow$ \textsc{params}[`prop'] \textcolor{gray}{\Comment{Percentage of contours known.}}
            % \State Get \textsc{dist} $\leftarrow$ \textsc{params}[`dist'] \textcolor{gray}{\Comment{Radius for dilation of pos and neg masks.}}
            \State Get \textsc{rad} $\leftarrow$ \textsc{params}[`radius'] \textcolor{gray}{\Comment{Radius for final dilation.}}
            
            \State Copy \textsc{y} to \textsc{w}, \textsc{w}$_{neg}$ and \textsc{w}$_{pos}$
            \State Set all pixels in \textsc{w} to unknown
            \State Set all pixels in \textsc{w}$_{neg}$ and \textsc{w}$_{pos}$ to \textsc{False}
            
            \State Copy \textsc{y} to \textsc{w}$_{dil}$ and \textsc{w}$_{ero}$
            \State \tikzmk{A}Dilate \textsc{w}$_{dil}$: \textsc{w}$_{dil}$ $\leftarrow$ \Call{dilation}{\textsc{w}$_{dil}$} \textcolor{gray}{\Comment{Finding outer shape.}}\tikzmk{B}\boxit{neg}
            \State \tikzmk{A}Erode \textsc{w}$_{ero}$: \textsc{w}$_{ero}$ $\leftarrow$ \Call{erosion}{\textsc{w}$_{ero}$} \textcolor{gray}{\Comment{Finding inner shape.}}\tikzmk{B}\boxit{pos}
            % \State \tikzmk{A}Dilate \textsc{w}$_{dil}$: \textsc{w}$_{dil}$ $\leftarrow$ \Call{dilation}{\textsc{w}$_{dil}$, \textsc{dist}} \textcolor{gray}{\Comment{Finding outer shape.}}\tikzmk{B}\boxit{neg}
            % \State \tikzmk{A}Erode \textsc{w}$_{ero}$: \textsc{w}$_{ero}$ $\leftarrow$ \Call{erosion}{\textsc{w}$_{ero}$, \textsc{dist}} \textcolor{gray}{\Comment{Finding inner shape.}}\tikzmk{B}\boxit{pos}
            
            \State \tikzmk{A}Set pixels with non-uniform 8-neighborhoods in \textsc{w}$_{dil}$ to \textsc{True} in \textsc{w}$_{neg}$\tikzmk{B}\boxit{neg}
            \State \tikzmk{A}Set pixels with non-uniform 8-neighborhoods in \textsc{w}$_{ero}$ to \textsc{True} in \textsc{w}$_{pos}$\tikzmk{B}\boxit{pos}
            
            \State \tikzmk{A}Sample \textsc{prop}\% of the contours in \textsc{w}$_{neg}$; set other pixels to \textsc{False}\tikzmk{B}\boxit{neg}
            \State \tikzmk{A}Sample \textsc{prop}\% of the contours in \textsc{w}$_{pos}$; set other pixels to \textsc{False}\tikzmk{B}\boxit{pos}
            
            \State \tikzmk{A}Set regions in \textsc{w} to label 0 where \textsc{w}$_{neg}$ is \textsc{True}\tikzmk{B}\boxit{neg}
            \State \tikzmk{A}Set regions in \textsc{w} to label 1 where \textsc{w}$_{pos}$ is \textsc{True}\tikzmk{B}\boxit{pos}
            
            \State Dilate known areas: \textsc{w} $\leftarrow$ \Call{dilate}{\textsc{w}, \textsc{rad}}
            \State \textsc{w} $\leftarrow$ \Call{fix\_integrity}{\textsc{y}, \textsc{w}}
            
            \State \Return \textsc{w}
            
        \EndFunction
        
        \Function{weak\_skeleton}{\textsc{w}, \textsc{params}}
            
            % \State Get \textsc{dist} $\leftarrow$ \textsc{params}[`dist'] \textcolor{gray}{\Comment{Radius for dilation of neg masks.}}
            \State Get \textsc{rad} $\leftarrow$ \textsc{params}[`radius'] \textcolor{gray}{\Comment{Radius for final dilation.}}
            
            \State Copy \textsc{y} to \textsc{w}, \textsc{w}$_{neg}$ and \textsc{w}$_{pos}$
            \State Set all pixels in \textsc{w} to unknown
            \State Set all pixels in \textsc{w}$_{neg}$ and \textsc{w}$_{pos}$ to \textsc{False}
            
            \State Copy \textsc{y} to \textsc{w}$_{dil}$
            \State Dilate \textsc{w}$_{dil}$: \textsc{w}$_{dil}$ $\leftarrow$ \Call{dilation}{\textsc{w}$_{dil}$} \textcolor{gray}{\Comment{Finding outer shape.}}
            % \State Dilate \textsc{w}$_{dil}$: \textsc{w}$_{dil}$ $\leftarrow$ \Call{dilation}{\textsc{w}$_{dil}$, \textsc{dist}} \textcolor{gray}{\Comment{Finding outer shape.}}
            \State Set pixels with non-uniform 8-neighborhoods in \textsc{w}$_{dil}$ to \textsc{True} in \textsc{w}$_{neg}$
            \State Set regions in \textsc{w} to label 0 where \textsc{w}$_{neg}$ is \textsc{True}
            
            \State Copy \textsc{y} to \textsc{w}$_{skl}$
            \State Skeletonize \textsc{w}$_{skl}$: \textsc{w}$_{skl}$ $\leftarrow$ \Call{skeletonize}{\textsc{w}$_{skl}$} \textcolor{gray}{\Comment{Finding skeleton.}}
            % \State Skeletonize \textsc{w}$_{skl}$: \textsc{w}$_{skl}$ $\leftarrow$ \Call{skeletonize}{\textsc{w}$_{skl}$, \textsc{dist}} \textcolor{gray}{\Comment{Finding skeleton.}}
            \State Set pixels in \textsc{w}$_{pos}$ to \textsc{True} where \textsc{w}$_{skl}$ is equal to 1
            \State Set regions in \textsc{w} to label 1 where \textsc{w}$_{skl}$ is \textsc{True}
            
            \State Dilate known areas: \textsc{w} $\leftarrow$ \Call{dilate}{\textsc{w}, \textsc{rad}}
            \State \textsc{w} $\leftarrow$ \Call{fix\_integrity}{\textsc{y}, \textsc{w}}
            
            \State \Return \textsc{w}
            
        \EndFunction
        
    \end{algorithmic}
    % \setstretch{2.0}
\end{algorithm}

For each annotation style we vary the sparsity of annotations according to hyperparameters and analyse how label density influences the performance of meta-learners. Table~\ref{tab:sparsity} explicits the parameters analysed in our experiments for each annotation style.

\begin{table}[!th]
    % \setstretch{1.0}
    \centering
    \caption{Parameters analysed for each annotation style studied during the experiments described in this work.}
    \label{tab:sparsity}
    \begin{tabular}{@{}cccccc@{}}
        \toprule
        \textbf{\begin{tabular}[c]{@{}c@{}}annotation\\style\end{tabular}} & \textbf{parameter}  &  & \textbf{training} &  & \textbf{testing}         \\ \cmidrule(r){1-2} \cmidrule(lr){4-4} \cmidrule(l){6-6} 
        \multirow{2}{*}{\textbf{points}}                                   & \textsc{n\_pix}  &  & \{1-20\}          &  & \{1, 5, 10\}             \\
                                                                           & \textsc{rad}     &  & \{1-5\}           &  & \{1, 2, 3\}              \\ \cmidrule(r){1-2} \cmidrule(lr){4-4} \cmidrule(l){6-6} 
        \multirow{2}{*}{\textbf{grid}}                                     & \textsc{spc}    &  & \{20-12\}         &  & \{20, 16, 12\}           \\
                                                                           & \textsc{rad}     &  & \{1-5\}           &  & \{1, 2, 3\}              \\ \cmidrule(r){1-2} \cmidrule(lr){4-4} \cmidrule(l){6-6} 
        \multirow{2}{*}{\textbf{scribbles}}                                & \textsc{prop} &  & {[}0.1-1.0{]}     &  & \{0.1, 0.25, 0.50, 1.0\} \\
                                                                           & \textsc{rad}     &  & \{1-8\}           &  & \{1, 2, 4, 8\}           \\ \cmidrule(r){1-2} \cmidrule(lr){4-4} \cmidrule(l){6-6} 
        \textbf{skeleton}                                                  & \textsc{radius}     &  & \{1-8\}           &  & \{1, 2, 4, 8\}           \\ \bottomrule
    \end{tabular}
    % \setstretch{2.0}
\end{table}

\subsection{Pseudo-algorithms for FSWSS Meta-Learners}
\label{sec:pseudoalgorithms}

\subsection{Gradient-based Pseudo-algorithms}
\label{sec:pseudoalgorithms_gradient}

Algorithm~\ref{alg:gradient} shows our pipeline for gradient-based algorithms for FSWSS in mode detail. For each inner loop the framework samples a support set $S_{i} = \{ \mathbf{x}_{i}^{sup}, \mathbf{y}_{i}^{sup} \}$ and a query set $Q_{i} = \{ \mathbf{x}_{i}^{qry}, \mathbf{y}_{i}^{qry} \}$, both initially with dense masks from the meta-training dataset. The dense masks $\mathbf{y}_{i}^{sup}$ from the support are sparsified procedurally according to a randomly generated seed. Predictions $\hat{\mathbf{y}}_{i}^{sup}$ are computed by forwarding the support images through the backbone $\phi$ and segmentation head $h$, allowing for the computation of the loss $\ell^{sup}_{i}$ through some a supervised criterion $\Ls_{sup}$. The method-dependent inner updating function $U_{in}$ is then used to compute the task-specific parameters $\theta^{\phi, h}_{i}$ for task $\tau_{i}$, which yield $\phi_{i}$ and $h_{i}$. $\phi_{i}$ and $h_{i}$ are then fed the query set images $\mathbf{x}_{i}^{qry}$, yielding predictions $\hat{\mathbf{y}}_{i}^{qry}$ that are used to compute the inner loop loss $\ell^{qry}_{i}$. After the whole meta-batch composed of tasks $\tau = \{ \tau_{0}, \tau_{1}, \dots \}$ is fed through the base model, yielding task-specific parameters $\{ \theta^{\phi, h}_{0}, \theta^{\phi, h}_{1}, \dots \}$ and inner losses $\{ \ell^{qry}_{0}, \ell^{qry}_{1}, \dots \}$, the outer loop gradients to update $\phi$ and $h$ can be computed by feeding the averaged inner gradients to the outer update function $U_{out}$, also method-dependent. This \textbf{while} process then repeats until convergence is met, always with randomly selected sets of tasks $\tau \sim p(\mathbfcal{T})$, support and query samples and sparsity parameters for support masks.

\begin{algorithm}[!ht]
    % \setstretch{1.0}
    \caption{Meta-training phase for gradient-based algorithms.}
    \label{alg:gradient}
    \begin{algorithmic}
        \Require $p(\mathbfcal{T})$: distribution over tasks
        \Require $\Ls$: supervised task loss
        \Require $U_{in}$: updating function for the specific models in the inner loop
        \Require $U_{out}$: updating function for the meta-model
        % \Require $\alpha, \beta$: inner and outer loop step sizes
        
        \State Randomly initialize parameters $\theta^{\phi, h}$
        \While{not done}
            \State Sample batch of tasks $\tau \sim p(\mathbfcal{T})$
            \For{\textbf{each} $\tau_{i} \in \tau$}
                \State Sample support batch $S_{i} \leftarrow \{ \mathbf{x}_{i}^{sup}, \mathbf{y}_{i}^{sup} \}$ from $\tau_{i}$
                \State Sample query batch $Q_{i} \leftarrow \{ \mathbf{x}_{i}^{qry}, \mathbf{y}_{i}^{qry} \}$ from $\tau_{i}$
                \State Convert $\mathbf{y}_{i}^{sup} \in S_{i}$ to sparse annotations
                % \State Compute gradients for $\phi$ and $h$ for $\tau_{i}$: $\nabla_{i}^{\phi, h}\Ls(\mathbf{y}_{i}^{sup}, h(\phi(\mathbf{x}_{i}^{sup})))$
                % \State Update parameters $\theta^{\phi, h}_{i} \leftarrow \theta^{\phi, h} - \alpha\nabla_{i}^{\phi, h}$
                \State Compute prediction $\hat{\mathbf{y}}_{i}^{sup} \leftarrow h(\phi(\mathbf{x}_{i}^{sup}))$
                \State Compute support loss $\ell^{sup}_{i} \leftarrow \Ls( \mathbf{y}_{i}^{sup}, \hat{\mathbf{y}}_{i}^{sup} )$
                \State Optimize specialized model parameters $\theta^{\phi, h}_{i}$ from $\ell^{sup}_{i}$ via $U_{in}(\theta^{h}, \ell^{sup}_{i})$
                \State Obtain specialized models $\phi_{i}$ and $h_{i}$ from $\theta^{\phi, h}_{i}$
                \State Compute prediction $\hat{\mathbf{y}}_{i}^{qry} = h_{i}(\phi_{i}(\mathbf{x}_{i}^{qry}))$
                \State Compute query loss $\ell^{qry}_{i} \leftarrow \Ls( \mathbf{y}_{i}^{qry}, \hat{\mathbf{y}}_{i}^{qry} )$
            \EndFor
            \State Update $\theta^{\phi, h} \leftarrow \theta^{\phi, h} + U_{out}(\theta^{\phi, h}, \sum_{\tau_{i}}^{\tau} \ell^{qry}_{i})$
            % \State Update $\theta^{\phi, h} \leftarrow \theta^{\phi, h} - \beta\nabla^{\phi, h} \sum_{\tau_{i}}^{\tau}\Ls(\mathbf{y}_{i}^{qry}, h_{i}(\phi_{i}(\mathbf{x}_{i}^{qry})))$
        \EndWhile
    \end{algorithmic}
    % \setstretch{2.0}
\end{algorithm}

A noteworthy variation of the gradient-based pipeline can be found in ANIL \cite{raghu2019rapid}. Feature reuse was observed to be more important than rapid learning on the MAML framework \cite{finn2017model}, allowing for only adapting task-specific classification -- or segmentation -- head during the inner loops, while keeping the feature extractor $\phi$ frozen for all tasks in the meta-batch. This strategy is much more memory efficient than optimizing a new $\phi_{\star}$ in each inner loop, while being almost as effective. In fact, by allowing for a larger meta-batch and/or more adaptation steps over the support set to yield $h_{\star}$, ANIL can even surpass the original MAML in similar GPU memory settings. Our framework also allows for the implementation of ANIL-like pipelines using second-order meta-learners \cite{finn2017model,li2017metasgd}, as described in Algorithm~\ref{alg:anil}.

\begin{algorithm}[!ht]
    % \setstretch{1.0}
    \caption{Meta-training phase for gradient-based algorithms with ANIL \cite{raghu2019rapid}.}
    \label{alg:anil}
    \begin{algorithmic}
        \Require $p(\mathbfcal{T})$: distribution over tasks
        \Require $\Ls$: supervised task loss
        \Require $U_{in}$: updating function for the specific models in the inner loop
        \Require $U_{out}$: updating function for the meta-model
        % \Require $\alpha, \beta$: inner and outer loop step sizes
        
        \State Randomly initialize parameters $\theta^{\phi}$ and $\theta^{h}$
        \While{not done}
            \State Sample batch of tasks $\tau \sim p(\mathbfcal{T})$
            \For{\textbf{each} $\tau_{i} \in \tau$}
                \State Sample support batch $S_{i} \leftarrow \{ \mathbf{x}_{i}^{sup}, \mathbf{y}_{i}^{sup} \}$ from $\tau_{i}$
                \State Sample query batch $Q_{i} \leftarrow \{ \mathbf{x}_{i}^{qry}, \mathbf{y}_{i}^{qry} \}$ from $\tau_{i}$
                \State Convert $\mathbf{y}_{i}^{sup} \in S_{i}$ to sparse annotations
                % \State Compute gradients for $h$ for $\tau_{i}$: $\nabla_{i}^{h}\Ls(\mathbf{y}_{i}^{sup}, h(\phi(\mathbf{x}_{i}^{sup})))$
                % \State Update head parameters $\theta^{h}_{i} \leftarrow \theta^{h} - \alpha\nabla_{i}^{h}$
                \State Compute prediction $\hat{\mathbf{y}}_{i}^{sup} \leftarrow h(\phi(\mathbf{x}_{i}^{sup}))$
                \State Compute support loss $\ell^{sup}_{i} \leftarrow \Ls( \mathbf{y}_{i}^{sup}, \hat{\mathbf{y}}_{i}^{sup} )$
                \State Optimize specialized head parameters $\theta^{h}_{i}$ from $\ell^{sup}_{i}$ via $U_{in}(\theta^{h}, \ell^{sup}_{i})$
                \State Obtain specialized head $h_{i}$ from $\theta^{h}_{i}$
                \State Compute prediction $\hat{\mathbf{y}}_{i}^{qry} = h_{i}(\phi(\mathbf{x}_{i}^{qry}))$
                \State Compute query loss $\ell^{qry}_{i} \leftarrow \Ls( \mathbf{y}_{i}^{qry}, \hat{\mathbf{y}}_{i}^{qry} )$
            \EndFor
            \State Update $\theta^{h} \leftarrow \theta^{h} + U_{out}(\theta^{h}, \sum_{\tau_{i}}^{\tau} \ell^{qry}_{i})$
            \State Update $\theta^{\phi} \leftarrow \theta^{\phi} + U_{out}(\theta^{\phi}, \sum_{\tau_{i}}^{\tau} \ell^{qry}_{i})$
        \EndWhile
    \end{algorithmic}
    % \setstretch{2.0}
\end{algorithm}

\subsection{Metric-based Pseudo-algorithm}
\label{sec:pseudoalgorithms_metric}

Algorithm~\ref{alg:metric} describes in more detail the proposed metric-based pipeline for FSWSS. For each meta-training iteration initially an accumulator $J$ is set to zero, followed by sampling a meta-batch of tasks $\tau \sim p(\mathbfcal{T})$. Support $S_{i} = \{ \mathbf{x}_{i}^{sup}, \mathbf{y}_{i}^{sup} \}$ and query set $Q_{i} = \{ \mathbf{x}_{i}^{qry}, \mathbf{y}_{i}^{qry} \}$ sample/label pairs are sampled, compiling a batch for the task $\tau_{i}$. Feature representations for both the support $\mathbf{f}^{sup}$ and query $\mathbf{f}^{qry}$ sets are computed by forwarding the sets through $\phi$. The potentially few annotated support pixels $\mathbf{f}^{sup}$ are used to compute the centroids for the background $\mu_{0}$ and foreground $\mu_{1}$ classes by simply averaging their feature representations, resulting in the supporting embedding space $\mathbf{e}^{sup}$. Subsequently the query features are projected on the support embedding space, allowing for the computation of distances $d(\mathbf{f}^{qry}, \mu_{0})$ and $d(\mathbf{f}^{qry}, \mu_{1})$ to the support centroids. These distances can then be used compute logits of each query pixel in relation to the background $\mathbf{l}^{qry}_{0}$ and foreground $\mathbf{l}^{qry}_{1}$ classes by multiplying them by $-1.0$. At last, logits $\mathbf{l}^{qry}_{0}$ and $\mathbf{l}^{qry}_{1}$ can be plugged into $\Ls_{sup}$ or even to other selective supervised criterion coupled with $\mathbf{y}^{qry}$, yielding loss values that are accumulated into variable $J$. At the end of the meta-batch composed of multiple tasks $\tau_{i}$, the gradients $\nabla^{\phi} J$ are simply backpropagated through $\phi$, updating the feature extractor's weights. This algorithm simplifies the training process of metric-based algorithms to simple supervised training in deep neural networks.

\begin{algorithm}[!ht]
    % \setstretch{1.0}
    \caption{Training algorithm for metric-based methods.}
    \label{alg:metric}
    \begin{algorithmic}
        \Require $p(\mathbfcal{T})$: distribution over tasks
        \Require $\Ls$: supervised task loss
        \Require $d$: distance function
        
        \State Randomly initialize parameters from $\phi$
        \While{not done}
            \State $J \leftarrow 0$
            \State Sample batch of tasks $\tau \sim p(\mathbfcal{T})$
            \For{\textbf{each} $\tau_{i} \in \tau$}
                \State Sample support batch $S_{i} \leftarrow \{ \mathbf{x}_{i}^{sup}, \mathbf{y}_{i}^{sup} \}$ from $\tau_{i}$
                \State Sample query batch $Q_{i} \leftarrow \{ \mathbf{x}_{i}^{qry}, \mathbf{y}_{i}^{qry} \}$ from $\tau_{i}$
                \State Convert $\mathbf{y}_{i}^{sup} \in S_{i}$ to sparse annotations
                \State Obtain support feature representation $\mathbf{f}^{sup} \leftarrow \phi(\mathbf{x}_{i}^{sup})$
                \State Obtain query feature representation $\mathbf{f}^{qry} \leftarrow \phi(\mathbf{x}_{i}^{qry})$
                \State Compute $\mu_{0}$ and $\mu_{1}$ using $\mathbf{f}^{sup}$ %via Equation~\ref{eq:pixelwise_prototype}
                \State Project the query representations $\mathbf{f}^{qry}$ on the embedding space $\mathbf{e}^{sup}$
                \State Compute logits $\mathbf{l}^{qry}_{0} \leftarrow -1.0 \times d(\mathbf{f}^{qry}, \mu_{0})$ and $\mathbf{l}^{qry}_{1} \leftarrow - 1.0 \times d(\mathbf{f}^{qry}, \mu_{1})$
                \State $J \leftarrow J + \Ls(\mathbf{y}_{i}^{qry}, \{ \mathbf{l}^{qry}_{0}, \mathbf{l}^{qry}_{1} \})$
            \EndFor
            \State Update $\phi$ using gradient descent and $\nabla^{\phi} J$
        \EndWhile
    \end{algorithmic}
    % \setstretch{2.0}
\end{algorithm}

\subsection{Fusion-based Pseudo-algorithm}
\label{sec:pseudoalgorithms_fusion}

For fusion-based meta-learners, we employ the generalist algorithm described in Algorithm~\ref{alg:fusion}. Fusion-based methods use the support set (data and labels) to produce an embedding capable of guiding the task to be solved by the output head on the query set. This can be achieved through the use of merging operations $\otimes$ and $\oplus$ (i.e. addition, multiplication, concatenation, attention modules, or even simple tensorial regressors or support vector machines). Initially, $\mathbf{x}_{i}^{sup}$ is forwarded through $\phi$ and $\mathbf{y}_{i}^{sup}$ is passed through $\phi^{m}$, resulting in embeddings $\mathbf{f}^{sup}$ and $\mathbf{f}^{m}$, which are then fused via $\otimes$ and shrunk through averaging in the spatial dimensions, resulting in $\mathbf{f}^{\otimes}$. This embedding contains information of both support data and labels, allowing it to ``guide'' the segmentation of the query pixels according to the data and task annotated in the support. $\mathbf{f}^{\otimes}$ is then tiled to match the query spatial dimensions and fused with the query embedding $\mathbf{f}^{qry}$ obtained by forwarding $\mathbf{x}_{i}^{qry}$ through $\phi$. The final embedding $\mathbf{f}^{\oplus}$ is the result from this last fusion operation $\oplus$ and then passed through the segmentation head $h$. As in other meta-learning algorithms, the loss $J$ is accumulated across the inner loop and the gradients $\nabla^{\phi} J$ are used to update $\phi$ in the outer loop.

\begin{algorithm}[!ht]
    % \setstretch{1.0}
    \caption{Training algorithm for fusion-based methods.}
    \label{alg:fusion}
    \begin{algorithmic}
        \Require $p(\mathbfcal{T})$: distribution over tasks
        \Require $\Ls$: supervised task loss
        \Require $\otimes$: support/mask identity mapping
        \Require $\oplus$: query/support identity mapping
        
        \State Randomly initialize parameters from $\phi$, $\phi^{m}$ and $h$
        \While{not done}
            \State $J \leftarrow 0$
            \State Sample batch of tasks $\tau \sim p(\mathbfcal{T})$
            \For{\textbf{each} $\tau_{i} \in \tau$}
                \State Sample support batch $S_{i} \leftarrow \{ \mathbf{x}_{i}^{sup}, \mathbf{y}_{i}^{sup} \}$ from $\tau_{i}$
                \State Sample query batch $Q_{i} \leftarrow \{ \mathbf{x}_{i}^{qry}, \mathbf{y}_{i}^{qry} \}$ from $\tau_{i}$
                \State Convert $\mathbf{y}_{i}^{sup} \in S_{i}$ to sparse annotations
                \State Obtain embeddings $\mathbf{f}^{sup} \leftarrow \phi(\mathbf{x}_{i}^{sup})$ and $\mathbf{f}^{m} \leftarrow \phi^{m}(\mathbf{y}_{i}^{sup})$
                \State Compute $\mathbf{f}^{\otimes} \leftarrow average(\mathbf{f}^{sup} \otimes \mathbf{f}^{m})$
                \State Obtain embedded representation $\mathbf{f}^{qry} \leftarrow \phi(\mathbf{x}_{i}^{qry})$
                \State Compute $\mathbf{f}^{\oplus} \leftarrow \mathbf{f}^{qry} \oplus \mathbf{f}^{\otimes}$
                \State Predict $\hat{\mathbf{y}}_{i}^{qry} = h(\mathbf{f}^{\oplus})$
                \State $J \leftarrow J + \Ls(\mathbf{y}_{i}^{qry}, \hat{\mathbf{y}}_{i}^{qry})$
            \EndFor
            \State Update $\phi$ using gradient descent and $\nabla^{\phi} J$
        \EndWhile
    \end{algorithmic}
    % \setstretch{2.0}
\end{algorithm}

\section{Zoomed-in Qualitative Results}
\label{sec:extra_results}

\begin{figure*}[!th]
    \centering
    \renewcommand{\currprop}{0.030\textheight}
    \includegraphics[page=1,clip,trim=0.00in 19.40in 13.38in 0.35in,height=\currprop]{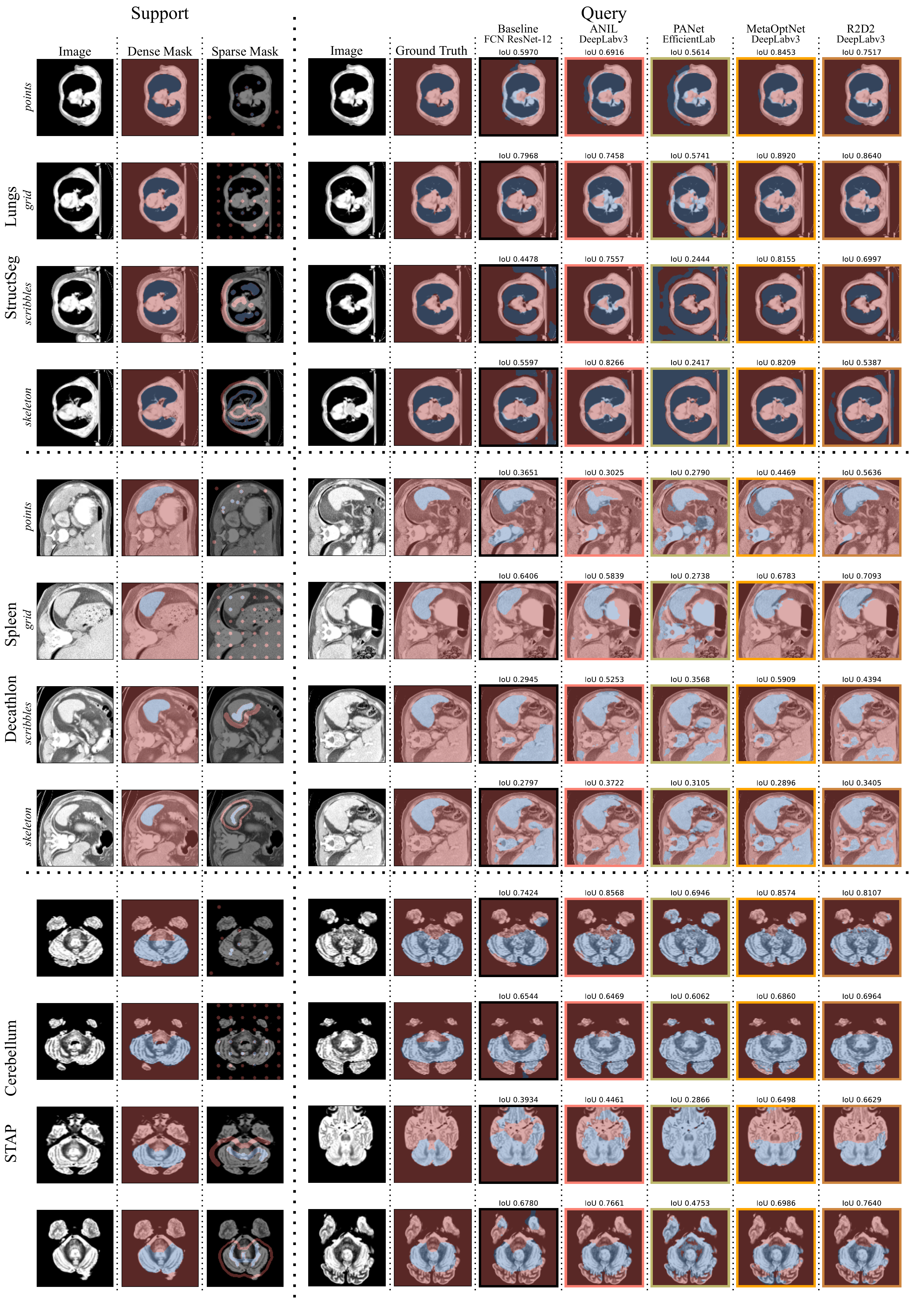}%
    \includegraphics[page=1,clip,trim=6.00in 19.40in 6.71in 0.35in,height=\currprop]{results_qualitative_3d_final.pdf}%
    \includegraphics[page=1,clip,trim=8.53in 19.40in 0.00in 0.35in,height=\currprop]{results_qualitative_3d_final.pdf}
    \\
    \renewcommand{\currprop}{0.555\textheight}
    \includegraphics[page=1,clip,trim=0.00in 13.21in 13.38in 0.705in,height=\currprop]{results_qualitative_3d_final.pdf}%
    \includegraphics[page=1,clip,trim=6.00in 13.21in 6.71in 0.705in,height=\currprop]{results_qualitative_3d_final.pdf}%
    \includegraphics[page=1,clip,trim=8.53in 13.21in 0.00in 0.705in,height=\currprop]{results_qualitative_3d_final.pdf}
    \caption{Zoomed-in qualitative results for FSWSS on the \textit{StructSeg-lungs} task with 4 distinct annotation styles (\textit{points}, \textit{grid}, \textit{scribbles} and \textit{skeleton}).}
    \label{fig:results_qualitative_structseg_lungs}
\end{figure*}

Qualitative results in FSWSS tasks for 2D (Figures~\ref{fig:results_qualitative_panoramic_mandible} and~\ref{fig:results_qualitative_openist_lungs}) and 3D (Figures~\ref{fig:results_qualitative_structseg_lungs},~\ref{fig:results_qualitative_msd_spleen} and~\ref{fig:results_qualitative_stap_cerebellum}) tasks. These figures are zoomed-in versions of the qualitative images from the main text, albeit with larger dimensions in order to allow for better visual inspection of the segmentation results.

\begin{figure*}[!th]
    \centering
    \renewcommand{\currprop}{0.028182667\textheight}
    \includegraphics[page=1,clip,trim=0.00in 13.01in 13.38in 0.35in,height=\currprop]{results_qualitative_final.pdf}%
    \includegraphics[page=1,clip,trim=6.00in 13.01in 6.71in 0.35in,height=\currprop]{results_qualitative_final.pdf}%
    \includegraphics[page=1,clip,trim=8.53in 13.01in 0.00in 0.35in,height=\currprop]{results_qualitative_final.pdf}
    \\
    \renewcommand{\currprop}{0.544\textheight}
    \includegraphics[page=1,clip,trim=0.00in 6.80in 13.38in 0.705in,height=\currprop]{results_qualitative_final.pdf}%
    \includegraphics[page=1,clip,trim=6.00in 6.80in 6.71in 0.705in,height=\currprop]{results_qualitative_final.pdf}%
    \includegraphics[page=1,clip,trim=8.53in 6.80in 0.00in 0.705in,height=\currprop]{results_qualitative_final.pdf}
    \caption{Zoomed-in qualitative results for FSWSS on the \textit{Panoramic-mandible} task with 4 distinct annotation styles (\textit{points}, \textit{grid}, \textit{scribbles} and \textit{skeleton}).}
    \label{fig:results_qualitative_panoramic_mandible}
\end{figure*}

\begin{figure*}[!th]
    \centering
    \renewcommand{\currprop}{0.028182667\textheight}
    \includegraphics[page=1,clip,trim=0.00in 13.01in 13.38in 0.35in,height=\currprop]{results_qualitative_final.pdf}%
    \includegraphics[page=1,clip,trim=6.00in 13.01in 6.71in 0.35in,height=\currprop]{results_qualitative_final.pdf}%
    \includegraphics[page=1,clip,trim=8.53in 13.01in 0.00in 0.35in,height=\currprop]{results_qualitative_final.pdf}
    \\
    \renewcommand{\currprop}{0.575\textheight}
    \includegraphics[page=1,clip,trim=0.00in 0.00in 13.38in 7.15in,height=\currprop]{results_qualitative_final.pdf}%
    \includegraphics[page=1,clip,trim=6.00in 0.00in 6.71in 7.15in,height=\currprop]{results_qualitative_final.pdf}%
    \includegraphics[page=1,clip,trim=8.53in 0.00in 0.00in 7.15in,height=\currprop]{results_qualitative_final.pdf}
    \caption{Zoomed-in qualitative results for FSWSS on the \textit{OpenIST-lungs} task with 4 distinct annotation styles (\textit{points}, \textit{grid}, \textit{scribbles} and \textit{skeleton}).}
    \label{fig:results_qualitative_openist_lungs}
\end{figure*}

\begin{figure*}[!th]
    \centering
    \renewcommand{\currprop}{0.030\textheight}
    \includegraphics[page=1,clip,trim=0.00in 19.40in 13.38in 0.35in,height=\currprop]{results_qualitative_3d_final.pdf}%
    \includegraphics[page=1,clip,trim=6.00in 19.40in 6.71in 0.35in,height=\currprop]{results_qualitative_3d_final.pdf}%
    \includegraphics[page=1,clip,trim=8.53in 19.40in 0.00in 0.35in,height=\currprop]{results_qualitative_3d_final.pdf}
    \\
    \renewcommand{\currprop}{0.5505\textheight}
    \includegraphics[page=1,clip,trim=0.00in 6.80in 13.38in 7.165in,height=\currprop]{results_qualitative_3d_final.pdf}%
    \includegraphics[page=1,clip,trim=6.00in 6.80in 6.71in 7.165in,height=\currprop]{results_qualitative_3d_final.pdf}%
    \includegraphics[page=1,clip,trim=8.53in 6.80in 0.00in 7.165in,height=\currprop]{results_qualitative_3d_final.pdf}
    \caption{Zoomed-in qualitative results for FSWSS on the \textit{MSD-spleen} task with 4 distinct annotation styles (\textit{points}, \textit{grid}, \textit{scribbles} and \textit{skeleton}).}
    \label{fig:results_qualitative_msd_spleen}
\end{figure*}

\begin{figure*}[!th]
    \centering
    \renewcommand{\currprop}{0.030\textheight}
    \includegraphics[page=1,clip,trim=0.00in 19.40in 13.38in 0.35in,height=\currprop]{results_qualitative_3d_final.pdf}%
    \includegraphics[page=1,clip,trim=6.00in 19.40in 6.71in 0.35in,height=\currprop]{results_qualitative_3d_final.pdf}%
    \includegraphics[page=1,clip,trim=8.53in 19.40in 0.00in 0.35in,height=\currprop]{results_qualitative_3d_final.pdf}
    \\
    \renewcommand{\currprop}{0.5878\textheight}
    \includegraphics[page=1,clip,trim=0.00in 0.00in 13.38in 13.55in,height=\currprop]{results_qualitative_3d_final.pdf}%
    \includegraphics[page=1,clip,trim=6.00in 0.00in 6.71in 13.55in,height=\currprop]{results_qualitative_3d_final.pdf}%
    \includegraphics[page=1,clip,trim=8.53in 0.00in 0.00in 13.55in,height=\currprop]{results_qualitative_3d_final.pdf}
    \caption{Zoomed-in qualitative results for FSWSS on the \textit{STAP-cerebellum} task with 4 distinct annotation styles (\textit{points}, \textit{grid}, \textit{scribbles} and \textit{skeleton}).}
    \label{fig:results_qualitative_stap_cerebellum}
\end{figure*}

\bibliography{mybibfile}